\documentclass[letterpaper]{article} 
\usepackage{aaai2026}  
\usepackage{times}  
\usepackage{helvet}  
\usepackage{courier}  
\usepackage[hyphens]{url}  
\usepackage{graphicx} 
\usepackage{natbib}  
\usepackage{caption} 
\usepackage{algorithm}
\usepackage{algorithmic}

\usepackage{newfloat}
\usepackage{listings}
\DeclareCaptionStyle{ruled}{labelfont=normalfont,labelsep=colon,strut=off} 
\floatstyle{ruled}
\newfloat{listing}{tb}{lst}{}
\floatname{listing}{Listing}
\usepackage{tikz}
\usepackage{caption}
\usepackage{subcaption}
\usepackage{amsmath} 
\usepackage{array} 
\usepackage{multirow}
\usepackage{amssymb}
\usepackage{booktabs}
\usepackage{adjustbox}

\newcommand*{\circled}[1]{\tikz[baseline=(char.base), scale=0.05]{
            \node[shape=circle,draw,inner sep=0.05pt] (char){\footnotesize #1};}}

\title{\emph{FreeInpaint}: Tuning-free Prompt Alignment and Visual Rationality Enhancement in Image Inpainting}
\author{
    Chao Gong\textsuperscript{\rm 1}\thanks{This work was performed at HiDream.ai.},
    Dong Li\textsuperscript{\rm 2},
    Yingwei Pan\textsuperscript{\rm 2},
    Jingjing Chen\textsuperscript{\rm 1,3}\thanks{Corresponding author.},
    Ting Yao\textsuperscript{\rm 2},
    Tao Mei\textsuperscript{\rm 2}
}
\affiliations{
    \textsuperscript{\rm 1}College of Computer Science and Artificial Intelligence, Fudan University \\
    \textsuperscript{\rm 2}HiDream.ai Inc.  \\
    \textsuperscript{\rm 3}Institute of Trustworthy Embodied AI, Fudan University

    cgong24@m.fudan.edu.cn, \{lidong, pandy\}@hidream.ai, chenjingjing@fudan.edu.cn, \{tiyao, tmei\}@hidream.ai
}

\usepackage{bibentry}

\begin{document}

\maketitle

\begin{abstract}
Text-guided image inpainting endeavors to generate new content within specified regions of images using textual prompts from users. The primary challenge is to accurately align the inpainted areas with the user-provided prompts while maintaining a high degree of visual fidelity. While existing inpainting methods have produced visually convincing results by leveraging the pre-trained text-to-image diffusion models, they still struggle to uphold both prompt alignment and visual rationality simultaneously. In this work, we introduce \emph{FreeInpaint}, a plug-and-play tuning-free approach that directly optimizes the diffusion latents on the fly during inference to improve the faithfulness of the generated images. Technically, we introduce a prior-guided noise optimization method that steers model attention towards valid inpainting regions by optimizing the initial noise. Furthermore, we meticulously design a composite guidance objective tailored specifically for the inpainting task. This objective efficiently directs the denoising process, enhancing prompt alignment and visual rationality by optimizing intermediate latents at each step. Through extensive experiments involving various inpainting diffusion models and evaluation metrics, we demonstrate the effectiveness and robustness of our proposed \emph{FreeInpaint}.
\end{abstract}

\begin{links}
    \link{Code}{https://github.com/CharlesGong12/FreeInpaint}
\end{links}

\section{Introduction}

Recent years have witnessed the success of large text-to-image (T2I) diffusion models \cite{rombach2022ldm,saharia2022imagen,cai2025hidream,yao2025denoising} and their exceptional ability to produce high-quality images. Despite the effectiveness of these generative models, users with specific design requirements often face the need for iterative interactions to achieve their desired outcomes. Text-guided image editing \cite{hertz2022p2p,brooks2023instructpix2pix,han2025unireditbench} stands out as a transformative approach that enriches the image generation process by facilitating interactive refinement. Particularly, text-guided image inpainting empowers users to create new content within specified regions of existing images, guided by textual prompts. This capability opens up various applications, including object removal \cite{suvorov2022resolution}, photo restoration \cite{lugmayr2022repaint}, and virtual try-on \cite{choi2024improving,wan2024improving,li2025pursuing}.

\begin{figure}[t]
    \centering
    \footnotesize 
    \setlength{\tabcolsep}{0.2pt} 
    \renewcommand{\arraystretch}{0.8} 

    \begin{tabular}{c p{0.14\linewidth} p{0.14\linewidth} p{0.14\linewidth} p{0.14\linewidth} p{0.14\linewidth} p{0.14\linewidth} p{0.14\linewidth}}
        & \multicolumn{6}{l}{\textit{Prompt: A black formal dress on the kitten.}} & \multicolumn{1}{c}{} \\

        & \includegraphics[width=\linewidth]{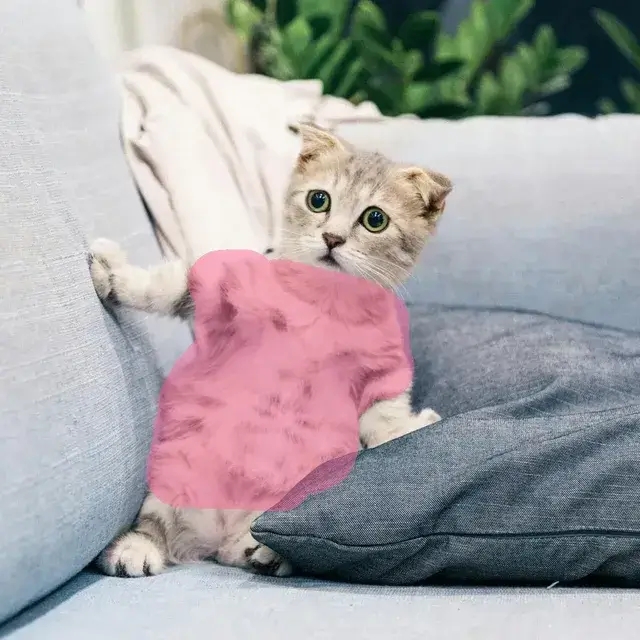} 
        & \includegraphics[width=\linewidth]{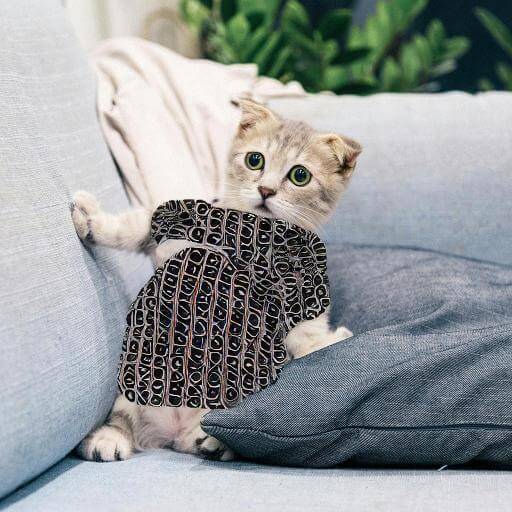} 
        & \includegraphics[width=\linewidth]{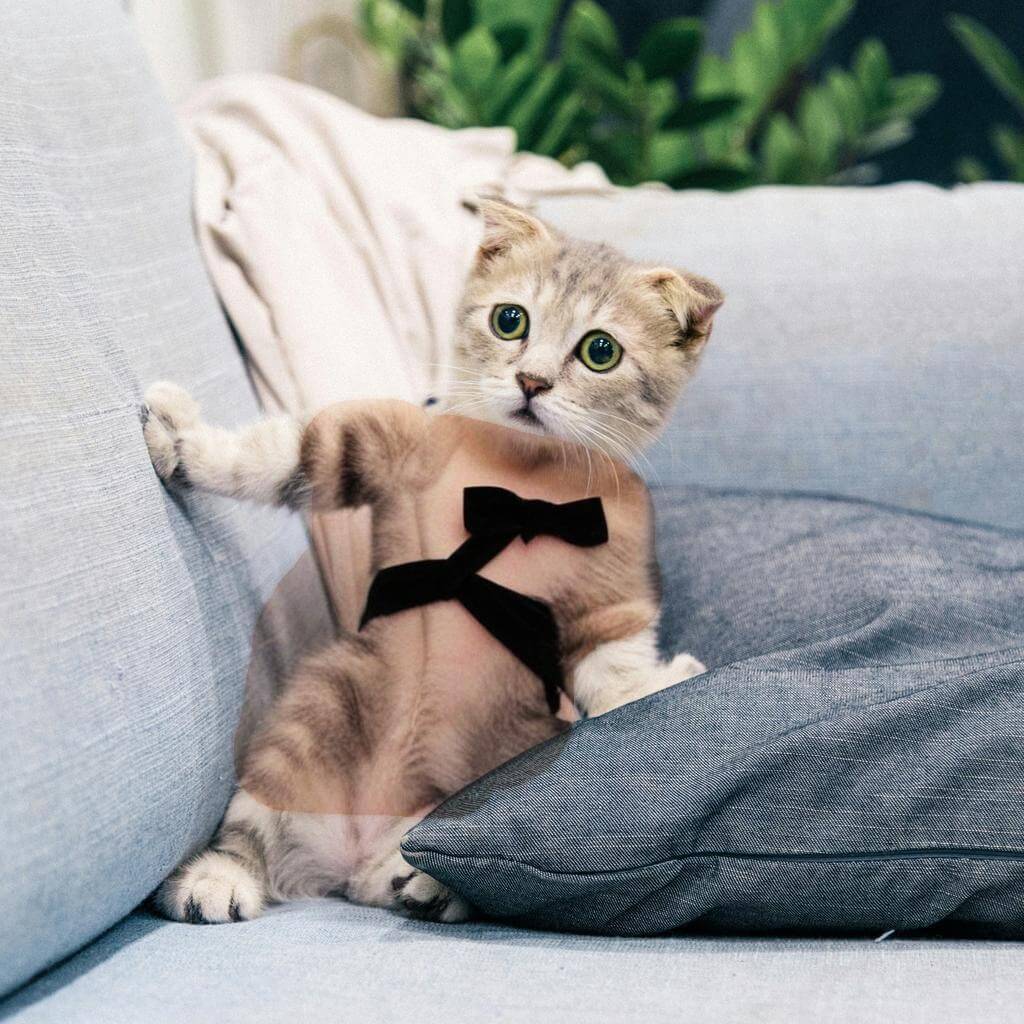}
        & \includegraphics[width=\linewidth]{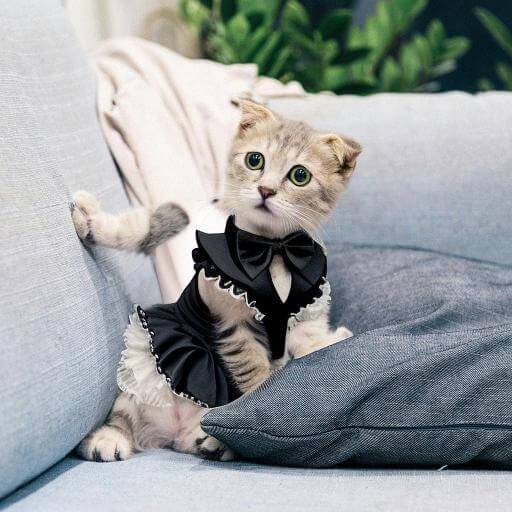} 
        & \includegraphics[width=\linewidth]{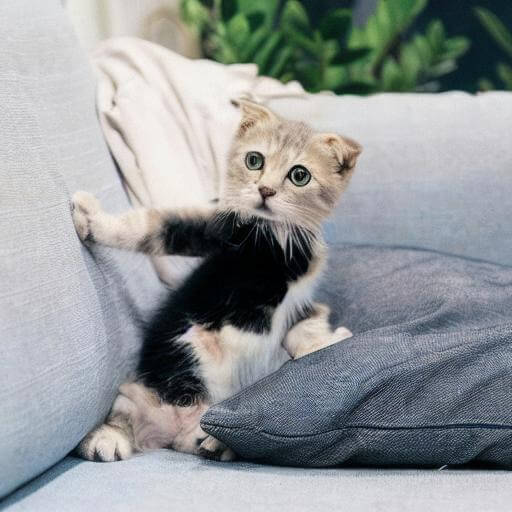} 
        & \includegraphics[width=\linewidth]{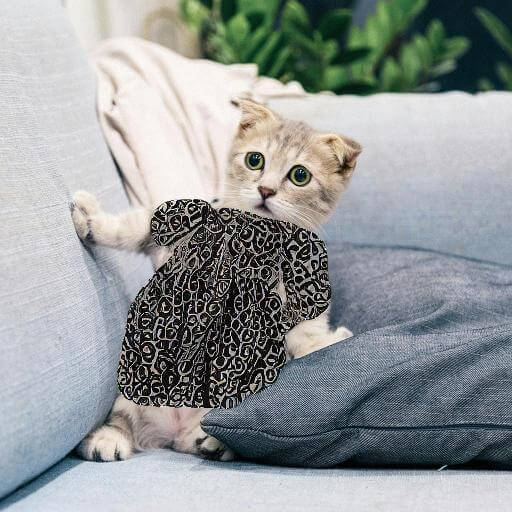} 
        & \includegraphics[width=\linewidth]{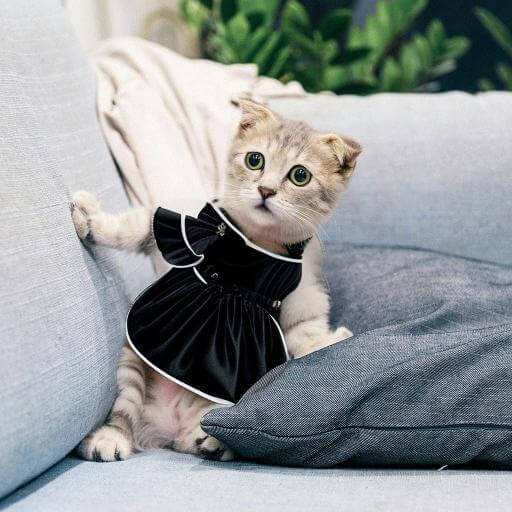} \\

        & \multicolumn{6}{l}{\textit{Prompt: A bouquet of colorful flowers in Hulk's hand.}} & \multicolumn{1}{c}{} \\

        & \includegraphics[width=\linewidth]{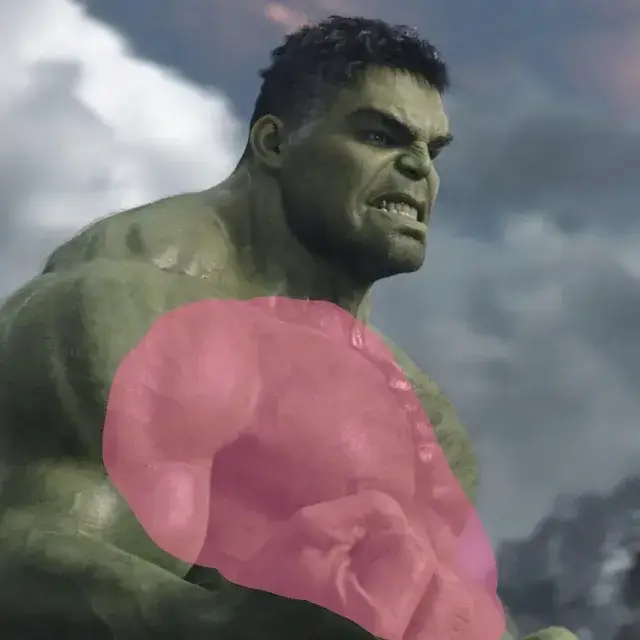} 
        & \includegraphics[width=\linewidth]{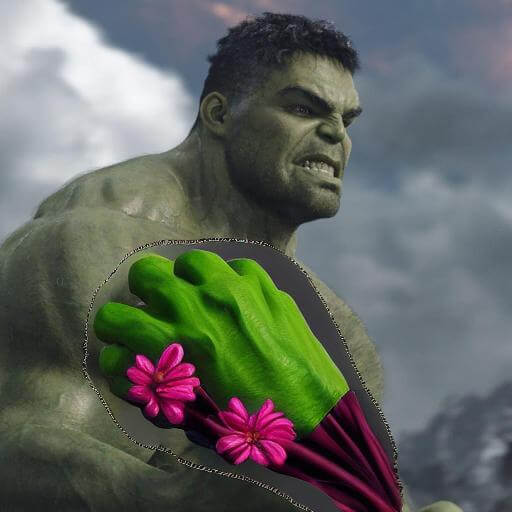} 
        & \includegraphics[width=\linewidth]{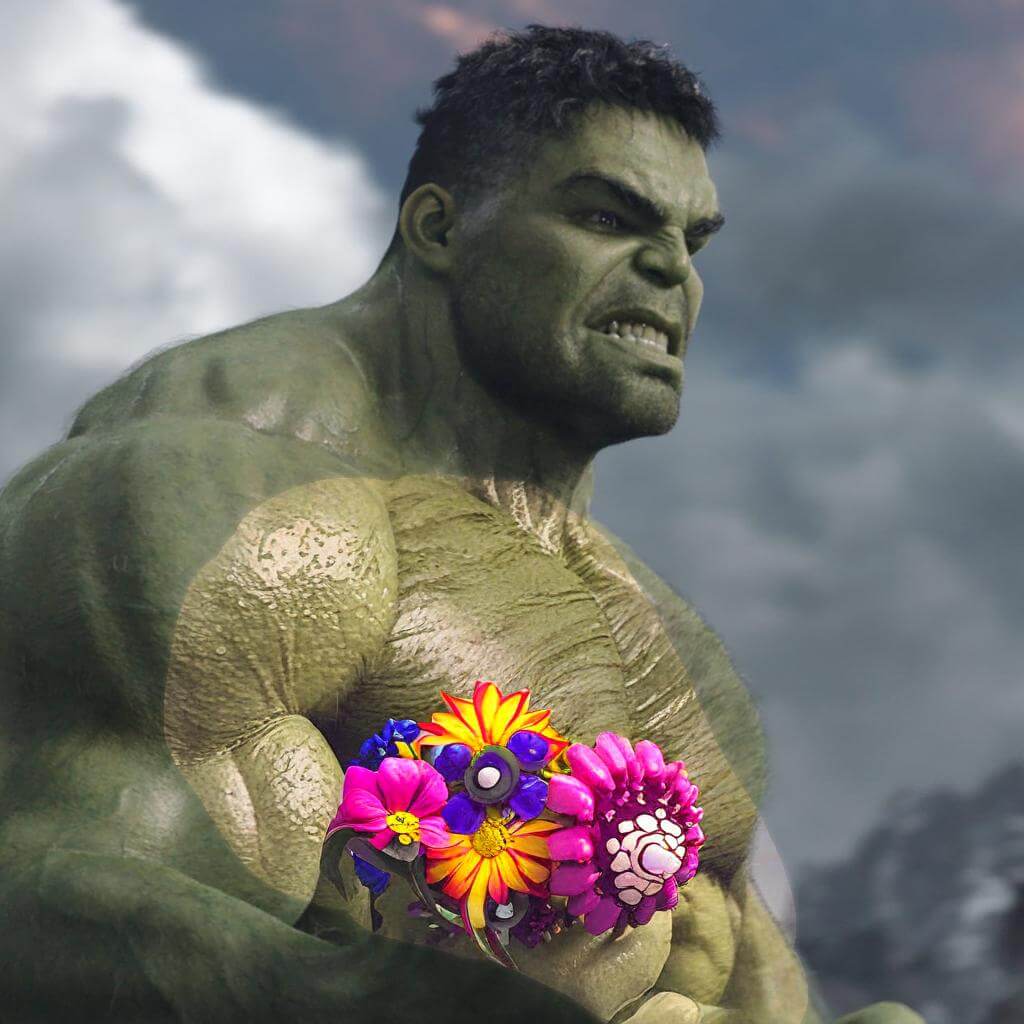}
        & \includegraphics[width=\linewidth]{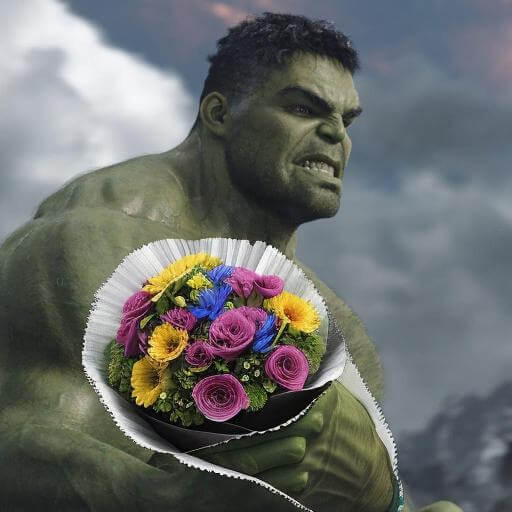} 
        & \includegraphics[width=\linewidth]{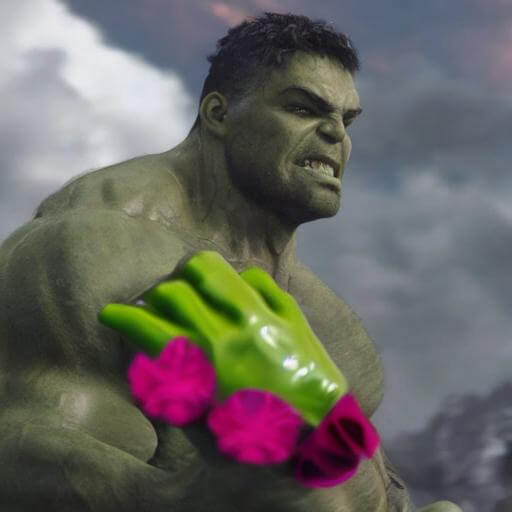} 
        & \includegraphics[width=\linewidth]{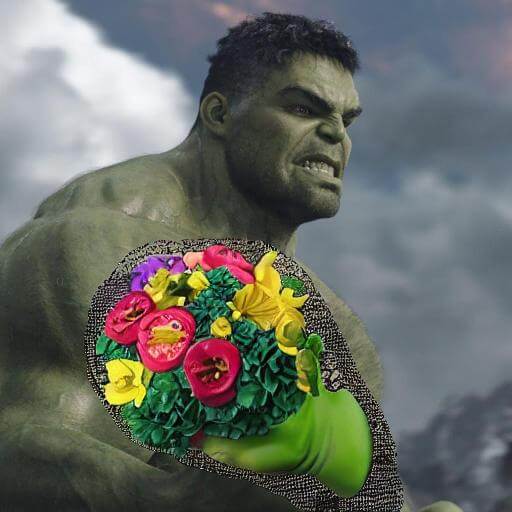} 
        & \includegraphics[width=\linewidth]{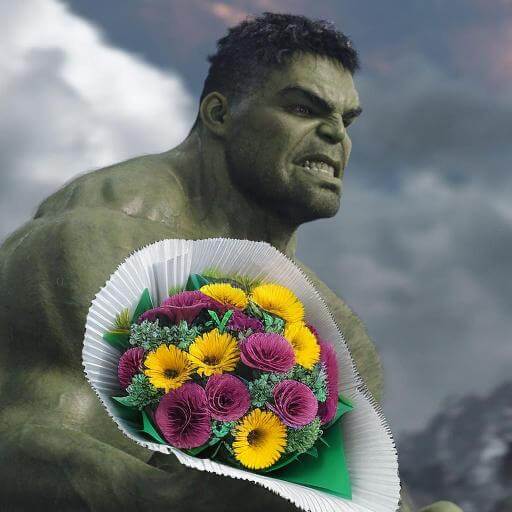} \\

        & \multicolumn{6}{l}{\textit{Prompt: A purple scarf around the bird's neck.}} & \multicolumn{1}{c}{} \\

        & \includegraphics[width=\linewidth]{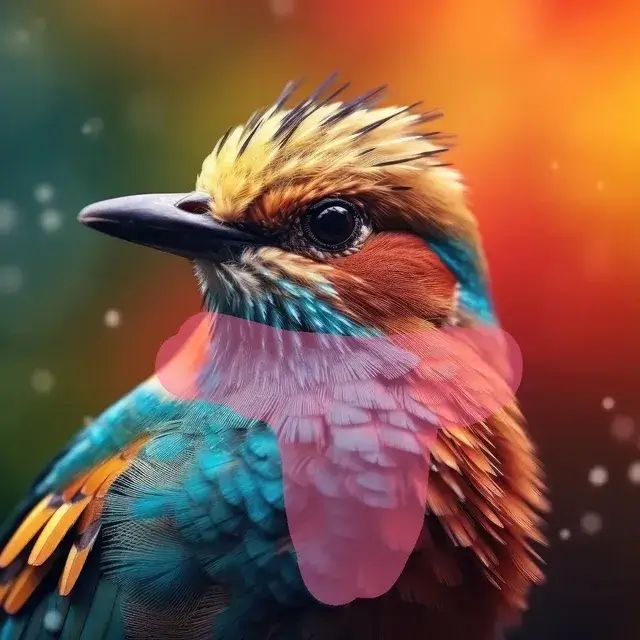} 
        & \includegraphics[width=\linewidth]{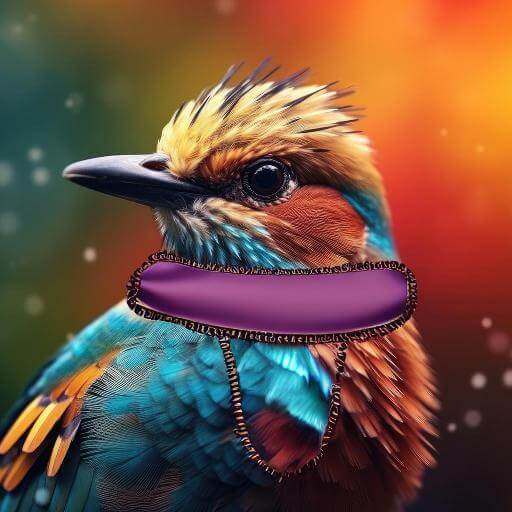} 
        & \includegraphics[width=\linewidth]{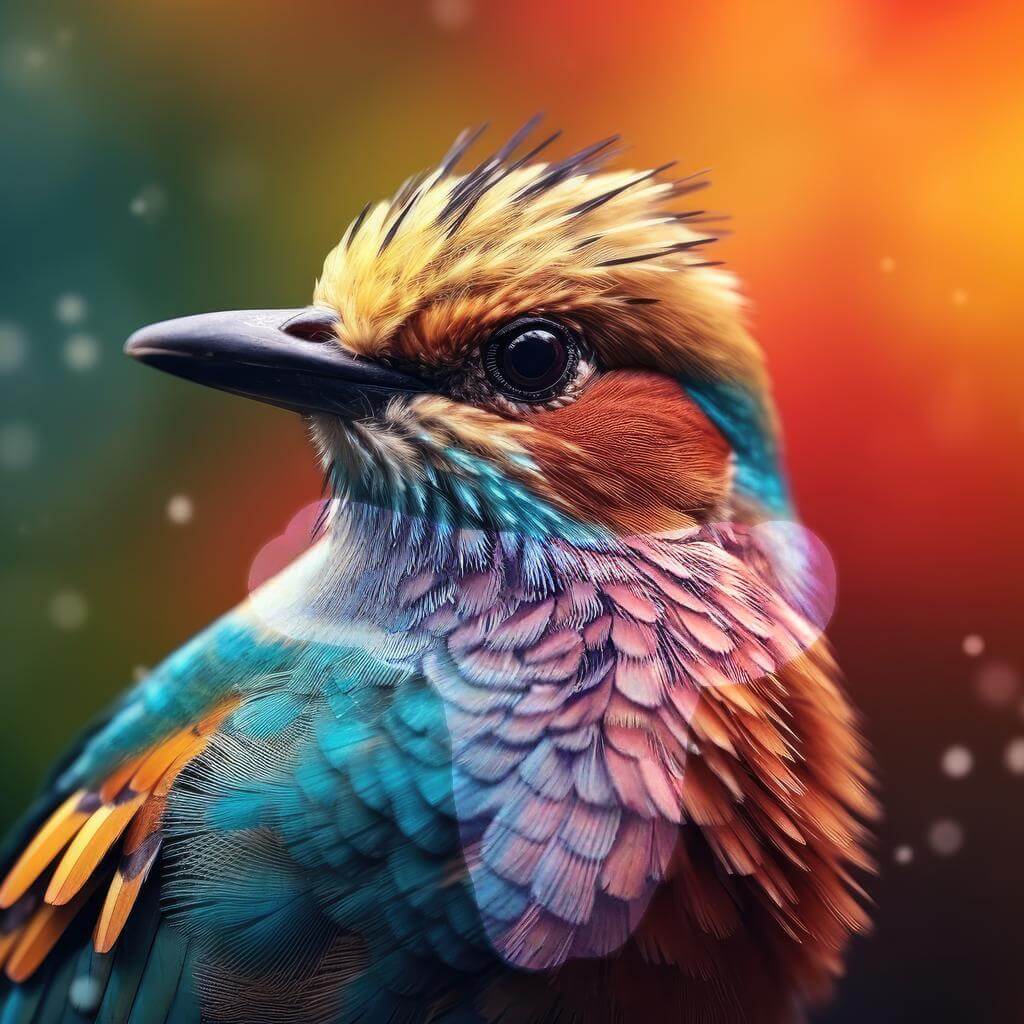}
        & \includegraphics[width=\linewidth]{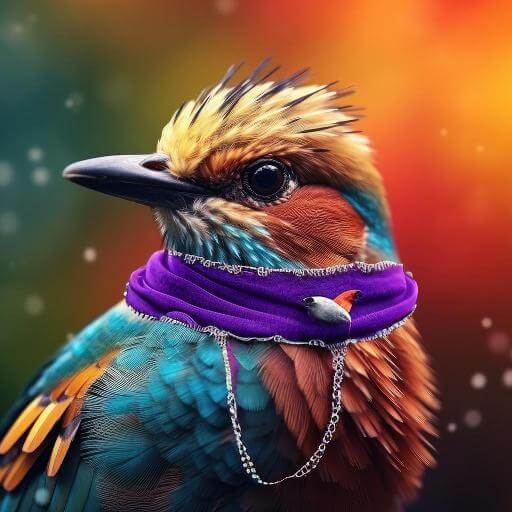} 
        & \includegraphics[width=\linewidth]{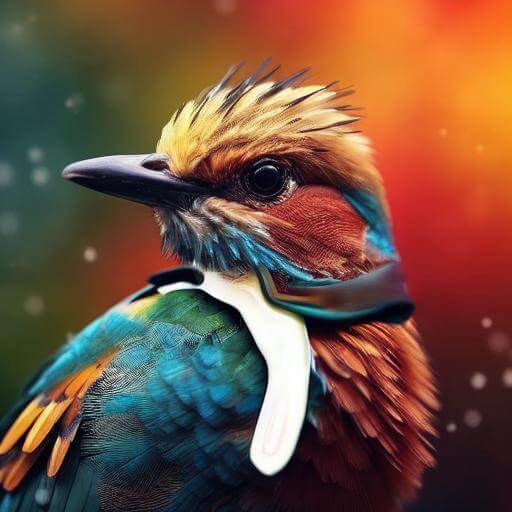} 
        & \includegraphics[width=\linewidth]{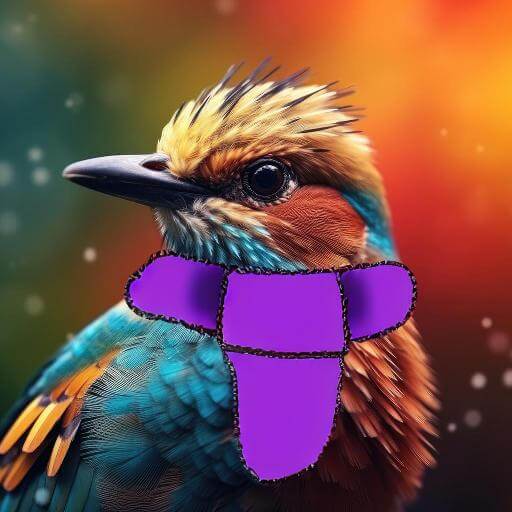} 
        & \includegraphics[width=\linewidth]{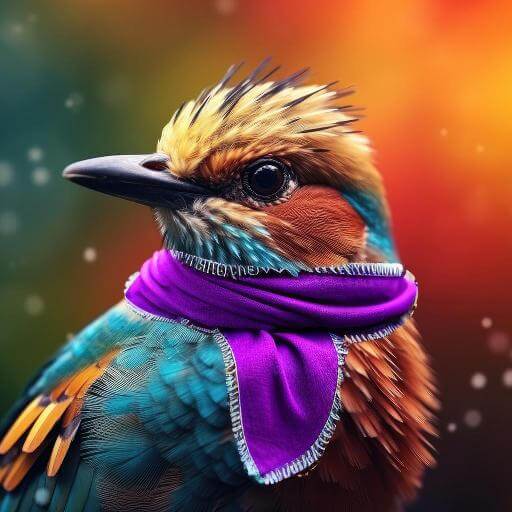} \\

        & \centering \scriptsize Input Image & \centering \scriptsize SD-Inpainting & \centering \scriptsize SDXL-Inpainting 
        & \centering \scriptsize BrushNet & \centering \scriptsize PowerPaint & \centering \scriptsize HD-Painter & \centering \scriptsize \emph{FreeInpaint} (Ours) \\
    \end{tabular}
  \caption{Comparisons between our \emph{FreeInpaint} and existing methods. \emph{FreeInpaint} simultaneously enhances prompt alignment and visual rationality. Zoom in for better view.}
  \label{fig:intro-comparisions}
\end{figure}

One straightforward approach to extending pre-trained T2I diffusion models for text-guided image inpainting is to blend diffused known regions with denoised unknown regions during the reverse diffusion process \cite{lugmayr2022repaint,avrahami2022blended}. However, this technique frequently produces incoherent inpainting results due to limited global scene comprehension and constrained perceptual understanding of mask boundaries. To tackle this issue, recent methods \cite{rombach2022ldm,nichol2021glide,podell2023sdxl} have shifted their focus towards finetuning T2I diffusion models by randomly removing portions of the input image and compelling the model to reconstruct the missing areas based on the corresponding caption. Despite reasonable performance, this training strategy leads to weak image-prompt alignment since randomly selected regions can often be feasibly inpainted purely based on image context, with limited consideration for the provided prompt. To improve prompt alignment, SmartBrush \cite{xie2023smartbrush}, ImagenEditor \cite{wang2023imageneditor}, and PowerPaint \cite{zhuang2024ppt} propose object-aware masking strategies and leverage object descriptions as prompts. Beyond that, BrushNet \cite{ju2024brushnet} employs a dual-branch architecture to reduce the training burden. While these methods alleviate prompt misalignment, the visual rationality of generated images has been compromised (See Fig.~\ref{fig:intro-comparisions}).

To achieve both prompt alignment and visual rationality concurrently, our investigation dives into the direct optimization of diffusion latents within the reverse diffusion process. Notably, we argue that two key ingredients should be taken into account during inference. One is \textbf{\textit{initial noise optimization}} at the beginning of the denoising process. Recent studies \cite{wallace2023doodl,samuel2024noiserare} highlight that pre-trained T2I diffusion models are sensitive to the initial random noise, with not all randomly sampled noise leading to visually consistent images. We reveal that this sensitivity extends to image inpainting models as well. As shown in Fig. \ref{fig:attention}, when different noise inputs are fed into the inpainting diffusion model with identical text prompts, a substantial inconsistency is observed in the image-prompt alignment. Therefore, we posit that a well-optimized initial noise can significantly enhance prompt alignment. The other is \textbf{\textit{task-specific guidance}} throughout the denoising process. While leveraging the generative capabilities of T2I models proves efficient across various downstream tasks, the generative data distributions derived from diverse training data often deviate from the specific requirements of individual tasks. Specifically, for the task of image inpainting, this discrepancy will degrade the visual rationality of generated images. Training-free guidance has proven effective in aligning generated images with task-specific objectives by utilizing differentiable rewards to optimize diffusion latents during inference \cite{dhariwal2021diffusiongan,chefer2023attend}. Nevertheless, an effective guidance objective tailored for the inpainting task remains underexplored.

By consolidating the idea of optimizing both initial noise and intermediate latents during the denoising process, we novelly present a plug-and-play tuning-free method, \emph{FreeInpaint}, to enhance prompt alignment and visual rationality in image inpainting. On the one hand, we introduce a Prior-Guided Noise Optimization method, dubbed \textit{PriNo}, to direct attention maps towards valid inpainting regions. Through contrastive analysis, we observe an obvious attention direction gap between prompt-aligned and unaligned inpainting results. Recognizing that object presence and low-frequency characteristics are predominantly established in the early stages of denoising \cite{yang2023freqslim,park2023freqriemannian}, we propose an attention loss to steer the attention maps of the first denoising step towards concentrating on the masked region, and then optimize the initial noise with respect to the gradients of the attention loss. This approach effectively directs the randomly sampled noise towards a direction that better aligns with the semantic context outlined in the prompt. On the other hand, we exquisitely craft a Decomposed Training-free Guidance objective, dubbed \textit{DeGu}, customized for the inpainting task to guide the denoising process. Specifically, we decompose the conditional distribution of the inpainting process into three distinct objectives: text alignment, visual rationality, and human preference. Such decomposition enables the seamless integration of off-the-shelf differentiable models to comprehensively enhance the inpainting model's perception and generation quality, without any training or fine-tuning requirements.

The main contribution of this work is the proposal of \emph{FreeInpaint}, a tuning-free approach that facilitates the text-guided image inpainting task by directly optimizing diffusion latents during inference. Our method elegantly explores how to optimize the initial noise with the prior knowledge tailored to image inpainting, and how to improve the denoising process with task-specific guidance. By conducting comprehensive experiments with a range of inpainting diffusion models on the EditBench and MSCOCO datasets, we demonstrate the effectiveness and robustness of \emph{FreeInpaint}.

\section{Related Work}
\paragraph{Image Inpainting}
Image inpainting \cite{bertalmio2000imageinpainting}, the task of filling missing regions in images, has advanced significantly with deep learning. While early methods employing Generative Adversarial Networks (GANs) \cite{goodfellow2020gan} can complete contexts, they often struggle to generate novel objects based on textual descriptions. A recent breakthrough comes with diffusion models \cite{dhariwal2021diffusiongan}. Early blending-based diffusion approaches \cite{avrahami2022blended} blend the edited region with the unchanged parts of the image at different noise levels. However, the edited areas often exhibit visible artifacts and lack global consistency with the overall scene. To address this limitation, current research is predominantly focused on fine-tuning pre-trained diffusion U-Nets. For example, Stable Diffusion Inpainting \cite{rombach2022ldm} refines a pre-trained U-Net by concatenating the noise latent, mask, and known image latent as input. ControlNet \cite{zhang2023controlnet} and BrushNet \cite{ju2024brushnet} use dedicated branches to diminish the training load. Despite these advancements, these methods typically rely on randomly scribbled or prompt-irrelevant masks during training, leading to potential neglect of the provided prompt. To enhance prompt alignment, subsequent methods have utilized paired mask-description data \cite{wang2023imageneditor,xie2023smartbrush} or introduced learnable prompts for specific scenarios \cite{zhuang2024ppt}. In contrast, HD-Painter \cite{manukyan2023hd} adopts a training-free approach. It reweights the self-attention scores according to the alignment of the pixels with the prompt. Nevertheless, its singular focus on prompt alignment often compromises visual rationality, resulting in a decline in overall quality. Our \emph{FreeInpaint} resolves this trade-off by decoupling the objectives: it prioritizes prompt alignment through optimizing initial noise before denoising, and then guides each denoising step to balance between prompt alignment and visual coherence. This approach yields significant improvements in both prompt adherence and overall quality.

\paragraph{Initial Noise Optimization}
Initial noise optimization refines the initial noise latent to enhance the quality of images generated by diffusion models. First introduced by DOODL \cite{wallace2023doodl} to improve guidance in T2I models, this technique has since been applied to diverse tasks like rare-concept generation \cite{samuel2024noiserare}, improving synthesis fidelity \cite{guo2024initno, 2025selfcross}, creating 3D motion priors \cite{karunratanakul2024noise3d}, audio generation \cite{novack2024dittomusic}, and solving inverse problems \cite{ben2024dflow}. Other works have focused on improving its efficiency \cite{samuel2024noiserare} or using reward-based optimization \cite{eyring2024reno}. Despite these advancements, how to optimize the initial noise specifically for the inpainting task has not yet been studied.

\paragraph{Post-Training Guidance}
In the T2I task, post-training guidance manipulates the denoising process of pre-trained models to steer generation towards specific objectives without retraining. This area is pioneered by Classifier Guidance \cite{dhariwal2021diffusiongan} and later advanced by CLIP guidance \cite{nichol2021glide} and classifier-free guidance \cite{ho2022cfg} for improved text-guided sample quality. Subsequent works leverage attention mechanisms to influence object attributes \cite{chefer2023attend, epstein2023selfguidance, xie2024dymo} under complex prompts. However, directly applying T2I guidance to image inpainting is ineffective. Unlike the T2I task, which primarily aligns the entire image with the prompt, image inpainting concentrates on the relevance between the masked local area and the prompt, alongside ensuring consistency between the masked and non-masked regions in the image. Therefore, crafting a tailored guidance objective for image inpainting is essential.

\section{Preliminaries}
\paragraph{Stable Diffusion Model}
Our work builds on the Stable Diffusion (SD) model \cite{rombach2022ldm}, which handles the T2I task in a latent space managed by a variational autoencoder (VAE) \cite{kingma2013vae}. The VAE's encoder $\mathcal{E}(\cdot)$ maps an image $I \in \mathbb{R}^{h\times w \times 3}$ to a latent $z_0$, and its decoder $\mathcal{D}(\cdot)$ reconstructs the image from the latent. SD's forward diffusion process \cite{ho2020ddpm} gradually adds Gaussian noise to $z_0$ over timesteps $t\in [0,T]$:
\begin{equation}
z_t = \sqrt{\bar{\alpha}_t}z_0 + \sqrt{1-\bar{\alpha}_t}\epsilon_t, 
\end{equation}
where $\epsilon_t\sim\mathcal{N}(\mathbf{0},\mathbf{1})$ is noise and $\bar{\alpha}_t$ is a noise schedule that decreases from $1$ to $0$ as $t$ increases. The reverse process uses a denoising model, $\epsilon_\theta$, to predict and remove the noise at each step, trained with the objective:
\begin{equation}
\mathcal{L} = \mathbb{E}_{z_t, t, \epsilon_t\sim\mathcal{N}(\mathbf{0},\mathbf{1})} \|\epsilon_t - \epsilon_\theta(z_t, t)\|_2^2.
\end{equation}

For text-guided inpainting, the model is fine-tuned by conditioning on a text prompt embedding $c$, a mask $M \in \mathbb{R}^{h\times w}$, and the masked image latent $z^m = \mathcal{E}(I\odot(1-M))$ to redraw the specified region. It estimates the noise $\epsilon_\theta(z_t, t, c, z^m, M^\prime)$, where $M^\prime$ is a downsampled mask matching the latent's dimensions. These conditions are typically injected either by concatenation with the latent $z_t$ \cite{rombach2022ldm,zhuang2024ppt} or through a dedicated dual-branch architecture \cite{ju2024brushnet}.

\paragraph{Diffusion Guidance} \label{sec:diffusion-guidance}
Conditional diffusion is often steered by guidance. In score-based diffusion models \cite{song2020scoresde}, the score function is directly related to the noise $\epsilon_t$:
\begin{equation}
\nabla_{z_{t}} \log p(z_{t}) = - \frac{1}{\sqrt{1-\bar{\alpha}_t}}\epsilon_t.
\label{eq:score_function}
\end{equation}
To introduce a condition $c$, Bayes' Theorem \cite{josang2016bayes} is used to derive a conditional score function:
\begin{equation}
\begin{aligned}
\nabla_{z_t}\log p(z_t|c) &\propto \nabla_{z_t}\log p(z_t) + \nabla_{z_t}\log p(c|z_t) \\
&= - \frac{1}{\sqrt{1-\bar{\alpha}_t}}\epsilon_t + \nabla_{z_t}\log p(c|z_t).
\end{aligned}
\end{equation}
This leads to a modified noise prediction with a guidance term \cite{dhariwal2021diffusiongan}:
\begin{equation}
\hat{\epsilon}_t = \epsilon_\theta(z_t, t) - \sqrt{1-\bar{\alpha}_t} \nabla_{z_t}\log p(c|z_t),
\end{equation}
where the guidance term $\nabla_{z_t}\log p(c|z_t)$ acts as a corrective gradient from a conditioning model (e.g., a classifier) to guide the output towards satisfying condition $c$.

\begin{figure*}[t]
  \includegraphics[width=\textwidth]{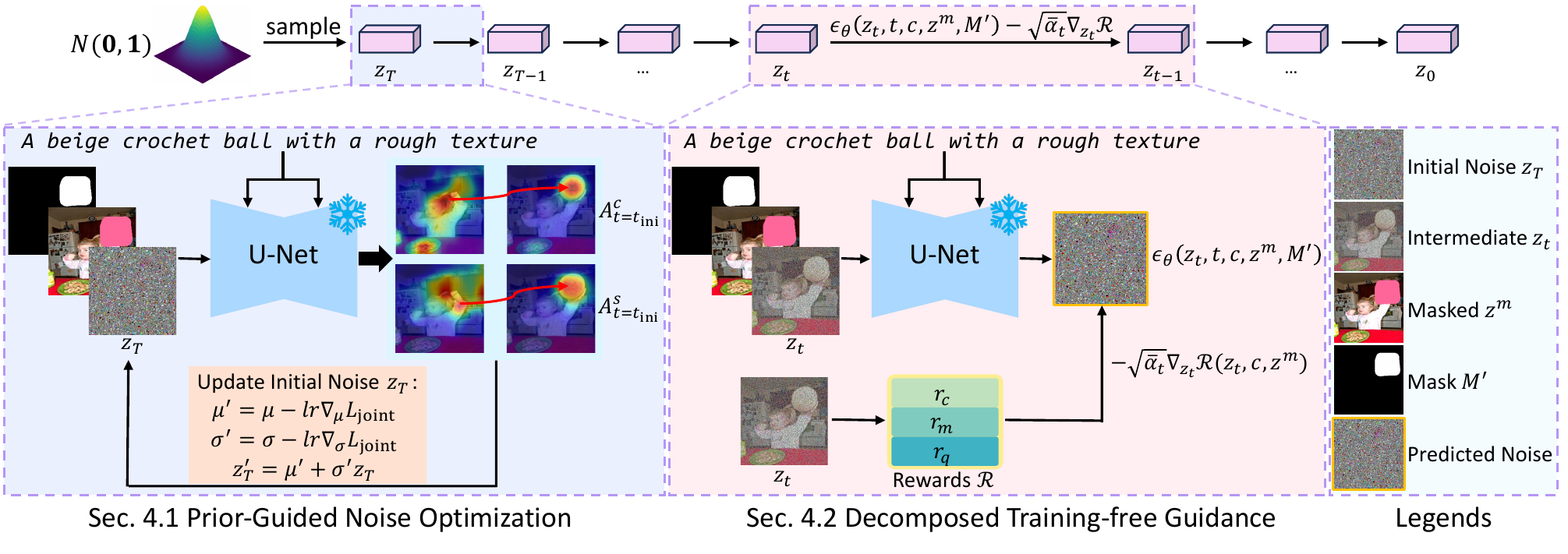}
  \caption{Our tuning-free \emph{FreeInpaint} framework consists of two key stages: (1) Prior-Guided Noise Optimization (Sec. \ref{sec:NoiseOpt}), which optimizes the initial noise $z_T$ at the first denoising step $t=t_\text{ini}$ to concentrate cross-attention ($A^c_{t=t_\text{ini}}$) and self-attention ($A^s_{t=t_\text{ini}}$) maps within the masked region, improving prompt alignment; and (2) Decomposed Training-Free Guidance (Sec. \ref{sec:guidance}), which decomposes the conditional distribution of inpainting and leverages text alignment $r_c$, visual rationality $r_m$, and human preference $r_q$ reward models to guide the predicted noise $\epsilon_\theta(z_t, t, c, z^m, M^\prime)$ during denoising, enhancing visual rationality.}
  \label{fig:pipeline}
\end{figure*}

\section{Method}

We propose \emph{FreeInpaint}, an innovative training-free framework for image inpainting, as depicted in Fig. \ref{fig:pipeline}. Our framework comprises two key components: Prior-Guided Noise Optimization (PriNo, Sec. \ref{sec:NoiseOpt}) and Decomposed Training-Free Guidance (DeGu, Sec. \ref{sec:guidance}). Initially, prior to the denoising process, we leverage prior knowledge of image inpainting to optimize the initial noise latent, enhancing prompt alignment. Subsequently, during denoising, we decompose the inpainting process into three conditional distributions and apply specific guidance strategies to each, leading to improved visual rationality and human preference.

\subsection{Prior-Guided Noise Optimization} \label{sec:NoiseOpt}

A primary limitation of existing inpainting methods is the failure to generate objects specified in the text prompt, a problem we denote as \textit{prompt-unaligned}. Since a model's output is highly sensitive to the initial random noise \cite{wallace2023doodl,samuel2024noiserare,guo2024initno}, we propose \textbf{PriNo}, a novel Prior-Guided Noise Optimization strategy leveraging the attention mechanisms within inpainting models. PriNo optimizes the initial noise latent $z_T$ to steer the model's attention, promoting the generation of the desired object during the early, structure-defining stages of denoising \cite{yang2023freqslim,yi2024freqtowards,park2023freqriemannian}.

\subsubsection{Misdirected Attention}
\begin{figure}[t]
    \centering
    \footnotesize 
    \setlength{\tabcolsep}{0.5pt} 
    \renewcommand{\arraystretch}{0.8} 

    \begin{tabular}{c p{0.16\linewidth} p{0.16\linewidth} p{0.16\linewidth} p{0.16\linewidth} p{0.16\linewidth} p{0.16\linewidth}}
        & \multicolumn{6}{l}{\scriptsize Prompt-Aligned - \textit{Prompt: A yellow bird is diving towards the coral}}\\

        & \includegraphics[width=\linewidth]{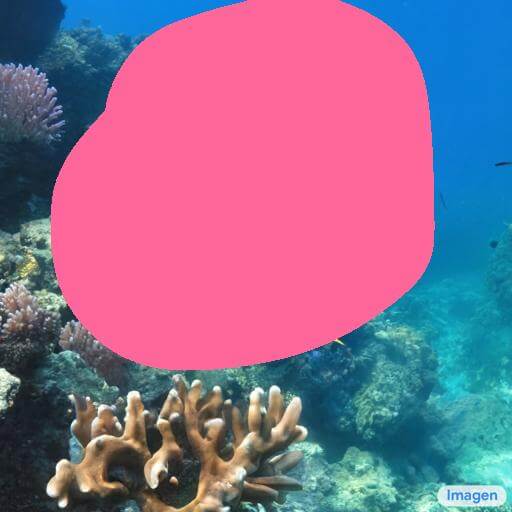} 
        & \includegraphics[width=\linewidth]{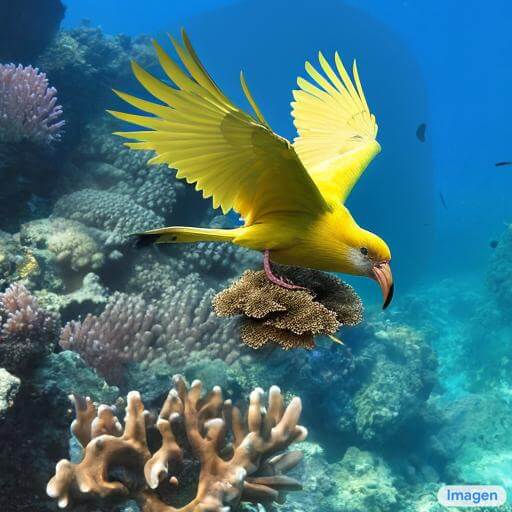} 
        & \includegraphics[width=\linewidth]{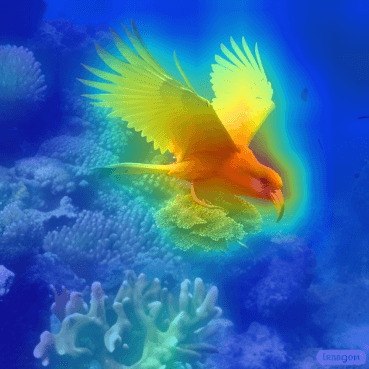} 
        & \includegraphics[width=\linewidth]{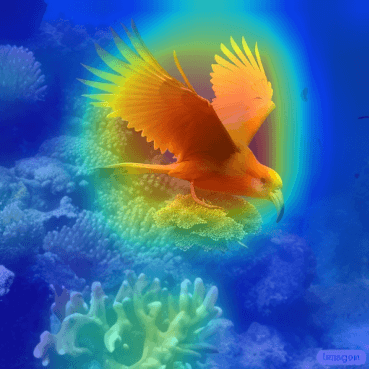} 
        & \includegraphics[width=\linewidth]{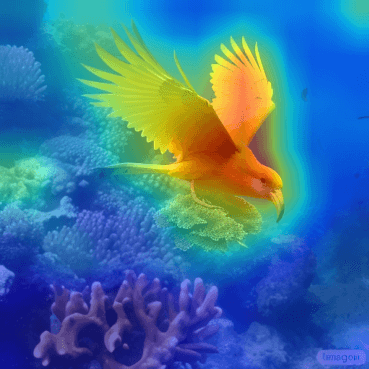} 
        & \includegraphics[width=\linewidth]{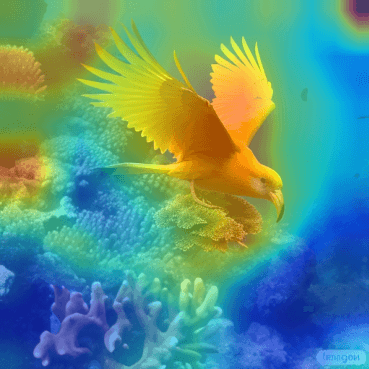} \\

        & \multicolumn{6}{l}{\scriptsize Prompt-Unaligned - \textit{Prompt: A woman facing away in a light-colored shirt and pants}} \\

        & \includegraphics[width=\linewidth]{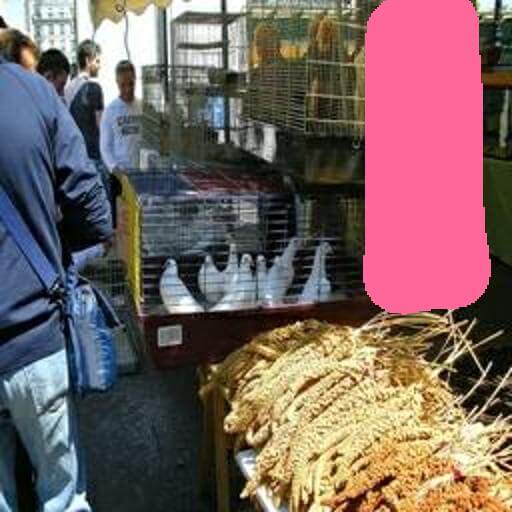} 
        & \includegraphics[width=\linewidth]{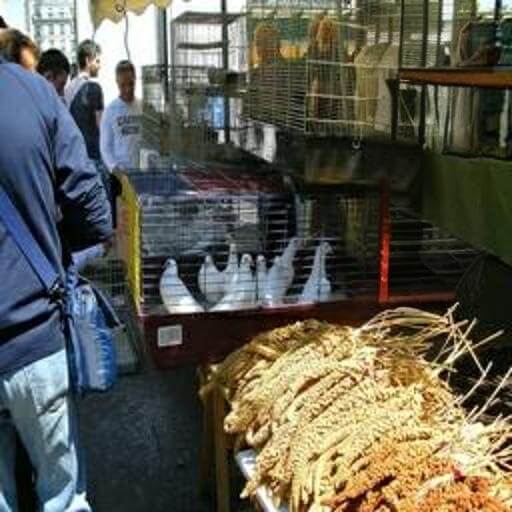} 
        & \includegraphics[width=\linewidth]{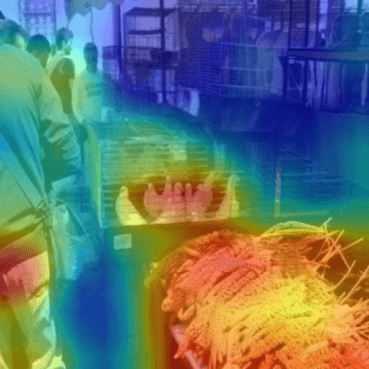} 
        & \includegraphics[width=\linewidth]{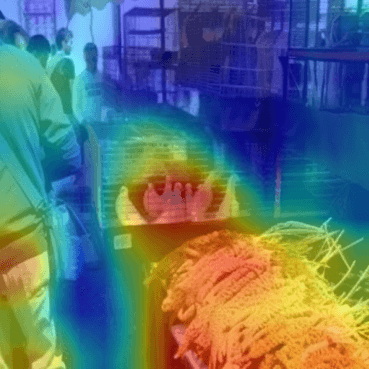} 
        & \includegraphics[width=\linewidth]{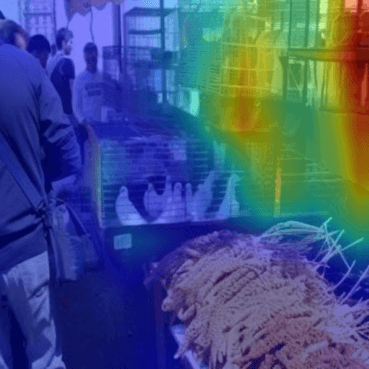} 
        & \includegraphics[width=\linewidth]{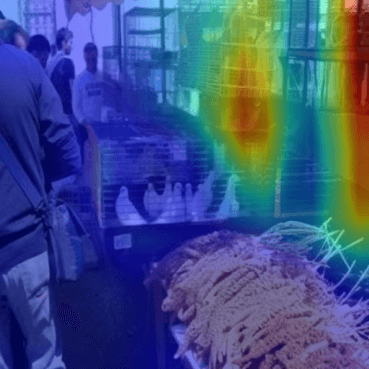} \\

        & \multicolumn{6}{l}{\scriptsize Prompt-Unaligned - w/ optimized initial noise}\\

        & \includegraphics[width=\linewidth]{images/attention/32_init.jpg} 
        & \includegraphics[width=\linewidth]{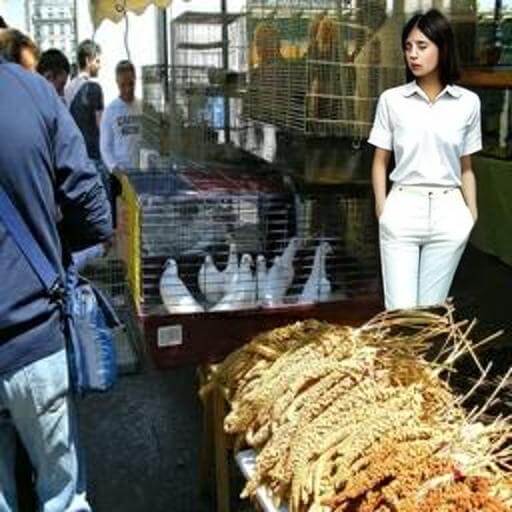} 
        & \includegraphics[width=\linewidth]{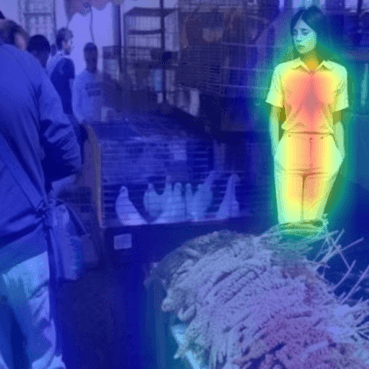} 
        & \includegraphics[width=\linewidth]{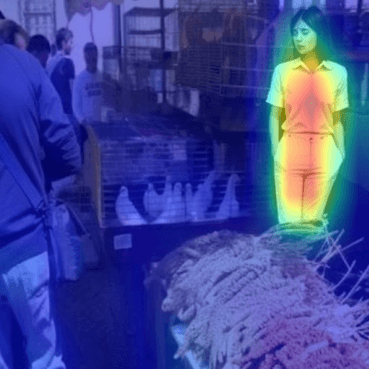} 
        & \includegraphics[width=\linewidth]{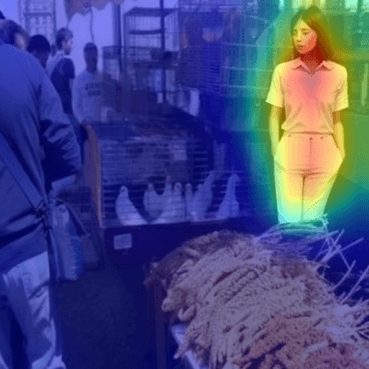} 
        & \includegraphics[width=\linewidth]{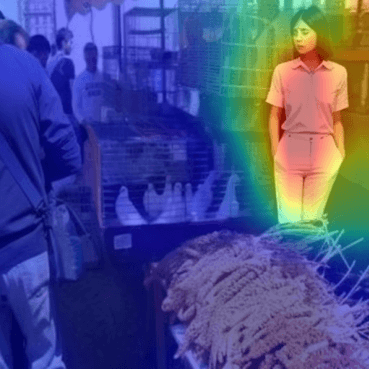} \\

        & \centering \scriptsize Masked Image & \centering \scriptsize Output Image & \centering $\bar{A}^c$ 
        & \centering $A^c_{t=t_\text{ini}}$ & \centering $\bar{A}^s$ & \centering $A^s_{t=t_\text{ini}}$ \\
    \end{tabular}

  \caption{Visualization of cross-attention (cols 3-4) and self-attention (cols 5-6). Row 2 shows misdirected attention causing an unaligned result, while row 3 shows our optimized noise concentrates attention for a successful alignment. The similarity between the first-step ($t=t_\text{ini}$) and averaged maps validates our efficient optimization.}
  \label{fig:attention}
\end{figure}

Our approach is motivated by an analysis of the attention mechanism \cite{vaswani2017attention} in inpainting diffusion models. In SD, the attention map is computed as $A = \mathrm{softmax}\left( QK^T / \sqrt{d} \right)$, where $d$ is the feature dimension. The query $Q$ is a projection of the visual features, while the key $K$ is a projection of either the prompt embedding $c$ (for cross-attention) or the visual features themselves (for self-attention). Thus, cross-attention maps $A^c$ indicate relevance between text tokens and visual patches, while self-attention maps $A^s$ indicate relevance between different visual patches.

To visualize attention maps across different inpainting results, we aggregate a cross-attention map $A^c$ by averaging them across all text tokens at the $h^\prime \times w^\prime$ spatial resolution of the model's feature maps. Similarly, a self-attention map $A^s$ is derived by averaging across all visual patches within the masked region. Columns 3 and 5 in Fig. \ref{fig:attention} display the attention maps using BrushNet \cite{ju2024brushnet}, averaged across all timesteps ($\bar{A}^c$ and $\bar{A}^s$), for a prompt-aligned and unaligned inpainting example. We observe that prompt-aligned inpainting results (row 1) exhibit highly concentrated attention within the mask, whereas unaligned instances (row 2) suffer from misdirected attention that fails to concentrate on the masked region. We hypothesize that this is a side effect of training on global image captions with randomly scribbled masks \cite{wang2023imageneditor, xie2023smartbrush}, which teaches the model to prioritize background context. 

\subsubsection{Noise Optimization for Attention Steering}
To counteract attention misdirection, we propose an attention-steering objective: \textbf{minimize attention outside the mask while maximizing it inside}. This principle is applied distinctly to each attention type to resolve their specific failure modes:
\begin{itemize}
    \item Cross-attention $A^c$ measures text-visual patch relevance. Misdirected $A^c$ incorrectly associates prompt content with the background, failing to generate the desired object within the mask. We correct this by forcing relevance between the prompt and visual patches inside the mask, ensuring prompt concepts manifest in the correct place.
    \item Self-attention $A^s$ measures inter-patch visual relevance. Misdirected $A^s$ causes the inpainted region to be over-influenced by the surrounding context. Our objective mitigates this, guiding the new content to focus on its own structure, not the background.
\end{itemize}
To implement this objective, we design two corresponding loss functions. The cross-attention loss encourages relevance between the prompt and the masked region:
\begin{equation}
\mathcal{L}_\text{c} = \sum_{i=1}^{h^\prime} \sum_{j=1}^{w^\prime} \left[ (1-M^\prime_{ij}) \cdot A^c_{ij} - M^\prime_{ij} \cdot A^c_{ij} \right],
\label{eq:loss_cross}
\end{equation}
where $M^\prime \in \mathbb{R}^{h^\prime \times w^\prime}$ is the downsampled input mask. Similarly, the self-attention loss encourages the process to focus on the masked region:
\begin{equation}
\mathcal{L}_\text{s} = \sum_{i=1}^{h^\prime} \sum_{j=1}^{w^\prime} \left[ (1-M^\prime_{ij}) \cdot A^s_{ij} - M^\prime_{ij} \cdot A^s_{ij} \right].
    \label{eq:loss_self}
\end{equation}

Optimizing the initial noise latent using the attention map averaged across all denoising steps is computationally prohibitive. However, we observe that attention maps at the initial denoising step $t=t_\text{ini}$ closely resemble the full average attention maps, as depicted in cols. 4 and 6 of Fig. \ref{fig:attention}. Thus, we compute $\mathcal{L}_\text{c}$ and $\mathcal{L}_\text{s}$ only at this first step.

We employ an effective update mechanism proposed by \cite{guo2024initno}, which tunes the mean $\mu$ and standard deviation $\sigma$ of the noise distribution $z_T \sim \mathcal{N}(\mu, \sigma^2)$, initialized with $\mu=\mathbf{0}$ and $\sigma=\mathbf{1}$. Our full objective is shown as:
\begin{equation}
    \mathcal{L}_\text{joint} = \lambda_{1}\mathcal{L}_\text{c} + \lambda_{2}\mathcal{L}_\text{s} + \lambda_{3}\mathcal{L}_\text{KL},
\label{eq:loss_joint}
\end{equation}
where $\lambda_{1}$, $\lambda_{2}$, and $\lambda_{3}$ are weighting hyperparameters, and $\mathcal{L}_\text{KL}$ is a KL divergence \cite{kullback195kl} to constrain the optimized distribution to remain close to $\mathcal{N}(\mathbf{0},\mathbf{1})$. We iteratively update $\mu$ and $\sigma$ with a learning rate $lr$ for $\tau_{\text{iter}}$ iterations to minimize $\mathcal{L}_\text{joint}$, then sample a new noise latent $z_T^\prime = \mu^\prime + \sigma^\prime z_T$ for the inpainting process. The complete procedure is detailed in the pseudocode in Appendix A. The steered attention maps and refined inpainting results are illustrated in row 3 of Fig. \ref{fig:attention}.

\subsection{Decomposed Training-free Guidance} \label{sec:guidance}
\begin{figure}[t]
    \centering
    \footnotesize
    \setlength{\tabcolsep}{0.5pt}

    \begin{tabular}{c p{0.2\linewidth} p{0.2\linewidth} p{0.2\linewidth} p{0.2\linewidth}}
        & \multicolumn{4}{l}{\textit{Prompt: The letters "water" on a brown paper box.}}\\
        & \includegraphics[width=\linewidth]{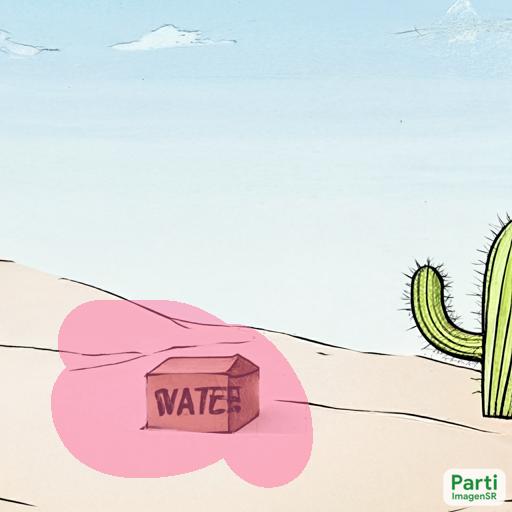} 
        & \includegraphics[width=\linewidth]{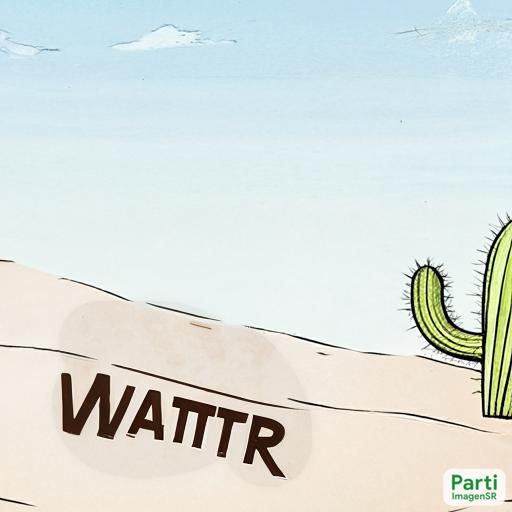} 
        & \includegraphics[width=\linewidth]{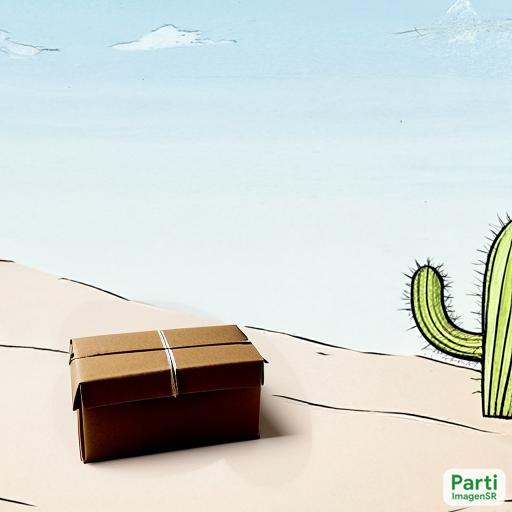}
        & \includegraphics[width=\linewidth]{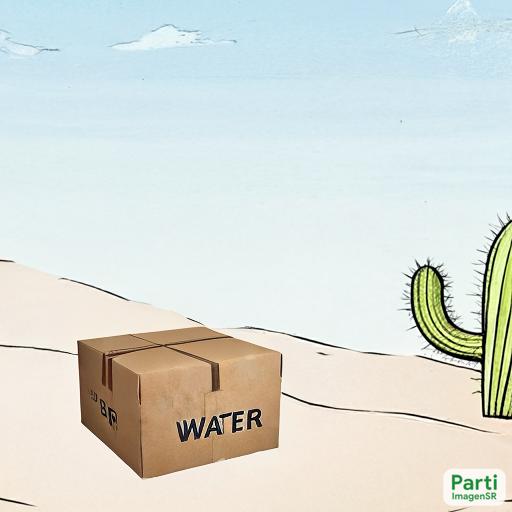}\\
        & \centering \scriptsize Input & \centering \scriptsize BrushNet & \centering \scriptsize BrushNet + PriNo & \centering \scriptsize BrushNet + PriNo + DeGu \\
    \end{tabular}
    \caption{Sec. \ref{sec:NoiseOpt} PriNo improves prompt alignment and Sec. \ref{sec:guidance} DeGu further refines image details.}

    \label{fig:role-of-guidance}
\end{figure}

While our PriNo strategy improves prompt alignment, it can leave finer details suboptimal. To enhance image detail and visual rationality, we introduce \textbf{DeGu}, a Decomposed Training-free Guidance approach. DeGu refines results by decomposing the inpainting process into three explicit objectives, each guided by an off-the-shelf reward model.

\subsubsection{Decomposing Inpainting Process}
As illustrated in Fig. \ref{fig:role-of-guidance}, even when the main object "paper box" is generated correctly via PriNo, finer details (e.g., the drawn letter) and image quality (e.g., brightness) can be lacking. To address these limitations in detail and quality, we revisit the inpainting process from the perspective of conditional probability. Then we decompose the inpainting process to tailor a fine-grained guidance mechanism that targets text alignment, visual rationality, and human preference.

Existing guidance mechanisms lack fine-grained control specifically tailored for the nuances of image inpainting \cite{dhariwal2021diffusiongan,kdiffusion,glid3xl,xie2024dymo}. We address this by formulating the inpainting process as modeling the conditional probability $p(z_{t} | c, z^m)$, and decompose it to explicitly address the key objectives for high-quality inpainting:
\begin{equation}
\begin{aligned}
    p(z_{t} | c, z^m)
    & \overset{\circled{1}}{\propto} p(c, z^m | z_{t}) \cdot p(z_{t}) \\
    & \overset{\circled{2}}{=} p(c | z_{t}) \cdot p(z^m | z_{t}) \cdot p(z_{t}).
\end{aligned}
\end{equation}
Here, proportionality \circled{1} follows from Bayes' Theorem \cite{josang2016bayes}, while equality \circled{2} assumes conditional independence between the text $c$ and masked latent $z^m$ \cite{Allan2013probability}. To further enhance human preference, we consider a human preference condition $q$, with $p(q|z_{t}) \propto \exp(r_{\text{q}}(z_{t}))$ based on a human preference reward function $r_{\text{q}}$ \cite{song2021ebm}. This allows us to augment the posterior $p(z_{t} | c, z^m)$ with condition $q$ for a more refined distribution:

\begin{equation}
\begin{aligned}
p(z_{t} | c, z^m, q)\propto p(c | z_{t}) \cdot p(z^m | z_{t}) \cdot p(q | z_{t}) \cdot p(z_{t}).
\end{aligned}
\end{equation}
Instead of treating text-guided inpainting as a monolithic process, we explicitly decompose this process into three distinct and guided objectives: (1) \textit{text alignment} $p(c | z_{t})$, enforcing alignment of the generated content to the text prompt, (2) \textit{visual rationality} $p(z^m | z_{t})$, ensuring coherence between the generated and known regions, and (3) \textit{human preference} $p(q|z_{t})$, promoting the overall aesthetic appeal and plausibility of the generated image. 

\subsubsection{Reward-Guided Denoising}
Building upon the guidance techniques in Sec. \ref{sec:diffusion-guidance}, we achieve multi-conditional guidance by employing a weighted combination of score functions \cite{song2020scoresde} during denoising:
\begin{equation}
\begin{split}
    & \nabla_{z_{t}} \log p(z_{t} | c, z^m, q)  = - \frac{\epsilon_t}{\sqrt{1-\bar{\alpha}_t}} + \gamma_c \nabla_{z_{t}} \log p(c | z_{t}) \\
    & + \gamma_m \nabla_{z_{t}}\log p(z^m | z_{t}) + \gamma_q \nabla_{z_{t}} \log p(q|z_{t}),
\end{split}
\end{equation}
where $\gamma$ terms control the guidance strength. This novel decomposition enables us to use specialized, off-the-shelf reward models for each objective, leading to more refined and controllable inpainting results. Specifically, we employ a reward model $r_c$ for text alignment, $r_m$ for visual rationality, and $r_q$ for human preference. Since the probability of each objective is proportional to the exponential of its reward score (i.e., $p(\cdot | z_{t}) \propto \exp(r(z_{t}))$) \cite{song2021ebm}, we can directly refine the predicted noise:
\begin{equation}
\begin{split}
    \hat\epsilon_t = & \ \epsilon_\theta(z_t, t, c, z^m, M^\prime)  - \gamma_c \sqrt{\bar{\alpha}_t} \nabla_{z_{t}} r_c(z_t, c) \\
    & - \gamma_m \sqrt{\bar{\alpha}_t} \nabla_{z_{t}} r_m(z_t, z^m) - \gamma_q \sqrt{\bar{\alpha}_t} \nabla_{z_{t}} r_q(z_{t}, c),
\end{split}
\label{eq:guidance-epsilon}
\end{equation}
Notably, we use $\sqrt{\bar{\alpha}_t}$ as the reward modulator instead of the conventional $\sqrt{1-\bar{\alpha}_t}$ in Sec. \ref{sec:diffusion-guidance}. As $\sqrt{\bar{\alpha}_t}$ is monotonically increasing during denoising, it down-weights the influence of unreliable predictions from initial noisy steps, which we find empirically yields better performance. Finally, to keep the unmasked region unchanged, we blend the inpainted image with the original input. See pseudocode in Appendix A.

\section{Experiment}

\subsection{Experimental Settings}

\subsubsection{Baselines}
We compare \emph{FreeInpaint} against a suite of competitive baselines, including SD1.5-Inpainting (SDI) \cite{rombach2022ldm}, PowerPaint (PPT) \cite{zhuang2024ppt}, BrushNet-Random (BN) \cite{ju2024brushnet}, SDXL-Inpainting (SDXLI) \cite{podell2023sdxl}, and the state-of-the-art DiT-based \cite{peebles2023dit} SD3-ControlNet-Inpainting (SD3I) \cite{SD3-Controlnet-Inpainting}. PPT uses paired mask-description data for training, while others do not. Our key competitor is HD-Painter (HDP) \cite{manukyan2023hd}, a training-free method focused only on prompt alignment, while the other baselines are training-based. The same random seeds are used across all experiments.

\subsubsection{Implementation Details}
We evaluate our method and HDP by integrating them with SDI, PPT, BN, SDXLI, SD3I. In DeGu, the reward models are: local CLIPScore \cite{radford2021clip} $r_c$ to assess the alignment between the prompt $c$ and the content generated within the mask $M$, InpaintReward \cite{liu2024prefpaint} $r_m$ to evaluate the visual coherence, and ImageReward \cite{xu2023imagereward} $r_q$ to measure human preference. Other details are in Appendix B.

\subsubsection{Benchmarks}
We employ two benchmarks: EditBench \cite{wang2023imageneditor} and MSCOCO \cite{lin2014mscoco}. 
EditBench provides handcrafted, scribble-like free-form masks. For extended evaluation, we also use precise, object-aligned layout masks from the MSCOCO validation set. As original MSCOCO labels are single words, we generate detailed mask captions using LLaVA \cite{liu2024llavanext} to better assess prompt alignment. More details are in Appendix B.

We define two prompt types: the \textit{global prompt}, which describes the entire image (the "Full prompt" in EditBench and the caption of an entire image in MSCOCO), and the \textit{local prompt}, which is a detailed, mask-level description (the "Mask-Rich prompt" in EditBench and our annotated mask caption in MSCOCO). For inference, we use the \textit{local prompt} as it better simulates real-world scenarios, such as adding content to a photograph.

\subsubsection{Metrics}
To prevent metric leakage, we employ multiple metrics beyond DeGu's rewards. We assess: human preference with ImageReward and HPSv2 \cite{wu2023hpsv2}, using \textit{global prompts} to be consistent with prior works \cite{ju2024brushnet} and better gauge overall image quality (distinct from the local prompts used during inference); prompt alignment with local and global CLIPScore; and visual rationality with InpaintReward and LPIPS \cite{zhang2018lpips}. The lower the LPIPS, the closer the inpainted image is to the ground truth. A user study was also conducted for thorough, unbiased comparison.

\subsection{Quantitative Results}

\subsubsection{Free-form Mask Inpainting}
\begin{table}[t]
\centering
\scriptsize 
\begin{tabular}{l @{\hspace{3pt}} c @{\hspace{3pt}} c @{\hspace{3pt}} c @{\hspace{3pt}} c @{\hspace{3pt}} c @{\hspace{3pt}} c @{\hspace{3pt}} c}
\toprule
                       &       & ImageReward     & HPSv2          & L.CLIP    & G.CLIP    & InpaintReward    & LPIPS $\downarrow$\\ \midrule
\multirow{3}{*}{SDI}   & Base  & -0.1341         & 23.36          & 23.90          & 26.13          & -0.1732          & 0.2073          \\
                       & +HDP  & -0.0110         & 22.99          & \textbf{26.09} & 26.85          & -0.3075          & 0.2419          \\
                       & +Ours & \textbf{0.1753} & \textbf{23.65} & 25.92          & \textbf{27.59} & \textbf{-0.0595} & \textbf{0.2011} \\ \midrule
\multirow{3}{*}{PPT}   & Base  & 0.0842          & 23.72          & 25.12          & 26.47          & -0.2133          & 0.2244          \\
                       & +HDP  & 0.0554          & 23.30          & 25.51          & 26.58          & -0.2445          & 0.2436          \\
                       & +Ours & \textbf{0.2361} & \textbf{23.94} & \textbf{25.69} & \textbf{27.52} & \textbf{-0.1262} & \textbf{0.2116} \\ \midrule
\multirow{3}{*}{BN}    & Base  & 0.2729          & 25.34          & 26.45          & 27.18          & -0.1791          & \textbf{0.1947} \\
                       & +HDP  & 0.3836          & 25.20          & 27.08          & 27.69          & -0.2124          & 0.2135          \\
                       & +Ours & \textbf{0.5006} & \textbf{25.64} & \textbf{27.81} & \textbf{28.28} & \textbf{-0.0878} & 0.2005          \\ \midrule
\multirow{3}{*}{SDXLI} & Base  & 0.2318          & 25.10          & 25.87          & 27.93          & -0.2792          & 0.2826          \\
                       & +HDP  & 0.1543          & 24.34          & \textbf{26.62} & \textbf{28.38} & -0.3463          & 0.3206          \\
                       & +Ours & \textbf{0.3455} & \textbf{25.14} & \textbf{26.62} & 28.29          & \textbf{-0.1888} & \textbf{0.2087} \\ \midrule
\multirow{3}{*}{SD3I} & Base  & 0.2993          & 25.48          & 26.26          & 27.43          & -0.2170          & 0.2155          \\
                       & +HDP  & -0.5020         & 21.56          & 22.83          & 25.36          & -0.2988          & 0.2516          \\
                       & +Ours & \textbf{0.5248} & \textbf{25.70} & \textbf{26.98} & \textbf{28.31} & \textbf{-0.0694} & \textbf{0.2057} \\ \bottomrule
\end{tabular}
  \caption{Quantitative comparisons on EditBench dataset with free-form masks. L/G.CLIP refer to Local/Global CLIP. \textbf{Bold} indicates the best performance.}
  \label{tab:editbench}
\end{table}
On EditBench with scribble-like free-form masks (Tab. \ref{tab:editbench}), \emph{FreeInpaint} comprehensively improves all baselines, including both U-Net and DiT architectures. As our guidance directly targets ImageReward, local CLIPScore, and InpaintReward, these metrics show substantial gains as expected. Notably, our method boosts the global-prompt-based ImageReward even while using only local prompts in DeGu. Non-guidance metrics like HPSv2, global CLIP, and LPIPS also improve obviously, indicating holistic enhancement.

Furthermore, \emph{FreeInpaint} consistently outperforms HDP. This superiority is especially clear in cases where HDP degrades image quality. For instance, with the SDI baseline, HDP worsens HPSv2, InpaintReward, and LPIPS scores, whereas our method provides substantial improvements. This highlights \emph{FreeInpaint}'s ability to simultaneously enhance prompt alignment and visual rationality, which we attribute to our strategy of decoupling and separately optimizing for these two objectives. Furthermore, unlike our approach, HDP is incompatible with the DiT-based SD3I, highlighting \emph{FreeInpaint}'s superior adaptability.

\subsubsection{Layout Mask Inpainting}
\begin{table}[t]
\centering
\scriptsize 
\begin{tabular}{l @{\hspace{3pt}} c @{\hspace{3pt}} c @{\hspace{3pt}} c @{\hspace{3pt}} c @{\hspace{3pt}} c @{\hspace{3pt}} c @{\hspace{3pt}} c}
\toprule
                       &       & ImageReward     & HPSv2          & L.CLIP    & G.CLIP    & InpaintReward    & LPIPS $\downarrow$\\ \midrule
\multirow{3}{*}{SDI}   & Base  & 0.2574          & 25.50          & 24.09          & 24.63          & -0.0206         & 0.1126          \\
                       & +HDP  & \textbf{0.2601} & 24.43          & \textbf{24.53} & 24.36          & -0.0346         & 0.1231          \\
                       & +Ours & 0.2591          & \textbf{25.69} & 24.41          & \textbf{24.94} & \textbf{0.0112} & \textbf{0.0844} \\ \midrule
\multirow{3}{*}{PPT}   & Base  & 0.2593          & 25.52          & 24.67          & 24.72          & -0.0149         & 0.1188          \\
                       & +HDP  & 0.2167          & 25.51          & 24.91          & 24.80          & -0.0201         & 0.1180          \\
                       & +Ours & \textbf{0.2671} & \textbf{25.79} & \textbf{25.09} & \textbf{24.91} & \textbf{0.0171} & \textbf{0.0861} \\ \midrule
\multirow{3}{*}{BN}    & Base  & 0.2701          & \textbf{26.02} & 24.51          & 24.76          & 0.0140          & 0.0867          \\
                       & +HDP  & 0.2766          & 25.98          & \textbf{24.87} & \textbf{24.98} & 0.0097          & 0.0876          \\
                       & +Ours & \textbf{0.2782} & \textbf{26.02} & 24.69          & 24.80          & \textbf{0.0568} & \textbf{0.0862} \\ \midrule
\multirow{3}{*}{SDXLI} & Base  & 0.2665          & 26.10          & 24.38          & \textbf{24.95} & \textbf{-0.0439}& 0.1802          \\
                       & +HDP  & 0.1810          & 24.22          & 23.73          & 24.54          & -0.0565         & 0.2640          \\
                       & +Ours & \textbf{0.3191} & \textbf{26.89} & \textbf{24.66} & 24.89          & -0.0504         & \textbf{0.0683} \\ \midrule
\multirow{3}{*}{SD3I} & Base  & 0.2795          & 26.51          & 24.62          & 24.84          & 0.0093           & 0.1008          \\
                       & +HDP  & -0.0010         & 24.99          & 23.52          & 24.62          & -0.1005          & 0.1072          \\
                       & +Ours & \textbf{0.3422} & \textbf{27.10} & \textbf{25.14} & \textbf{25.13} & \textbf{0.0273} & \textbf{0.0680} \\ \bottomrule
\end{tabular}
  \caption{Quantitative comparisons on MSCOCO dataset with layout masks. L/G.CLIP refer to Local/Global CLIP. \textbf{Bold} indicates the best performance.}
  \label{tab:coco}
\end{table}

\begin{table}[t]
\centering
\scriptsize 
\begin{tabular}{l @{\hspace{3.4pt}} c @{\hspace{3.4pt}} c @{\hspace{3.4pt}} c @{\hspace{3.4pt}} c @{\hspace{3.4pt}} c @{\hspace{3.4pt}} c @{\hspace{3.4pt}} c }
\toprule
                       & ImageReward     & HPSv2          & L.CLIP    & G.CLIP    & InpaintReward    & LPIPS $\downarrow$\\ \midrule
BN                                  & 0.2729          & 25.34          & 26.45          & 27.18          & -0.1791          & 0.1947          \\
BN + PriNo             & 0.3785          & 25.57          & 26.96          & 27.72          & -0.2124          & 0.2021          \\
BN + DeGu            & 0.3908          & 25.36          & 27.17          & 27.87          & \textbf{-0.0643} & 0.1944          \\
Constant & 0.3533          & 25.41          & 26.92          & 27.56          & -0.1088          & 0.1936          \\
$\sqrt{1-\bar{\alpha}_t}$ & 0.3454          & 25.27          & 26.85          & 27.53          & -0.1146          & \textbf{0.1922} \\
Ours                                & \textbf{0.5006} & \textbf{25.64} & \textbf{27.81} & \textbf{28.28} & -0.0878 & 0.2005          \\ \bottomrule
\end{tabular}
  \caption{Quantitative ablation study on EditBench dataset.}
  \label{tab:ablation}
\end{table}

To evaluate generalization, we test on MSCOCO's object-aligned layout masks (Tab. \ref{tab:coco}). Our method's strong performance extends to this dataset, confirming its robustness across different mask types. \emph{FreeInpaint} demonstrates remarkable effectiveness even on strong, specialized baselines, achieving top performance across all metrics for both PPT (which is trained on paired mask-description data) and the DiT-based SD3I.

While HDP is competitive on some metrics (e.g., Local CLIP on BN), our method provides a better overall trade-off. Take BN for example, \emph{FreeInpaint} achieves better ImageReward, InpaintReward, and LPIPS at the cost of a narrow margin on CLIP score. This confirms that our decoupled optimization strategy achieves more robust enhancements in both prompt alignment and visual rationality.

\subsubsection{User Study}
Considering that quantitative metrics may not perfectly align with human preferences in image generation tasks, we conducted a user study. We randomly selected 30 samples from EditBench and MSCOCO datasets. 59 evaluators with diverse educational backgrounds independently assessed the generated results from SDI, SDI + HDP, and SDI + Ours, selecting the best based on prompt alignment and overall quality. The win rates of the three models are 16.16\%, 19.32\%, and 64.52\%, respectively, which demonstrate a significant user preference for our method.

\subsection{Qualitative Results}
\begin{figure}[t]
    \centering
    \renewcommand{\arraystretch}{0.4} 
    \footnotesize 
    \setlength{\tabcolsep}{0.5pt} 

    \begin{tabular}{
        >{\centering\arraybackslash}b{0.03\linewidth} 
        >{\centering\arraybackslash}p{0.185\linewidth} 
        >{\centering\arraybackslash}p{0.185\linewidth} 
        >{\centering\arraybackslash}p{0.185\linewidth} 
        >{\centering\arraybackslash}p{0.185\linewidth} 
        >{\centering\arraybackslash}p{0.185\linewidth}}

        & \scriptsize A wooden cat in beige color
        & \scriptsize One cat facing away
        & \scriptsize A rough beige crochet ball
        & \scriptsize A vandalized stop sign 
        & \scriptsize Man holding a cell phone \\

        \rotatebox{90}{\scriptsize Input}
        & \includegraphics[width=\linewidth]{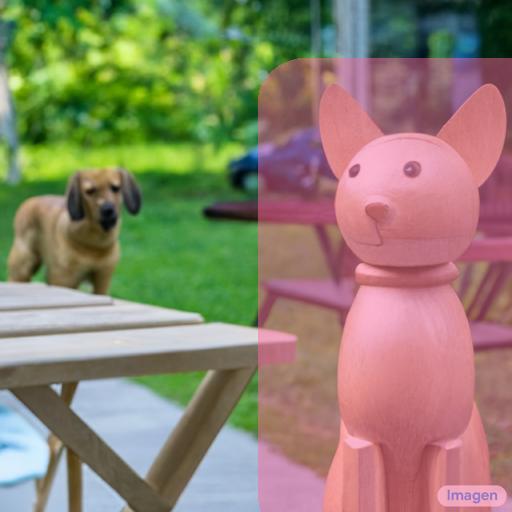}
        & \includegraphics[width=\linewidth]{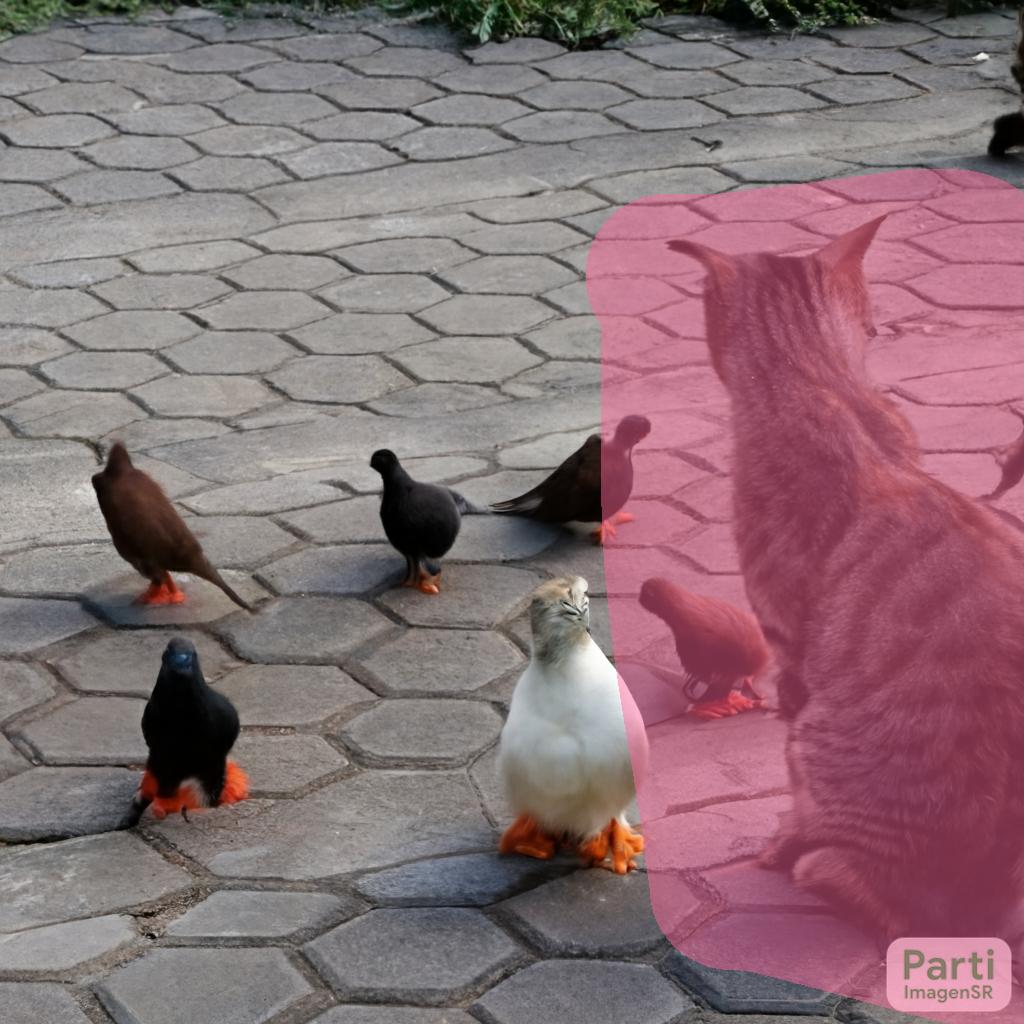}
        & \includegraphics[width=\linewidth]{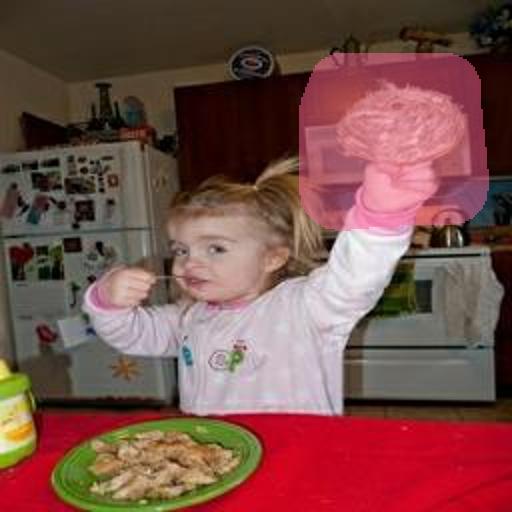}
        & \includegraphics[width=\linewidth]{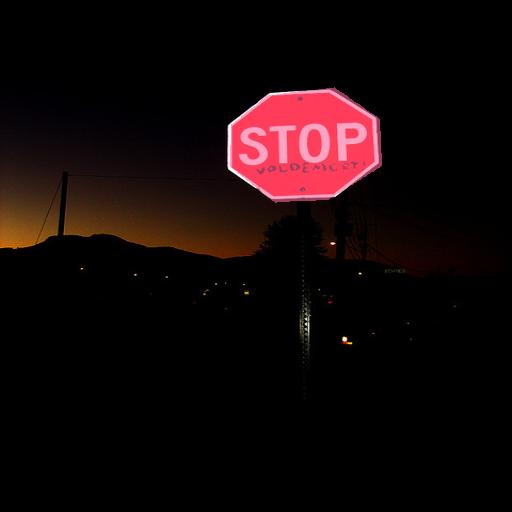}
        & \includegraphics[width=\linewidth]{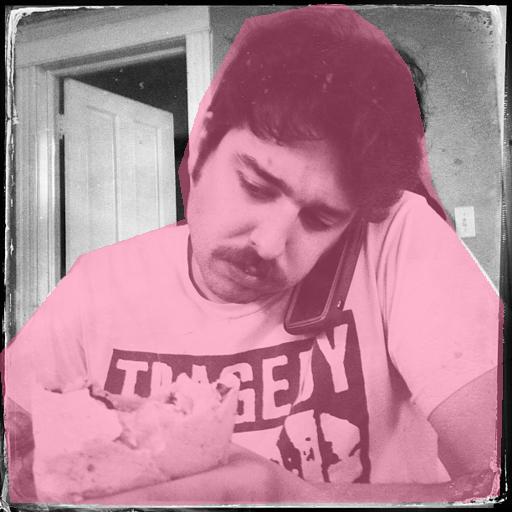} \\

        \rotatebox{90}{\scriptsize SDI}
        & \includegraphics[width=\linewidth]{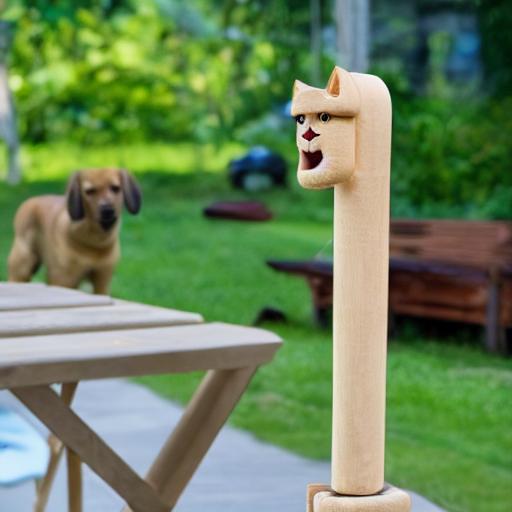}
        & \includegraphics[width=\linewidth]{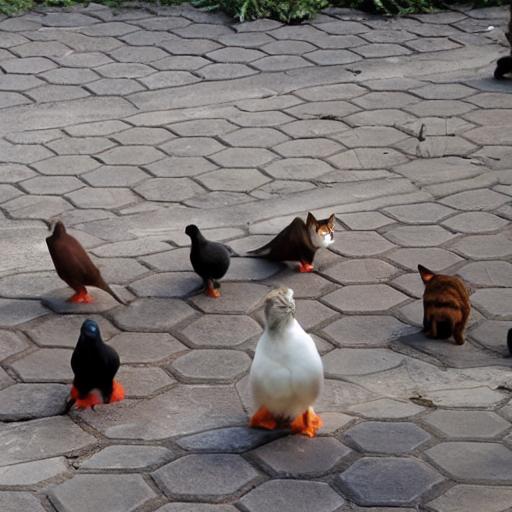}
        & \includegraphics[width=\linewidth]{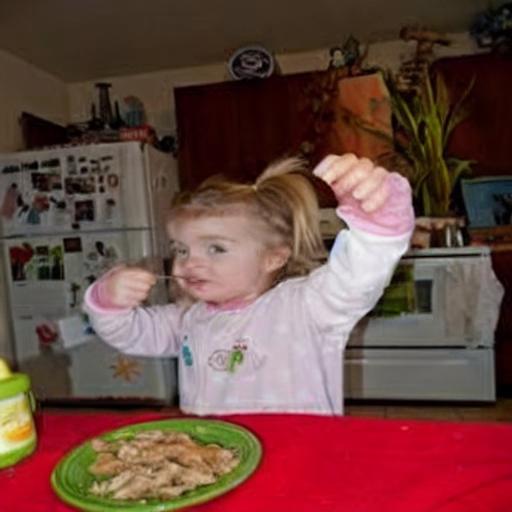}
        & \includegraphics[width=\linewidth]{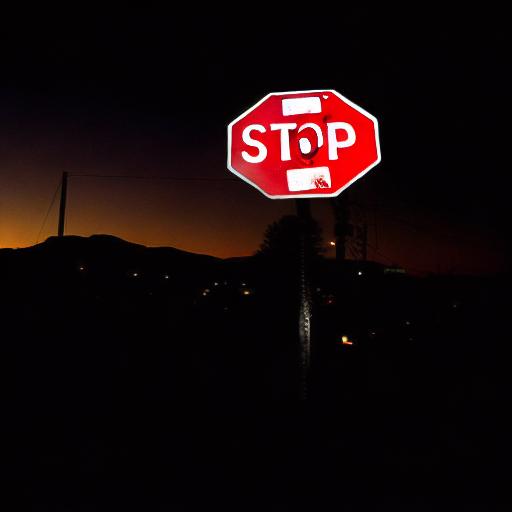}
        & \includegraphics[width=\linewidth]{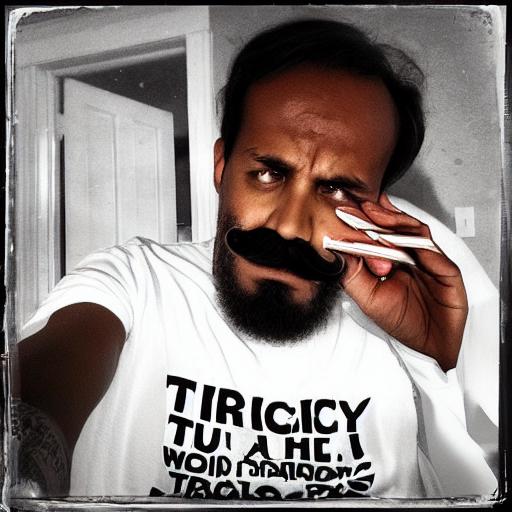} \\

        \rotatebox{90}{\scriptsize SDI+HDP}
        & \includegraphics[width=\linewidth]{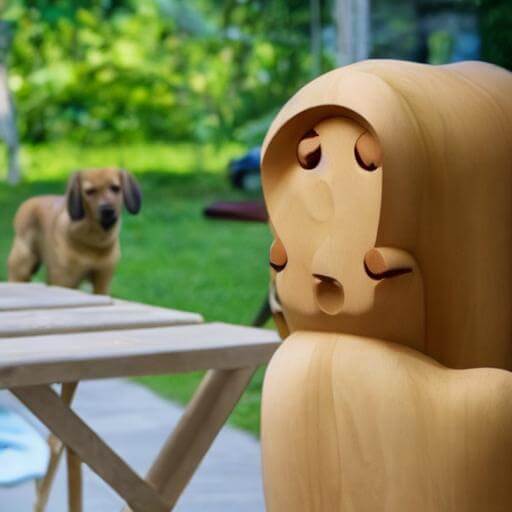}
        & \includegraphics[width=\linewidth]{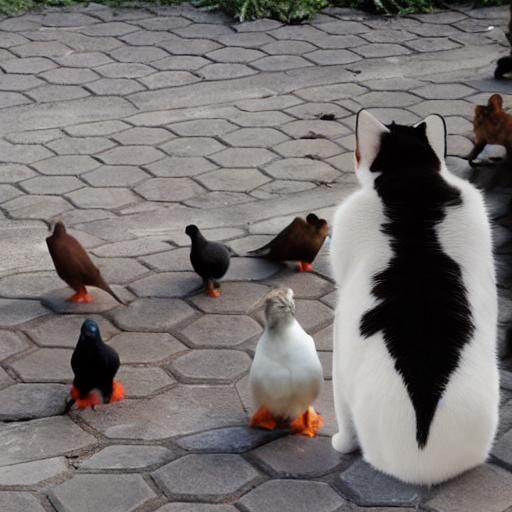}
        & \includegraphics[width=\linewidth]{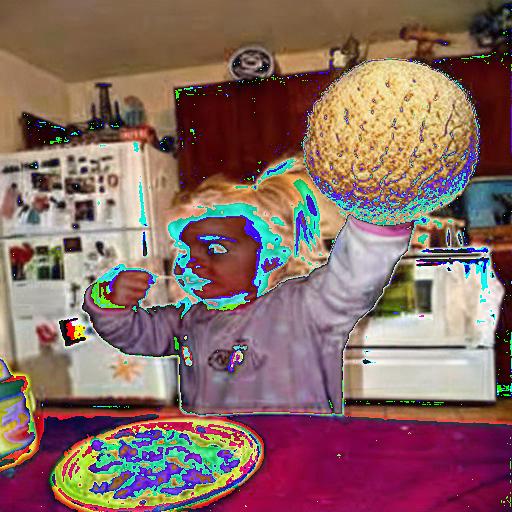}
        & \includegraphics[width=\linewidth]{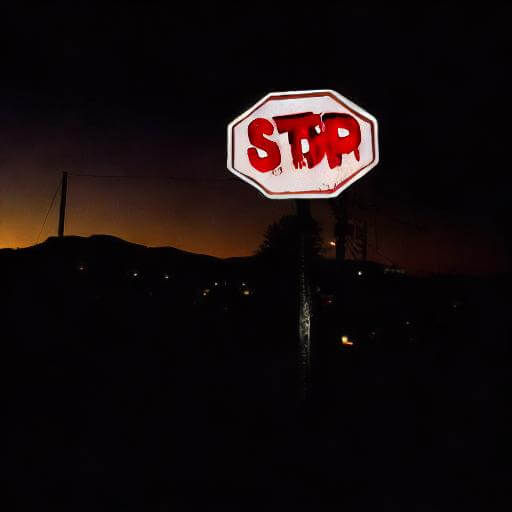}
        & \includegraphics[width=\linewidth]{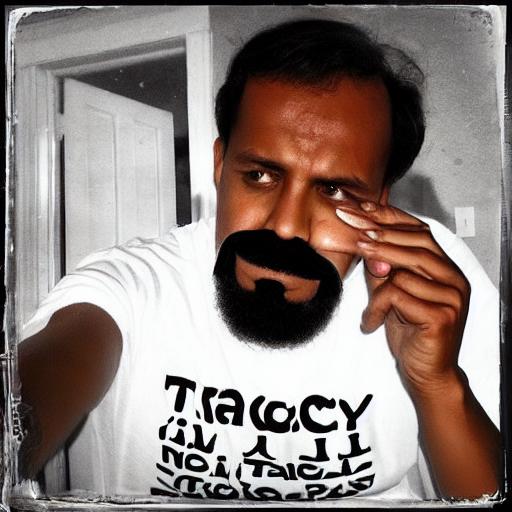} \\

        \rotatebox{90}{\scriptsize SDI+Ours}
        & \includegraphics[width=\linewidth]{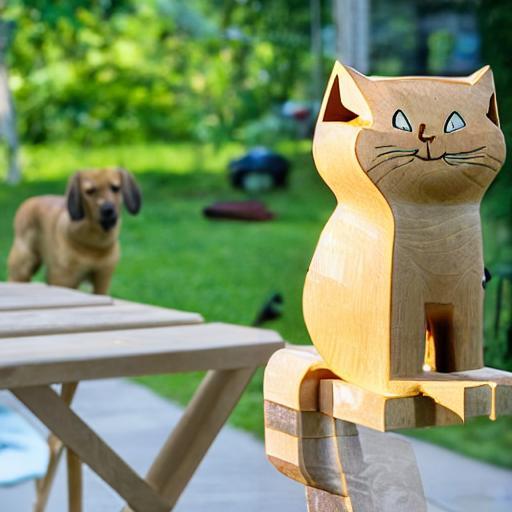}
        & \includegraphics[width=\linewidth]{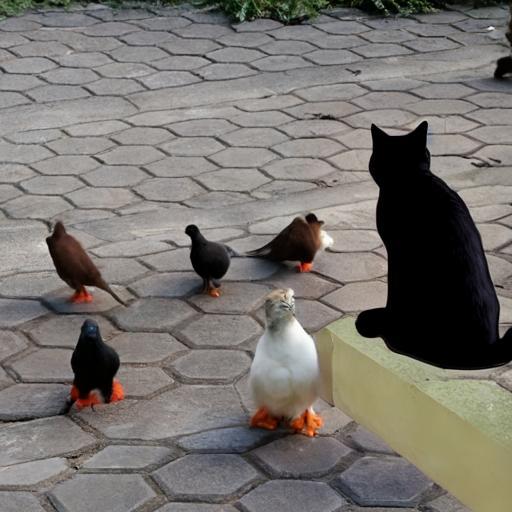}
        & \includegraphics[width=\linewidth]{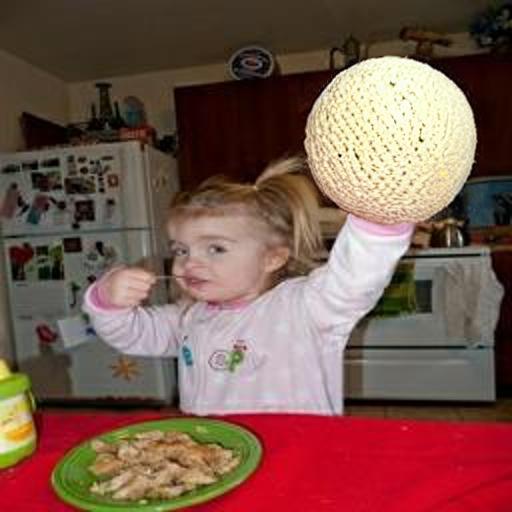}
        & \includegraphics[width=\linewidth]{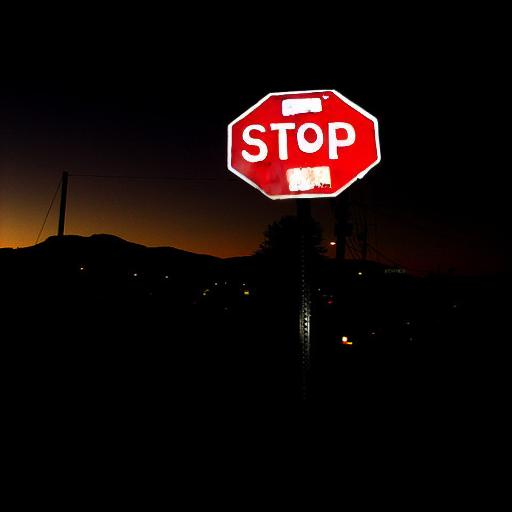}
        & \includegraphics[width=\linewidth]{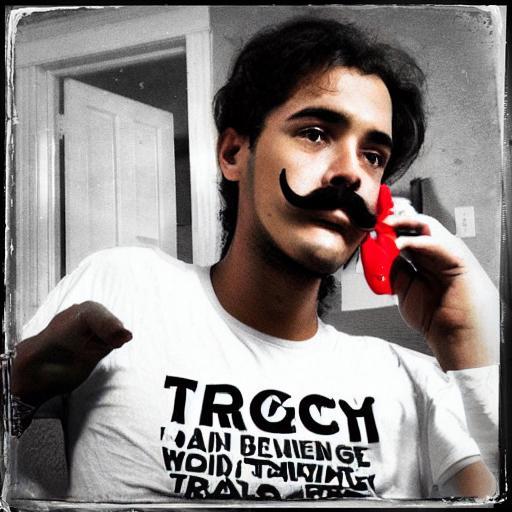} \\

        & \scriptsize I
        & \scriptsize II
        & \scriptsize III
        & \scriptsize IV
        & \scriptsize V \\

    \end{tabular}
    \caption{Comparison against SDI-based approaches.}

    \label{fig:qualitative-sdi}
\end{figure}

\begin{figure}[!t]
    \centering
    \renewcommand{\arraystretch}{0.4} 
    \footnotesize 
    \setlength{\tabcolsep}{0.5pt} 

    \begin{tabular}{
        >{\centering\arraybackslash}b{0.03\linewidth} 
        >{\centering\arraybackslash}p{0.185\linewidth} 
        >{\centering\arraybackslash}p{0.185\linewidth} 
        >{\centering\arraybackslash}p{0.185\linewidth} 
        >{\centering\arraybackslash}p{0.185\linewidth} 
        >{\centering\arraybackslash}p{0.185\linewidth}}

        & \scriptsize Letters "bike" on a board
        & \scriptsize Cat reaching for a basketball
        & \scriptsize Yellow dog
        & \scriptsize Man sitting in a wheelchair
        & \scriptsize Snowboarder tricking \\

        \rotatebox{90}{\scriptsize Input}
        & \includegraphics[width=\linewidth]{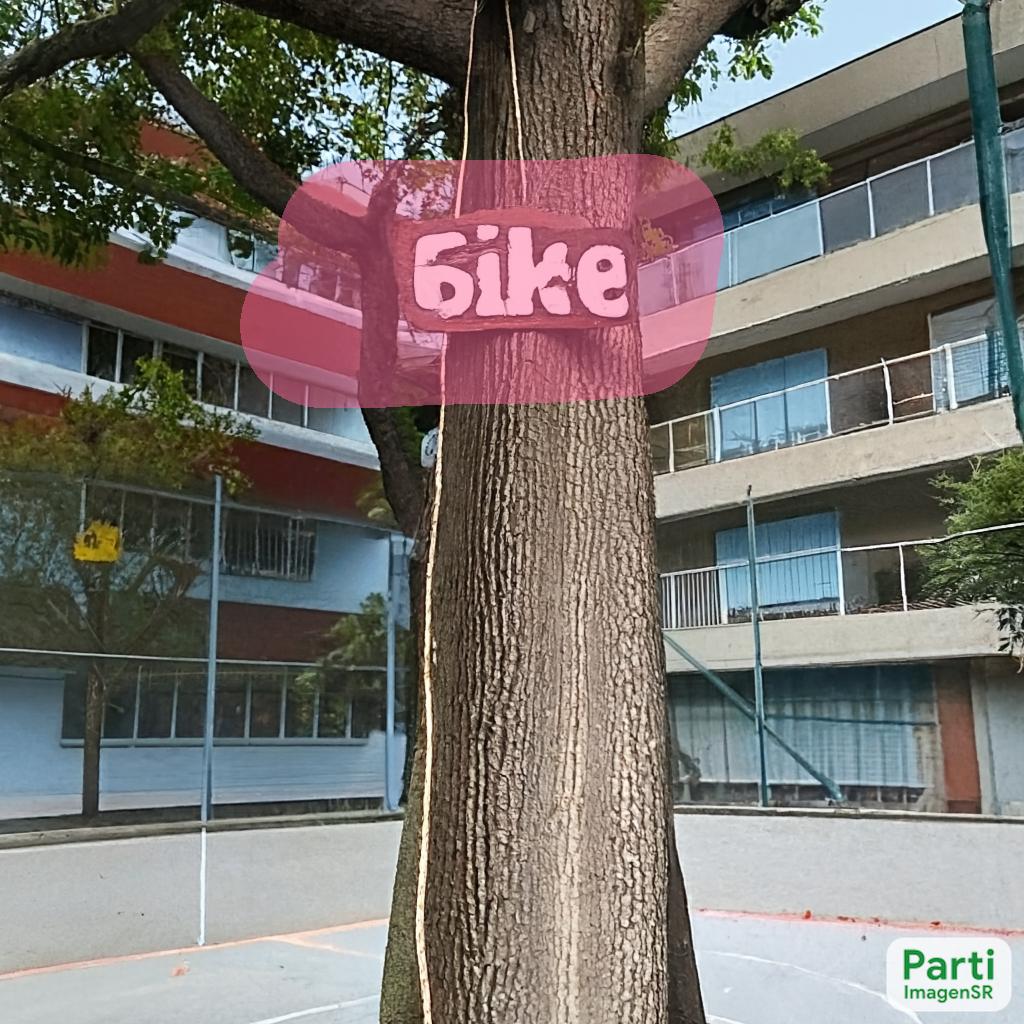}
        & \includegraphics[width=\linewidth]{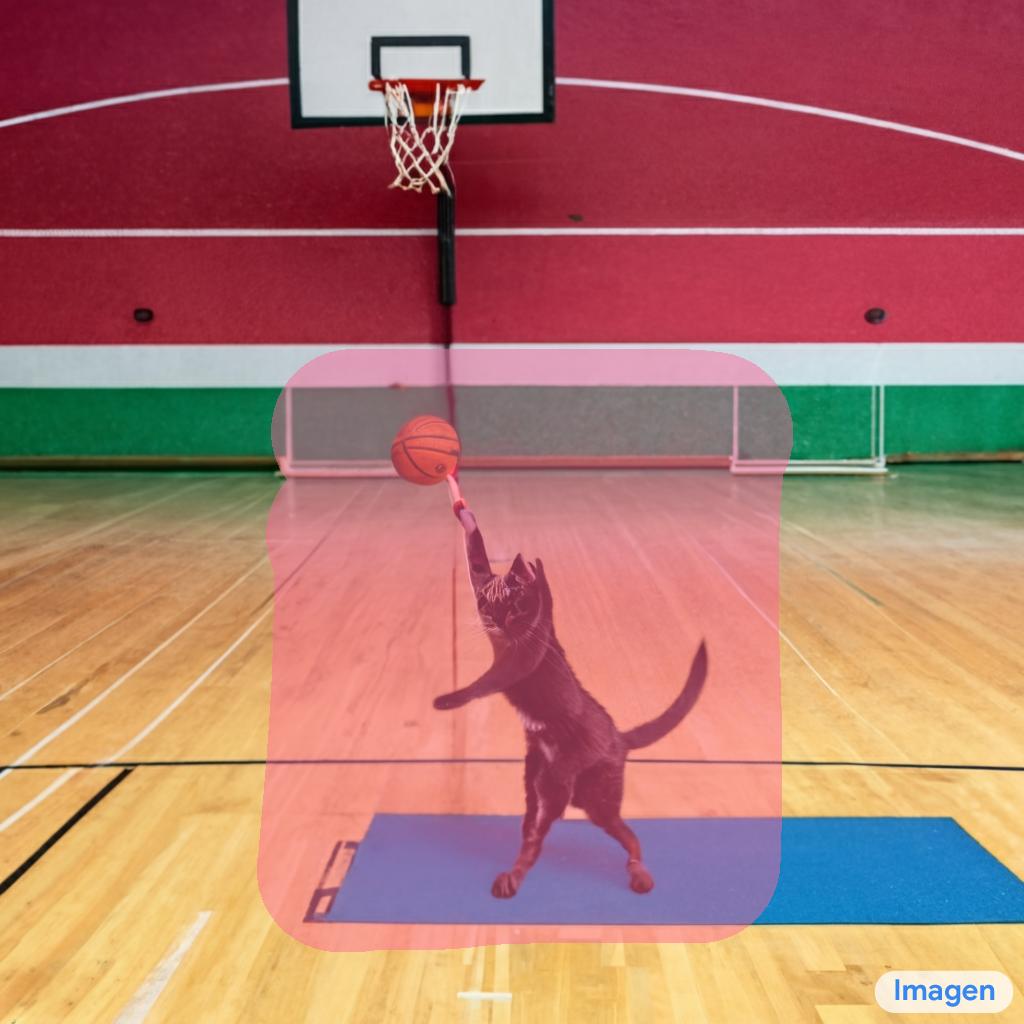}
        & \includegraphics[width=\linewidth]{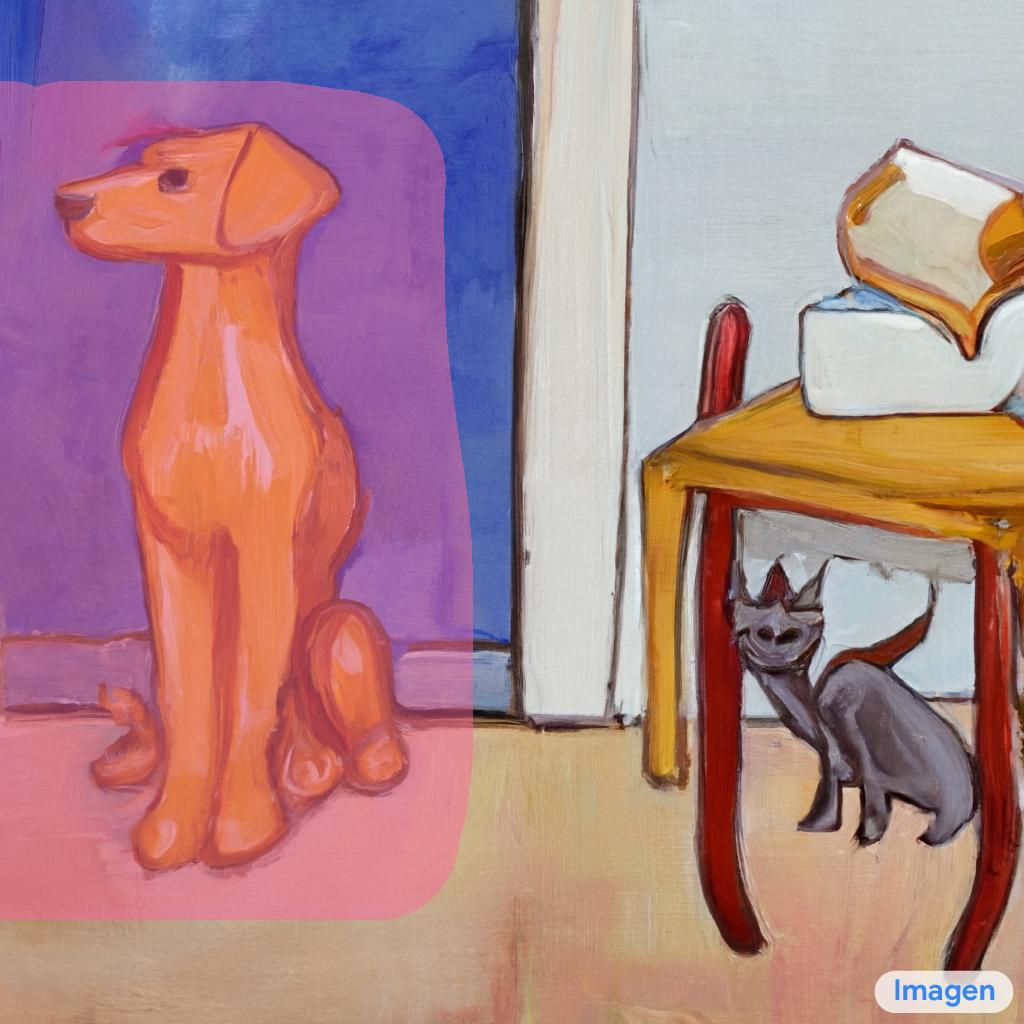}
        & \includegraphics[width=\linewidth]{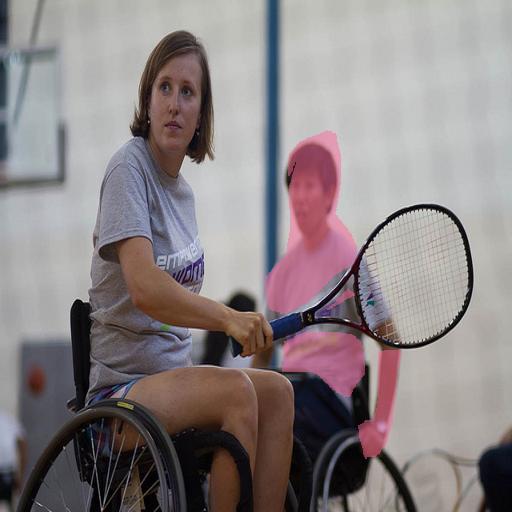}
        & \includegraphics[width=\linewidth]{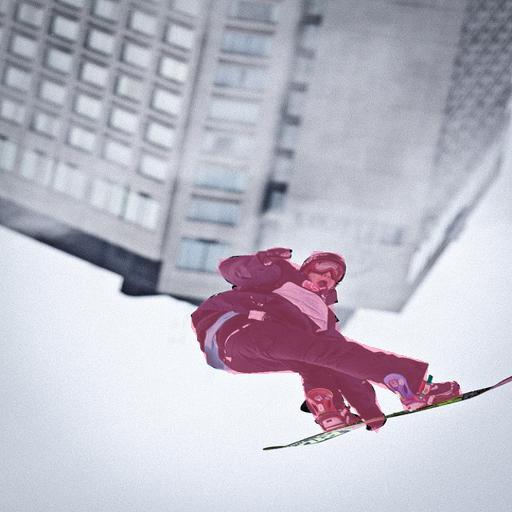} \\

        \rotatebox{90}{\scriptsize BN}
        & \includegraphics[width=\linewidth]{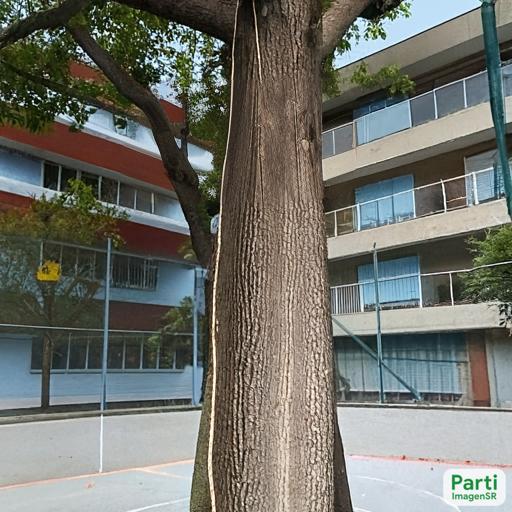}
        & \includegraphics[width=\linewidth]{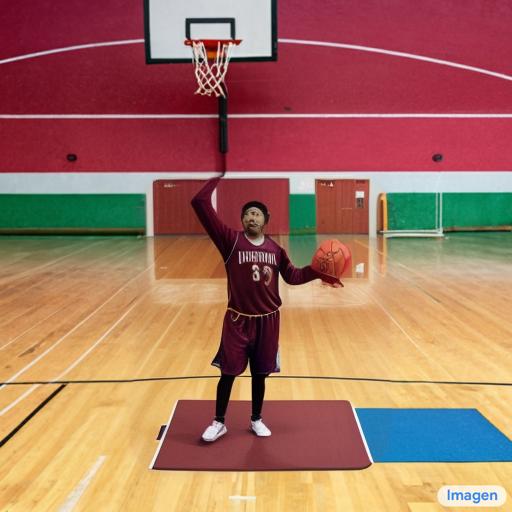}
        & \includegraphics[width=\linewidth]{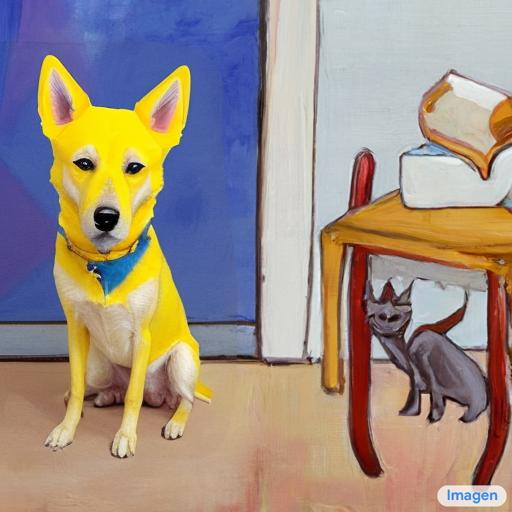}
        & \includegraphics[width=\linewidth]{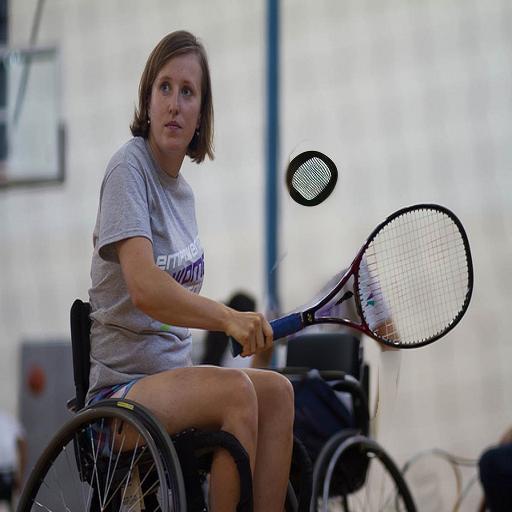}
        & \includegraphics[width=\linewidth]{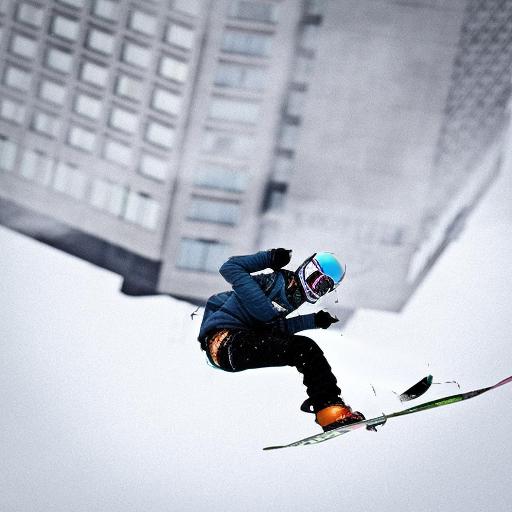} \\

        \rotatebox{90}{\scriptsize BN+HDP}
        & \includegraphics[width=\linewidth]{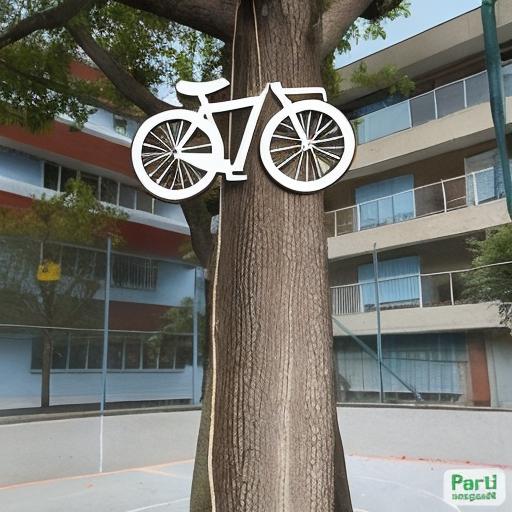}
        & \includegraphics[width=\linewidth]{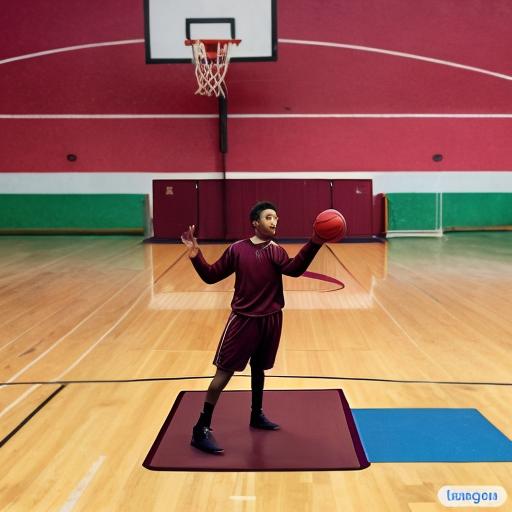}
        & \includegraphics[width=\linewidth]{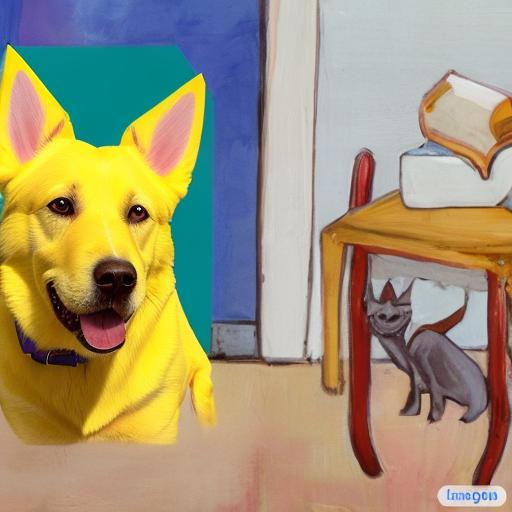}
        & \includegraphics[width=\linewidth]{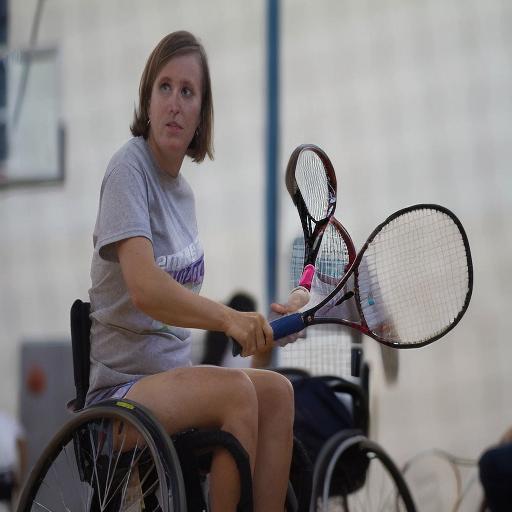}
        & \includegraphics[width=\linewidth]{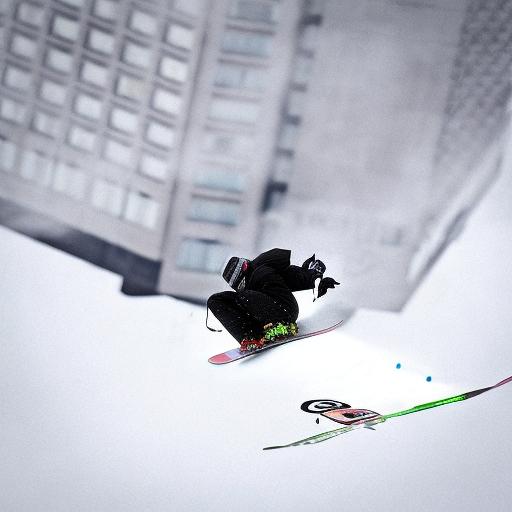} \\

        \rotatebox{90}{\scriptsize BN+Ours}
        & \includegraphics[width=\linewidth]{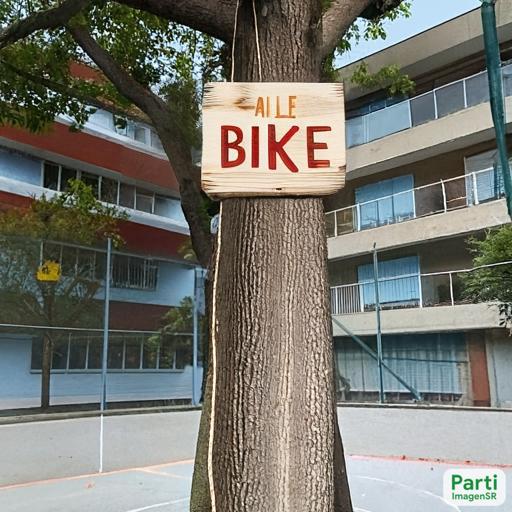}
        & \includegraphics[width=\linewidth]{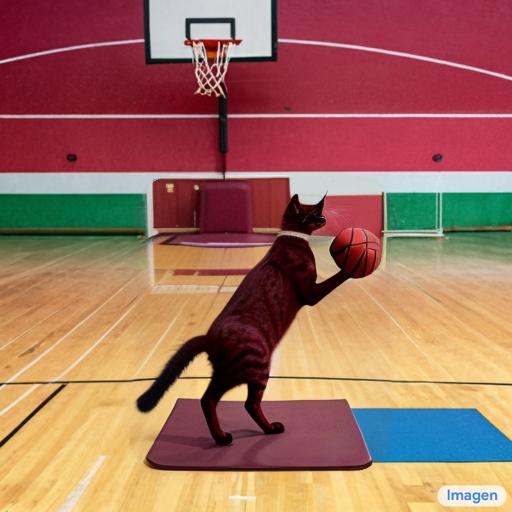}
        & \includegraphics[width=\linewidth]{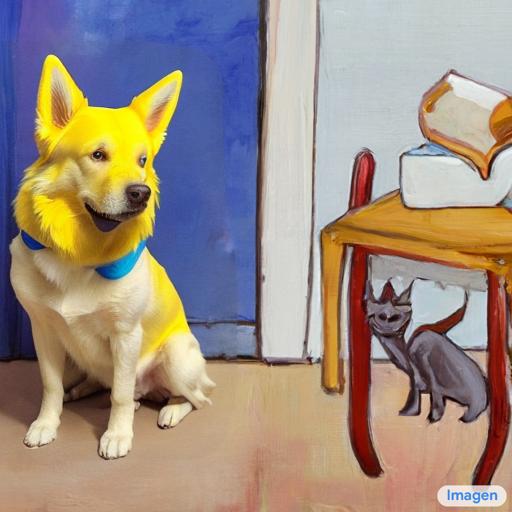}
        & \includegraphics[width=\linewidth]{images/qualitative/coco_custom/258_000000044877-ours-brush.jpg}
        & \includegraphics[width=\linewidth]{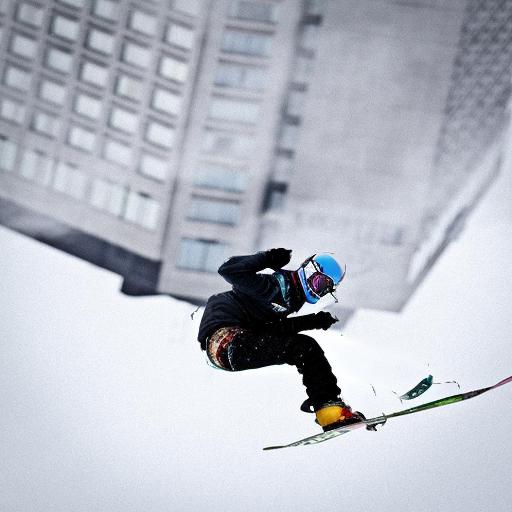} \\
        
        & \scriptsize I
        & \scriptsize II
        & \scriptsize III
        & \scriptsize IV
        & \scriptsize V \\

    \end{tabular}
    \caption{Comparison against BN-based approaches.}

    \label{fig:qualitative-brush}
\end{figure}

Qualitative results are shown in Fig. \ref{fig:qualitative-sdi} and \ref{fig:qualitative-brush} (prompts simplified for presentation). In Fig. \ref{fig:qualitative-brush}, both baseline methods BN and HDP struggle to render text (I) or generate the correct content (II, IV), while HDP produces an unnatural boundary (III). In Fig. \ref{fig:qualitative-sdi}, the outputs of the baselines are either unaesthetic (I) or visually flawed (III). In contrast, our method resolves all these shortcomings.

\subsection{Ablation Study}
We conduct an ablation study on EditBench using BrushNet to analyse each component's effect (Tab. \ref{tab:ablation} and Fig. \ref{fig:role-of-guidance}). The results confirm that our main components, Sec. \ref{sec:NoiseOpt} PriNo and Sec. \ref{sec:guidance} DeGu, are individually effective and their benefits are additive. PriNo improves prompt-related metrics but reduces the visual rationality metric InpaintReward, while DeGu compensates for this. We also analyze our guidance modulator,$\sqrt{\bar{\alpha}_t}$, against alternatives $\sqrt{1-\bar{\alpha}_t}$ and a constant 0.5. Our chosen modulator demonstrates superior performance across most metrics. For more ablation analyses, please refer to Appendix C.

\section{Conclusion}
In this work, we present \emph{FreeInpaint}, a plug-and-play tuning-free framework for text-guided image inpainting. By optimizing the initial noise with priors tailored for inpainting and guiding intermediate latents with a task-specific composite objective, our method achieves both superior prompt alignment and visual rationality. Specifically, the proposed noise optimization steers attention toward masked regions in the initial denoising process, while the composite reward—combining text alignment, visual rationality, and human preference—effectively guides the inpainting process without training. Experiments on multiple inpainting diffusion models and benchmarks demonstrate the effectiveness and generalizability of our \emph{FreeInpaint}.

\section*{Acknowledgments}
This work was supported by NSFC project (No. 62232006).

\bibliography{aaai2026}

\clearpage
\appendix
\section*{Appendix}

\section{Algorithms}
\subsection{Prior-Guided Noise Optimization}
In PriNo, we iteratively optimize the initial noise over $\tau_{\text{round}}$ rounds. During each round, the process operates for a maximum of $\tau_{\text{iter}}$ iterations utilizing an SGD optimizer with a specified learning rate $lr$. Furthermore, we incorporate an early-stopping mechanism with predefined thresholds $\tau_\text{c}$, $\tau_\text{s}$, and $\tau_\text{KL}$ for the losses $\mathcal{L}_\text{c}$, $\mathcal{L}_\text{s}$, and $\mathcal{L}_\text{KL}$, respectively. The complete procedure is detailed in Algorithm \ref{alg:noise_opt}.

\begin{algorithm}[h]
\caption{\textsc{PriNo}}
\label{alg:noise_opt}
\textbf{Input}: Inpainting model $\epsilon_\theta$, text condition $c$, masked latent $z^m$, downsampled mask $M^\prime$, thresholds $\tau_\text{c}$, $\tau_\text{s}$, $\tau_\text{KL}$, $\tau_{\text{iter}}$, $\tau_{\text{round}}$, learning rate $lr$, Optimizer SGD \\
\textbf{Output}: Optimized noise $\hat{\mathbf{z}}_T$
\begin{algorithmic}[1]
\STATE $\mathcal{Z} \leftarrow \emptyset$ \COMMENT{Initialize noise pool}
\FOR{$i = 1$ to $\tau_{\text{round}}$}
    \STATE $\mathbf{z}_T \sim \mathcal{N}(\mathbf{0}, \mathbf{1})$; $\mu \leftarrow \mathbf{0}$; $\sigma \leftarrow \mathbf{1}$; $t\leftarrow$ initial step;
    \FOR{$j = 1$ to $\tau_{\text{iter}}$}
        \STATE Extract $\mathbf{A}^c_{t}$, $\mathbf{A}^s_{t}$ from $\epsilon_\theta(\mu + \sigma \mathbf{z}_T, t, c, z^m, M^\prime)$;
        \STATE Compute $\mathcal{L}_\text{c}$, $\mathcal{L}_\text{s}$, $\mathcal{L}_\text{KL}$, and $\mathcal{L}_{\text{joint}}$;
        \IF{$\mathcal{L}_\text{c} < \tau_\text{c}$ \textbf{and} $\mathcal{L}_\text{s} < \tau_\text{s}$}
            \STATE \textbf{return} $\mu + \sigma \mathbf{z}_T$;
        \ENDIF
        \IF{$\mathcal{L}_\text{KL} > \tau_\text{KL}$}
            \STATE \textbf{break};
        \ENDIF
        \STATE $\mu, \sigma \leftarrow \text{SGD}(\mu, \sigma, \mathcal{L}_{\text{joint}}, lr)$;
    \ENDFOR
    \STATE $\mathcal{Z} \leftarrow \mathcal{Z} \cup \{ \mu + \sigma \mathbf{z}_T \}$;
\ENDFOR
\STATE \textbf{return} $\underset{\mathbf{z}_T\in \mathcal{Z}}{\text{argmin}} \ \mathcal{L}_{\text{joint}}(\mathbf{z}_T)$
\end{algorithmic}
\end{algorithm}

\subsection{Decomposed Training-free Guidance}
In DeGu, we guide the denoising process by applying differentiable rewards to optimize the predicted noise. Specifically, we utilize three distinct reward models: $r_c$ for prompt alignment, $r_m$ for visual rationality, and $r_q$ for human preference. The full procedure is detailed in Algorithm \ref{alg:reward_guidance}.

\begin{algorithm}[h]
\caption{\textsc{DeGu}}
\label{alg:reward_guidance}
\textbf{Input}: Inpainting model $\epsilon_\theta$, text condition $c$, masked latent $z^m$, mask $M$, text alignment reward model $r_c$, visual rationality reward model $r_m$, human preference reward model $r_q$, decoder $\mathcal{D}$, timesteps $T$, guidance scales $\gamma_c$, $\gamma_m$, and $\gamma_q$ \\
\textbf{Output}: Inpainted Image $I^o$
\begin{algorithmic}[1]
\STATE Obtain $z_T$ from Algorithm \ref{alg:noise_opt};
\FOR{$t = T$ to $1$}
    \STATE $\hat\epsilon_t \leftarrow \epsilon_\theta(z_t, t, c, z^m, \text{down}(M))$     \COMMENT{Predict noise}
    \STATE $\mathcal{R}(z_t, c, z^m) \leftarrow \gamma_c r_c(z_t, c) + \gamma_m r_m(z_t, z^m) + \gamma_q r_q(z_{t}, c)$;
    \STATE $\hat\epsilon_t \leftarrow \hat\epsilon_t - \sqrt{\bar{\alpha}_t} \nabla_{z_{t}}\mathcal{R}(z_t, c, z^m)$;
    \STATE $z_{t-1} \leftarrow \text{DenoiseStep}(z_t, t, \hat\epsilon_t);$
\ENDFOR
\STATE $I^o \leftarrow \mathcal{D}(z_0) \odot M + \mathcal{D}(z^m) \odot (1-M)$;    \COMMENT{Blending}
\STATE \textbf{return} $I^o$
\end{algorithmic}
\end{algorithm}

\section{Experimental Setup}

\subsection{Implementation Details}

\subsubsection{General Settings}
All experiments are conducted on a single NVIDIA A100 GPU. For all models, the inpainting strength is set to 1.0 to ensure accurate generation within the masked region. The conditioning scale for BN and SD3I is set to 1.0, following their official recommendations. All hyperparameters listed below are kept identical for experiments on both the EditBench (free-form masks) and MSCOCO (layout masks) datasets.

\subsubsection{Hyperparameters}
Our hyperparameter strategy is designed for both robustness and flexibility. A substantial portion of parameters is shared across all models, establishing a robust foundation for our method. For parameters that are model-specific, we group them based on the underlying backbone, ensuring tailored optimization for each model.

Firstly, we introduce the shared parameters. The following PriNo parameters are kept constant for all base models, demonstrating the core stability of our noise optimization strategy. We fix the cross-attention loss weight $\lambda_1=1$. The self-attention loss weight is $\lambda_2=5$ and the sensitivity analysis is presented in Appendix \ref{sec:Hyperparameter-Sensitivity}. The KL divergence weight is dynamically adjusted as $\lambda_3 = 500 + \mathcal{L}_\text{KL} \times 10^6$ to constrain the optimized noise such that it remains close to the initial Gaussian noise. We also use shared early-stopping thresholds ($\tau_\text{c}=0.1, \tau_\text{s}=0.1, \tau_\text{KL}=0.003$) and an iteration limit $\tau_\text{iter}=40$ per round.

Secondly, we introduce the model-specific parameters. To accommodate the diverse architectures and scales of modern diffusion models, we define three sets of parameters for the remaining hyperparameters, as detailed in Tab. \ref{tab:hyperparameters}. This principled adaptation highlights the flexibility of our framework:
\begin{itemize}
\item SD1.5-based models: This group (SDI, PPT, BN) shares a largely consistent set of parameters due to their similar U-Net architecture.
\item SDXL-based models: As a significantly larger U-Net model, SDXLI benefits from a reduced number of optimization rounds to maintain efficiency.
\item DiT-based models: For the state-of-the-art DiT-based SD3I, the guidance scales and optimization rounds are adapted to its unique Transformer architecture, demonstrating our method's applicability to the latest models.
\end{itemize}

Additionally, the denoising steps and the Classifier-Free Guidance (CFG) scale for each backbone follow their officially recommended settings and are also listed in Tab. \ref{tab:hyperparameters}.

\begin{table}[t]
  \centering
  \begin{adjustbox}{max width=0.47\textwidth}
    \begin{tabular}{lcccccc}
      \toprule
      \textbf{Backbone} & \textbf{Model} & $\tau_{\text{round}}$ & $lr$ & $(\gamma_c, \gamma_m, \gamma_q)$ & \small{\# Steps} & \small{CFG Scale} \\ \midrule
      \multirow{3}{*}{SD1.5-based}  & SDI   & 5    & 0.01 & $25 \times (4.0, 1, 0.1)$      & 50               & 7.5               \\
                                    & PPT   & 5    & 0.1  & $10 \times (3.5, 1, 0.1)$      & 50               & 7.5               \\
                                    & BN    & 5    & 0.1  & $25 \times (4.0, 1, 0.1)$      & 50               & 7.5               \\ \midrule
      SDXL-based                    & SDXLI & 3    & 0.01 & $15 \times (3.5, 1, 0.1)$      & 20               & 7.5               \\ \midrule
      DiT-based                     & SD3I  & 1    & 0.1  & $40 \times (4.0, 1, 0.2)$      & 28               & 7                 \\ \bottomrule
    \end{tabular}
  \end{adjustbox}
\caption{Model-specific hyperparameters in our \emph{FreeInpaint}. Models are grouped by their backbone.}
  \label{tab:hyperparameters}
\end{table}

\subsection{Re-annotated MSCOCO Dataset}
The original object labels in the MSCOCO validation set are single words, which are insufficient for evaluating nuanced prompt alignment in inpainting scenarios. To create a more suitable benchmark, we re-annotate a subset of the data. The process involved cropping each object using its segmentation mask and using LLaVA-NeXT to generate a detailed caption for the cropped region. From the generated results, we curate a final set of 300 high-quality image-mask-description triplets by selecting those with the highest joint scores from ImageReward and local CLIPScore. This re-annotated dataset provides a more realistic testbed for evaluating the prompt-following capabilities of inpainting models.

\section{Experimental Results}
\subsection{Hyperparameter Sensitivity}  \label{sec:Hyperparameter-Sensitivity}
\begin{figure}[t]
\centering
  \includegraphics[width=\linewidth]{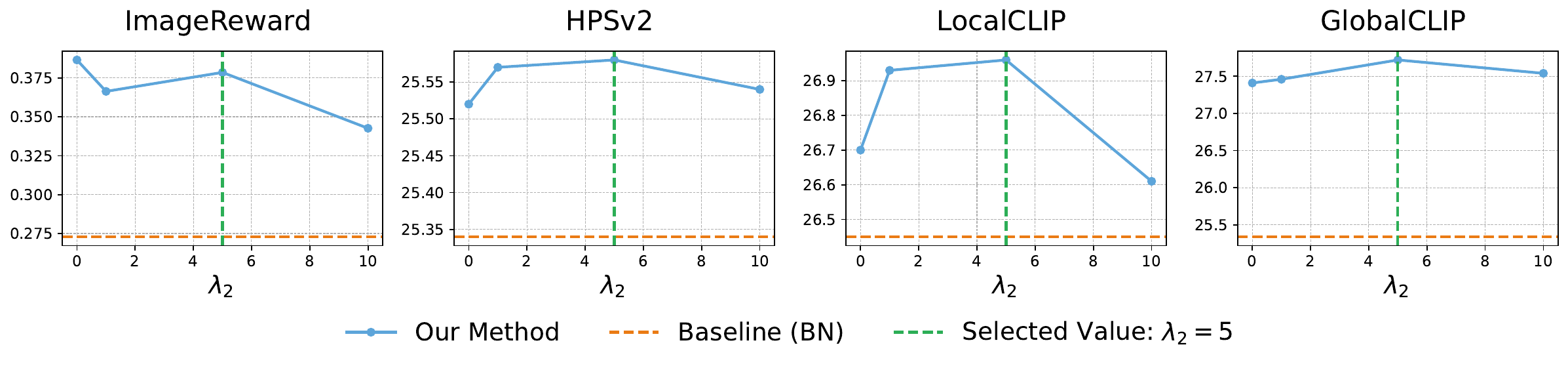}
  \caption{The sensitivity analysis of the $\mathcal{L}_\text{s}$ weight $\lambda_2$.}
  \label{fig:hyperparameters-lambda2}
\end{figure}

We analyze the sensitivity of PriNo's loss weights by fixing cross-attention $\lambda_1=1$ and varying self-attention $\lambda_2$. To isolate PriNo, DeGu is disabled, and only prompt-related metrics are reported. Fig. \ref{fig:hyperparameters-lambda2} shows that $\lambda_2=5$ offers an optimal trade-off. Despite a minor gap in ImageReward, it achieves the best results on other key metrics. Moreover, PriNo proves robust, significantly outperforming the baseline across all examined $\lambda_2$ values.

\begin{figure}[t]
  \includegraphics[width=\linewidth]{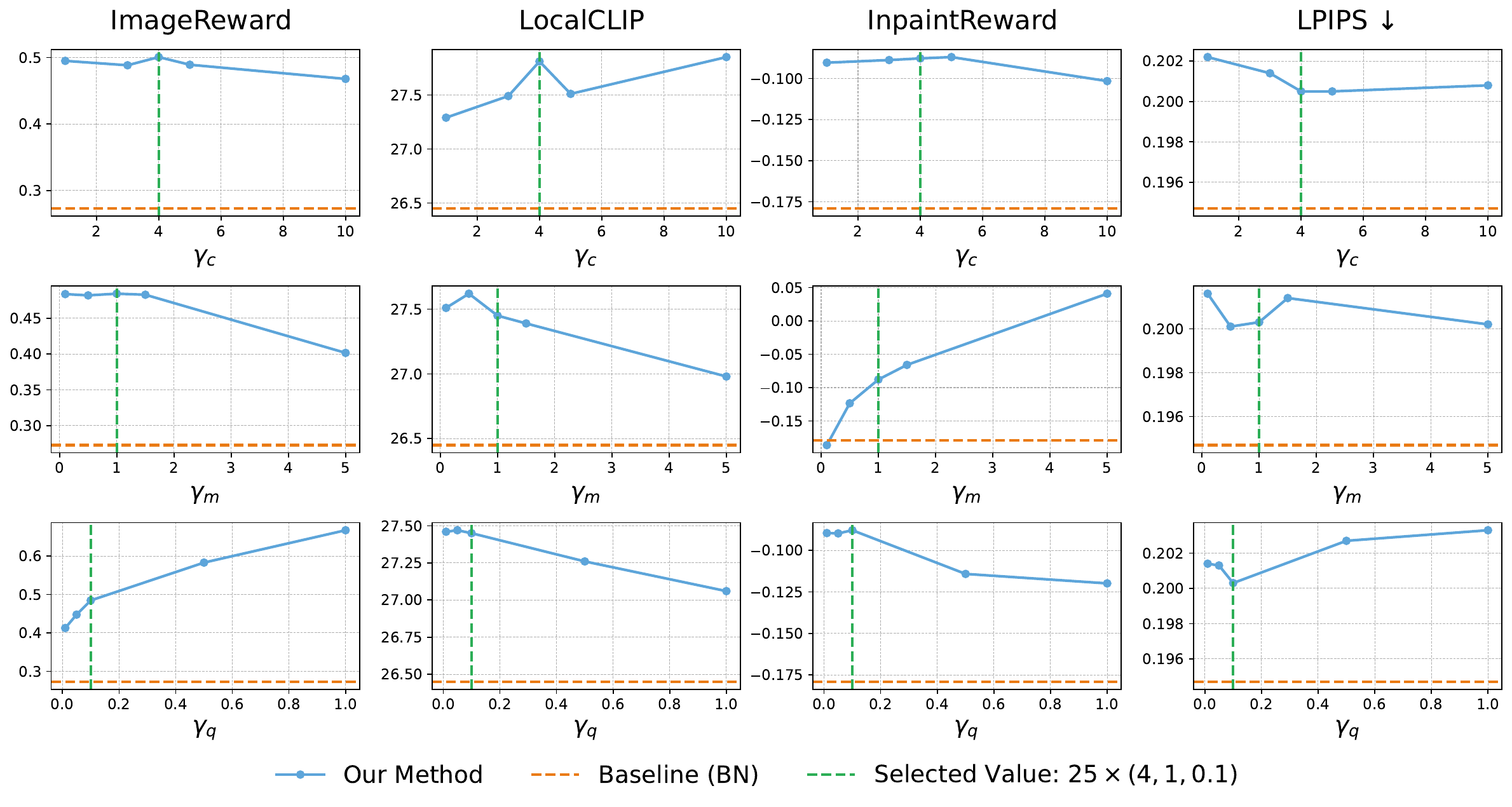}
  \caption{The sensitivity analysis of the guidance weights $\gamma_c$, $\gamma_m$, and $\gamma_q$.}
  \label{fig:hyperparameters-guidance}
\end{figure}

Simultaneously, we conduct a sensitivity analysis on the guiding weights in DeGu, as shown in Fig. \ref{fig:hyperparameters-guidance}. This result validates our method's robustness to hyperparameters, consistently outperforming the baseline across a wide range of values. Our selected weights, marked in the figure, balance competing objectives like prompt alignment (LocalCLIP), visual rationality (InpaintReward, LPIPS), and human preference (ImageReward).

\subsection{Efficiency Analysis}
The computational cost of \emph{FreeInpaint} is flexible, controlled by the noise optimization rounds $\tau_{\text{round}}$ and guidance frequency $s_\text{guide}$ (i.e., applying DeGu guidance once every $s_\text{guide}$ denoising steps). To demonstrate this adjustability, we evaluate an efficiency-focused setting ($\tau_{\text{round}}=1$ and $s_\text{guide}=10$) on EditBench using a single A100 GPU. As shown in Tab. \ref{tab:efficiency}, this configuration achieves a runtime comparable to HDP with lower memory usage, while delivering superior performance over both the BN baseline and HDP. This result confirms \emph{FreeInpaint}'s highly favorable trade-off between inference speed and output quality, allowing users to balance performance against their specific computational budget.

\begin{table}[t]
\begin{adjustbox}{max width=0.47\textwidth}
\begin{tabular}{lcccccccc}
\toprule
        & Time (s)      & Mem (GB)      & ImageReward     & HPSv2          & Local CLIP     & Global CLIP    & InpaintReward    & LPIPS $\downarrow$  \\ \midrule
BN      & 2.37 & 6.38 & 0.2729          & 25.34          & 26.45          & 27.18          & -0.1791          & \textbf{0.1947} \\
BN+HDP  & 6.51          & 13.82         & 0.3836          & 25.20          & 27.08          & 27.69          & -0.2124          & 0.2135          \\
BN+Ours & 7.86          & 12.24         & \textbf{0.4488} & \textbf{25.60} & \textbf{27.20} & \textbf{28.07} & \textbf{-0.1075} & 0.2011          \\ \bottomrule
\end{tabular}
\end{adjustbox}
\caption{Efficiency and performance analysis using an A100 GPU. Our method is evaluated in an efficiency-focused setting ($\tau_{\text{round}}=1$ and $s_\text{guide}=10$). \textbf{Bold} indicates the best.}
\label{tab:efficiency}
\end{table}

\subsection{Qualitative Results}
\begin{figure*}[!t]
    \centering
    \renewcommand{\arraystretch}{0.4} 
    \setlength{\tabcolsep}{0.5pt} 

    \begin{tabular}{
        >{\centering\arraybackslash}b{0.03\linewidth} 
        >{\centering\arraybackslash}p{0.12\linewidth} 
        >{\centering\arraybackslash}p{0.12\linewidth} 
        >{\centering\arraybackslash}p{0.12\linewidth} 
        >{\centering\arraybackslash}p{0.12\linewidth} 
        >{\centering\arraybackslash}p{0.12\linewidth}}

        & A yellow bird diving towards coral
        & A silver llama is standing
        & A black and yellow nylon flag
        & A black cat with white chest
        & A clock, white face with black numbers \\

        \rotatebox{90}{Input}
        & \includegraphics[width=\linewidth]{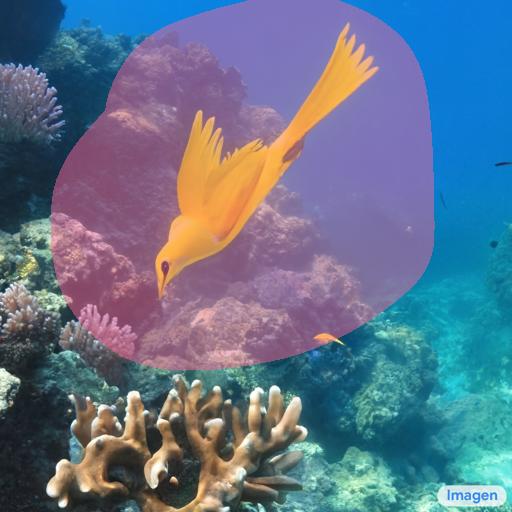}
        & \includegraphics[width=\linewidth]{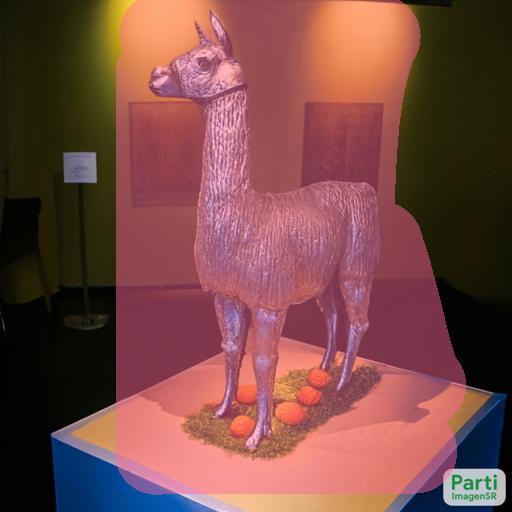}
        & \includegraphics[width=\linewidth]{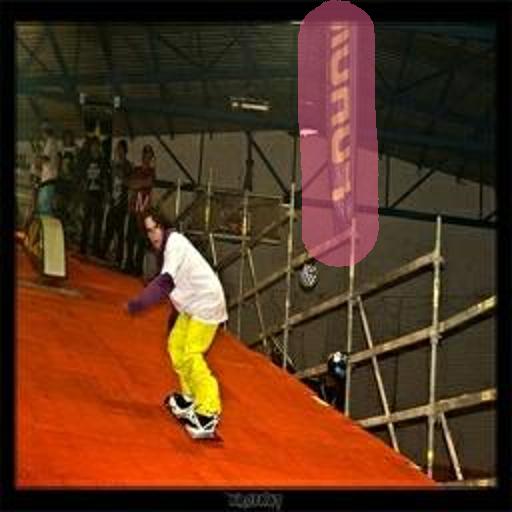}
        & \includegraphics[width=\linewidth]{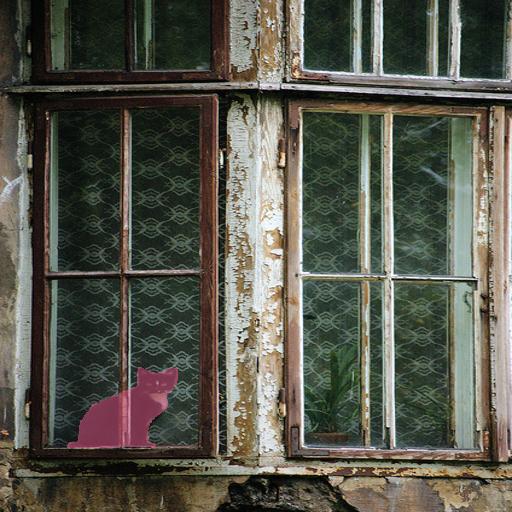}
        & \includegraphics[width=\linewidth]{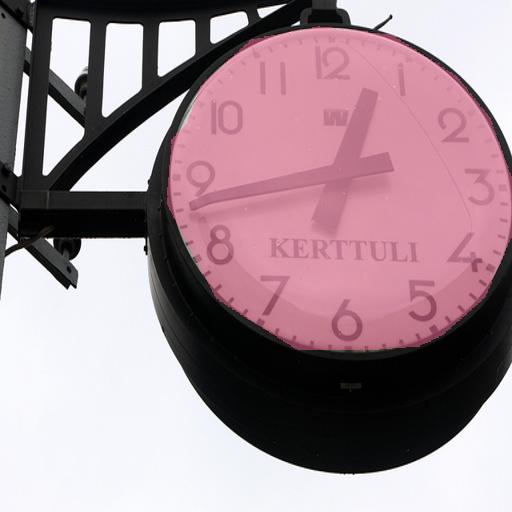} \\

        \rotatebox{90}{PPT}
        & \includegraphics[width=\linewidth]{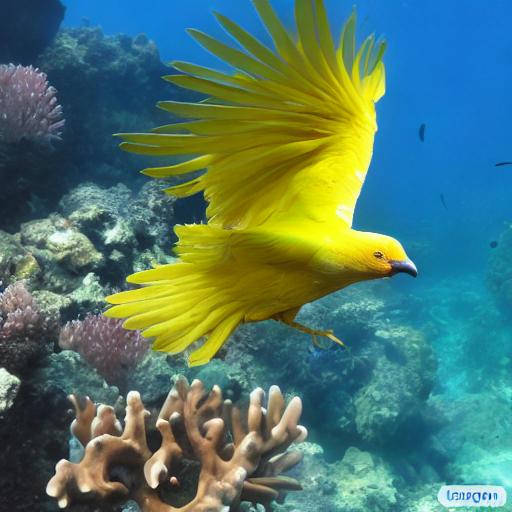}
        & \includegraphics[width=\linewidth]{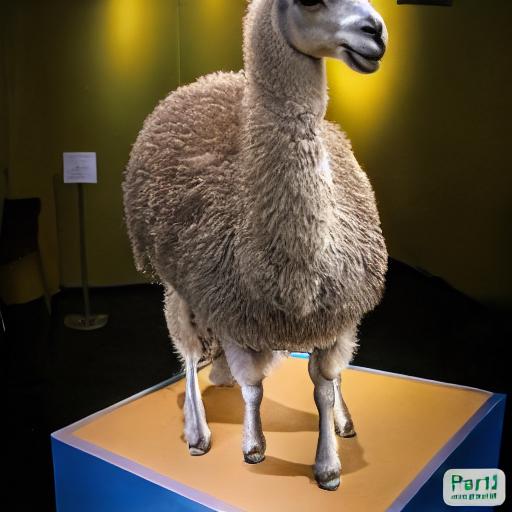}
        & \includegraphics[width=\linewidth]{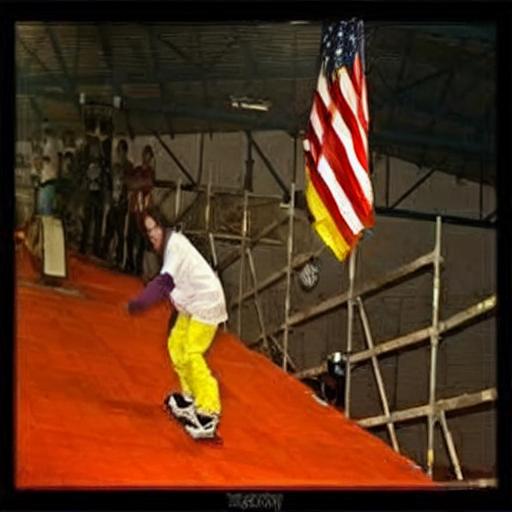}
        & \includegraphics[width=\linewidth]{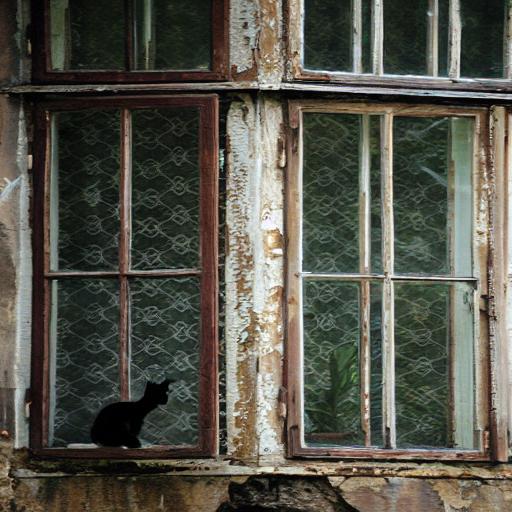}
        & \includegraphics[width=\linewidth]{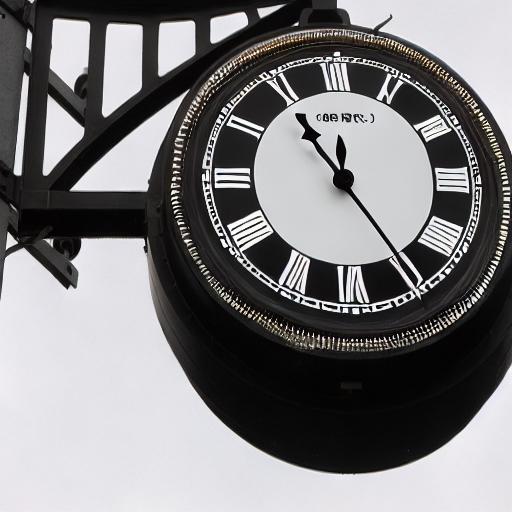} \\

        \rotatebox{90}{PPT+HDP}
        & \includegraphics[width=\linewidth]{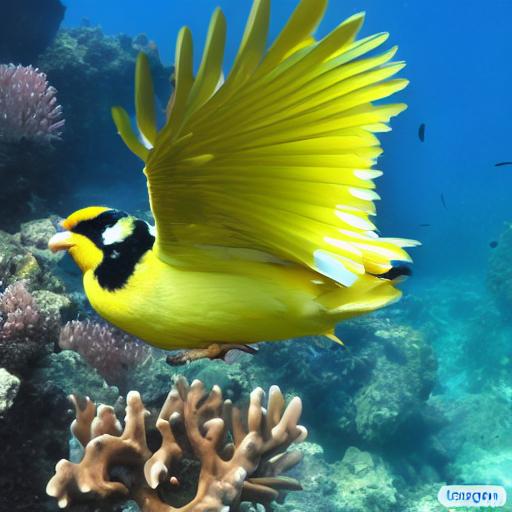}
        & \includegraphics[width=\linewidth]{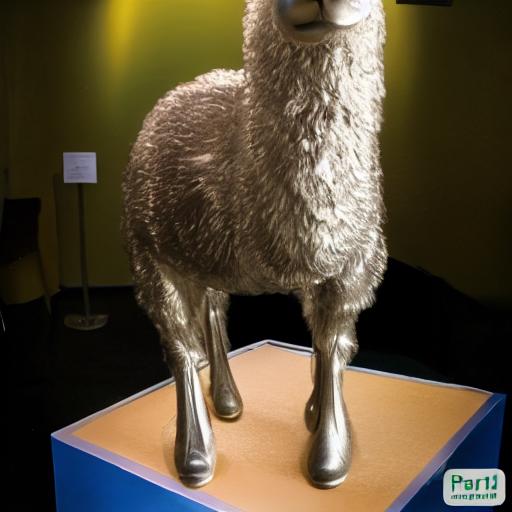}
        & \includegraphics[width=\linewidth]{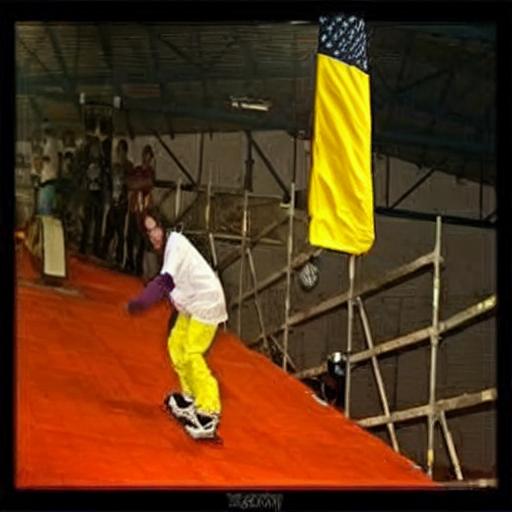}
        & \includegraphics[width=\linewidth]{images/qualitative/coco_custom/186_000000416170-ppt-hdp.jpg}
        & \includegraphics[width=\linewidth]{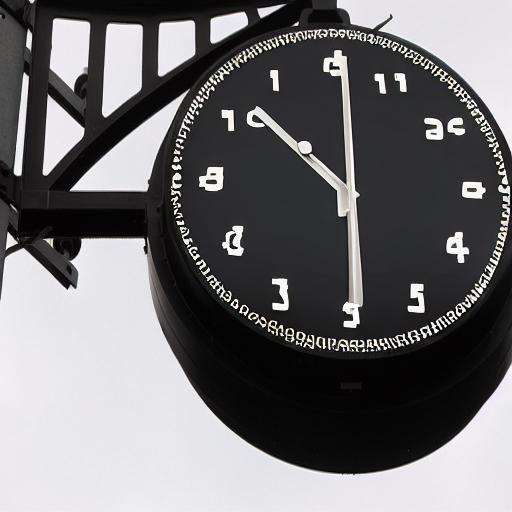} \\

        \rotatebox{90}{PPT+Ours}
        & \includegraphics[width=\linewidth]{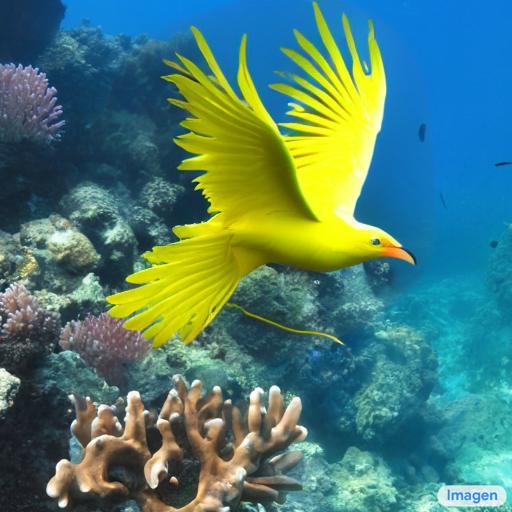}
        & \includegraphics[width=\linewidth]{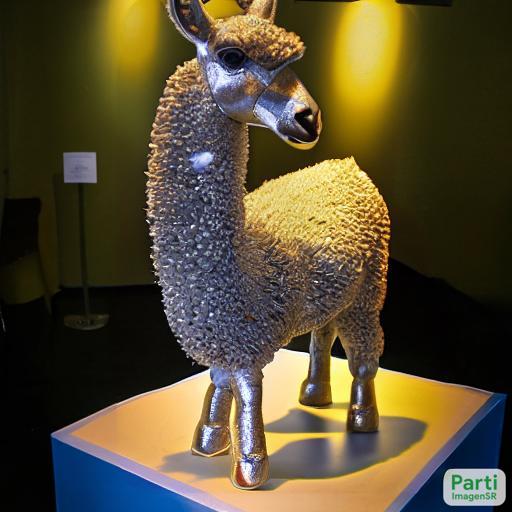}
        & \includegraphics[width=\linewidth]{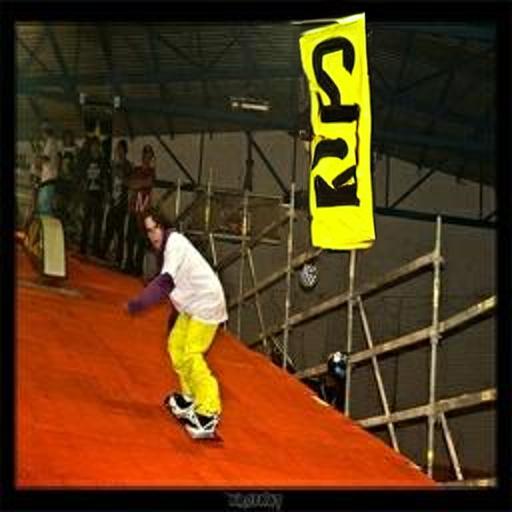}
        & \includegraphics[width=\linewidth]{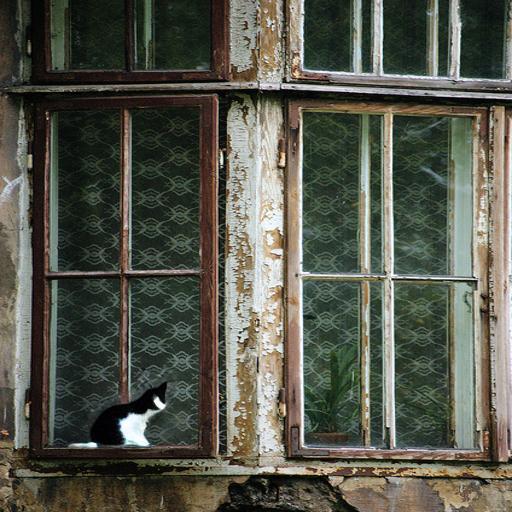}
        & \includegraphics[width=\linewidth]{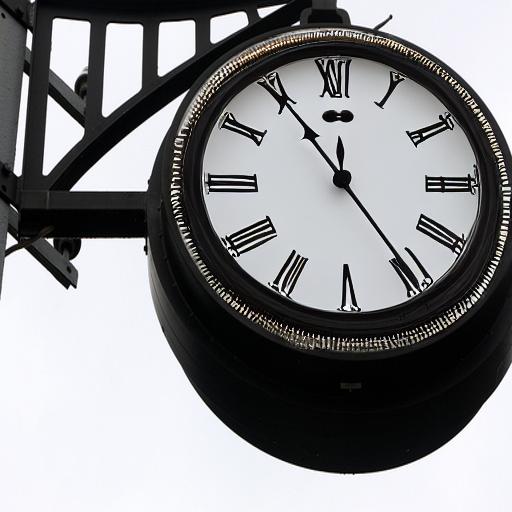} \\

    \end{tabular}
    \caption{Comparison against PPT-based approaches. Zoom in for a better view.}

    \label{fig:qualitative-ppt}
\end{figure*}

\begin{figure*}[!t]
    \centering
    \renewcommand{\arraystretch}{0.4} 
    \setlength{\tabcolsep}{0.5pt} 

    \begin{tabular}{
        >{\centering\arraybackslash}b{0.03\linewidth} 
        >{\centering\arraybackslash}p{0.12\linewidth} 
        >{\centering\arraybackslash}p{0.12\linewidth} 
        >{\centering\arraybackslash}p{0.12\linewidth} 
        >{\centering\arraybackslash}p{0.12\linewidth} 
        >{\centering\arraybackslash}p{0.12\linewidth}}

        & A yellow cloth sawfish doll
        & A blue monkey
        & A stack of white ceramic plates
        & A silver Pocket PC
        & A bowl in black and red \\

        \rotatebox{90}{Input}
        & \includegraphics[width=\linewidth]{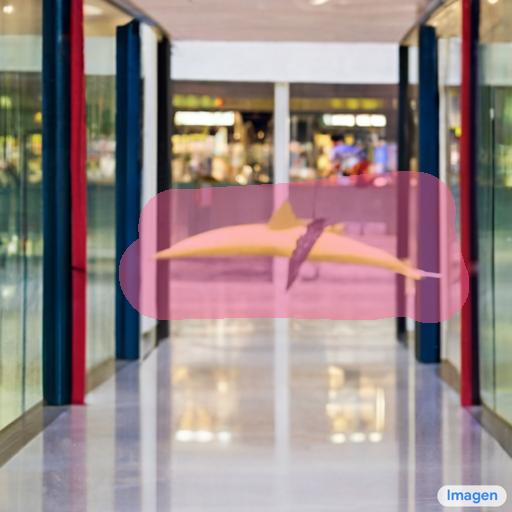}
        & \includegraphics[width=\linewidth]{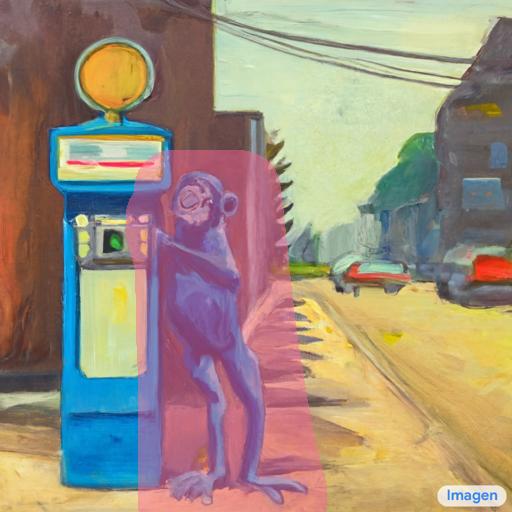}
        & \includegraphics[width=\linewidth]{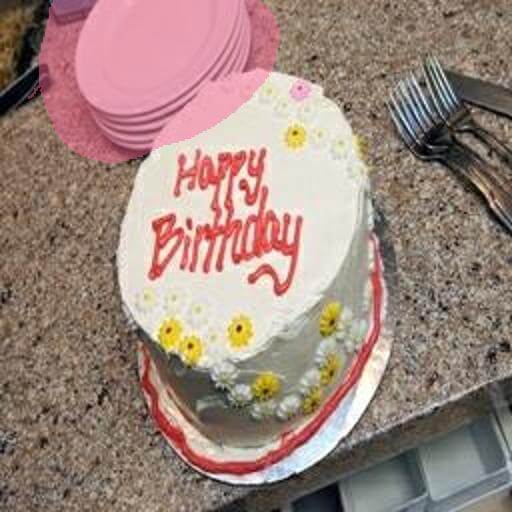}
        & \includegraphics[width=\linewidth]{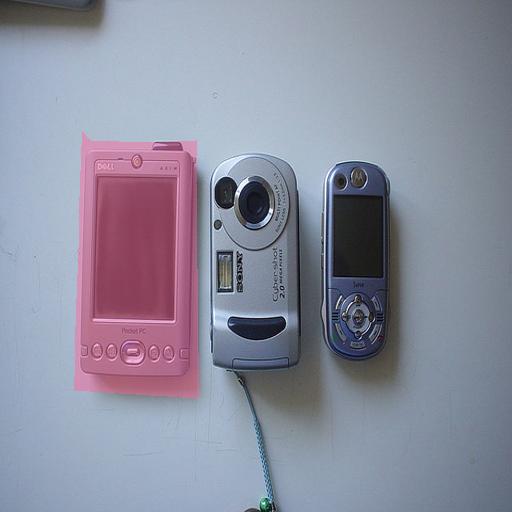}
        & \includegraphics[width=\linewidth]{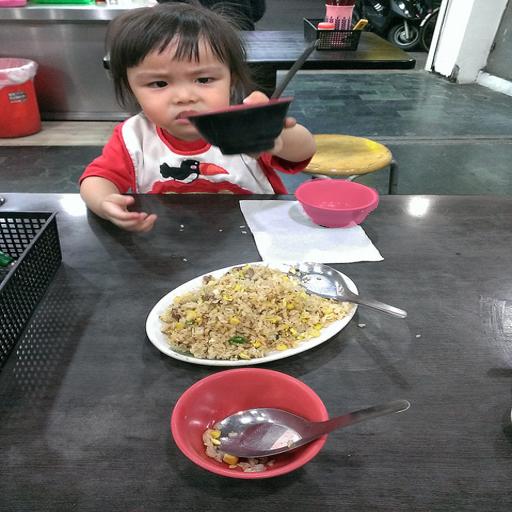} \\

        \rotatebox{90}{SDXLI}
        & \includegraphics[width=\linewidth]{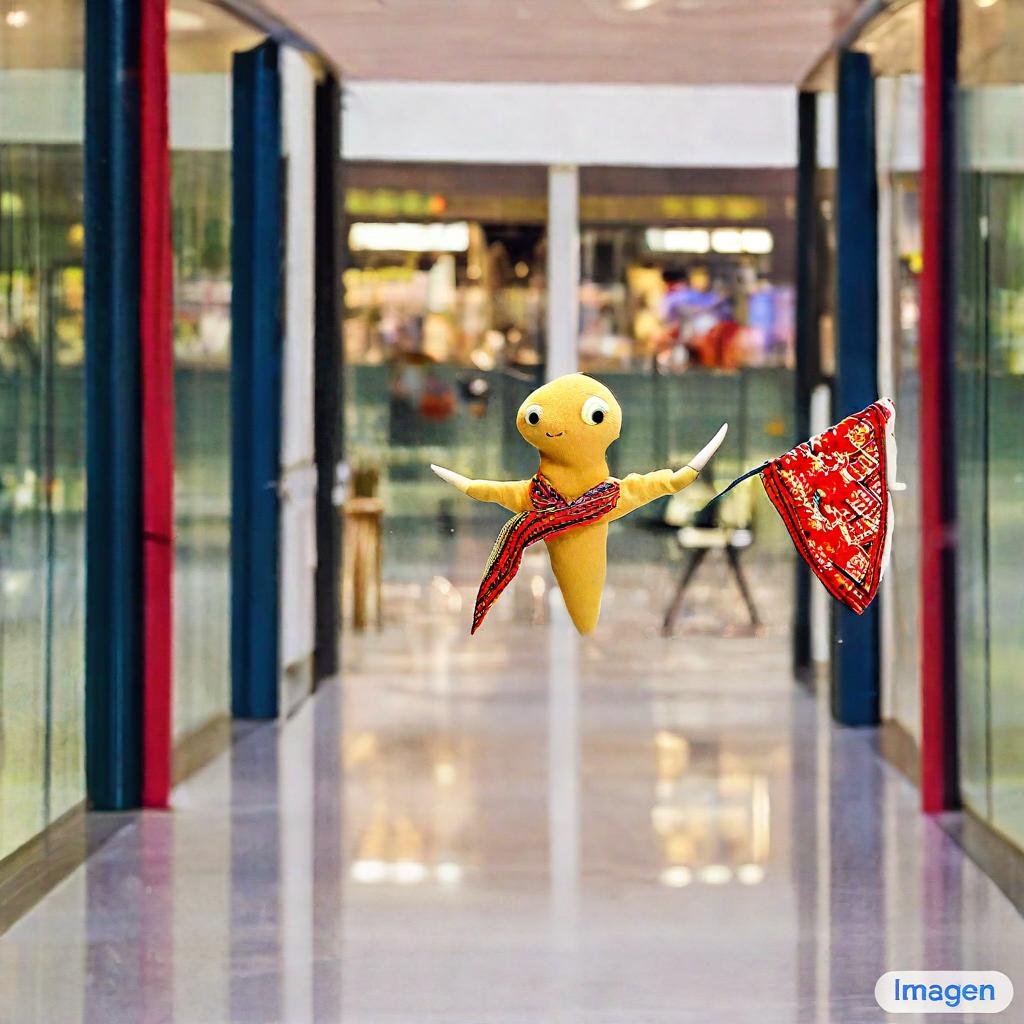}
        & \includegraphics[width=\linewidth]{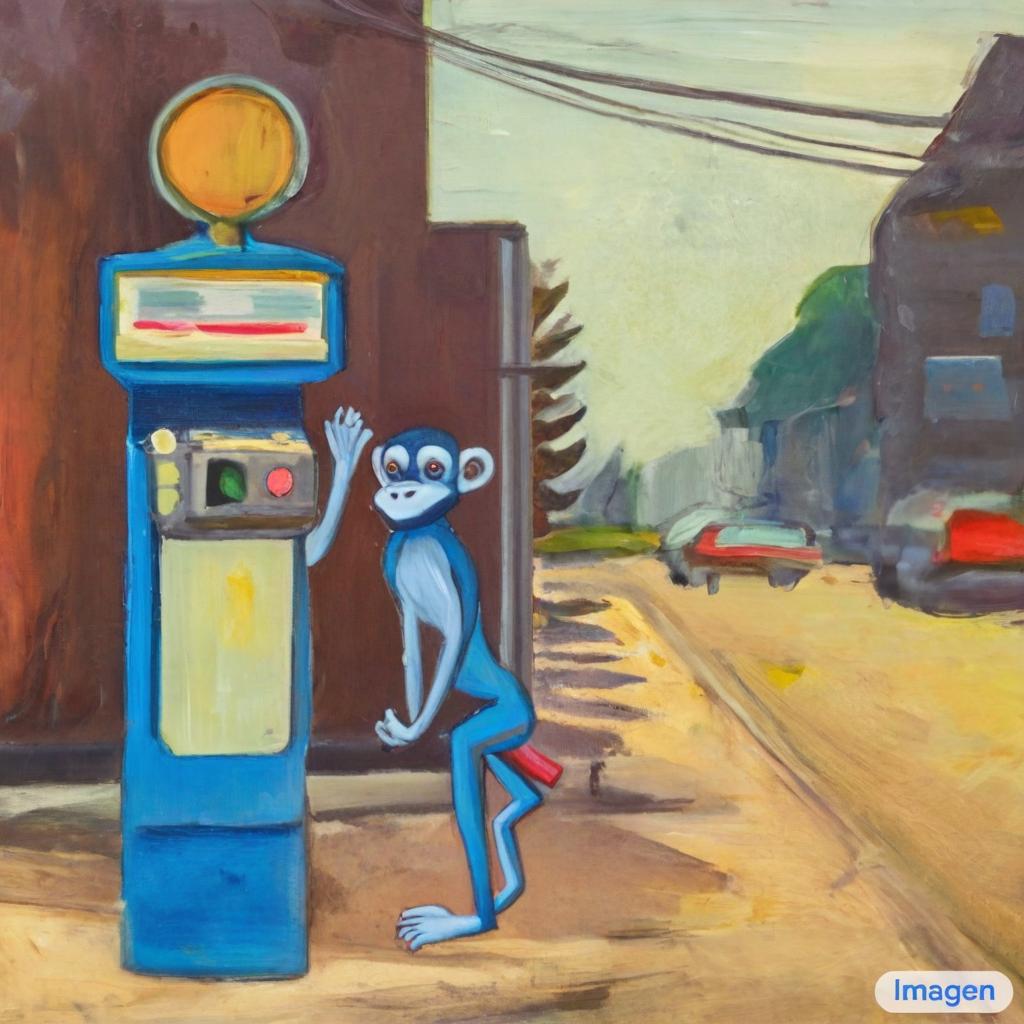}
        & \includegraphics[width=\linewidth]{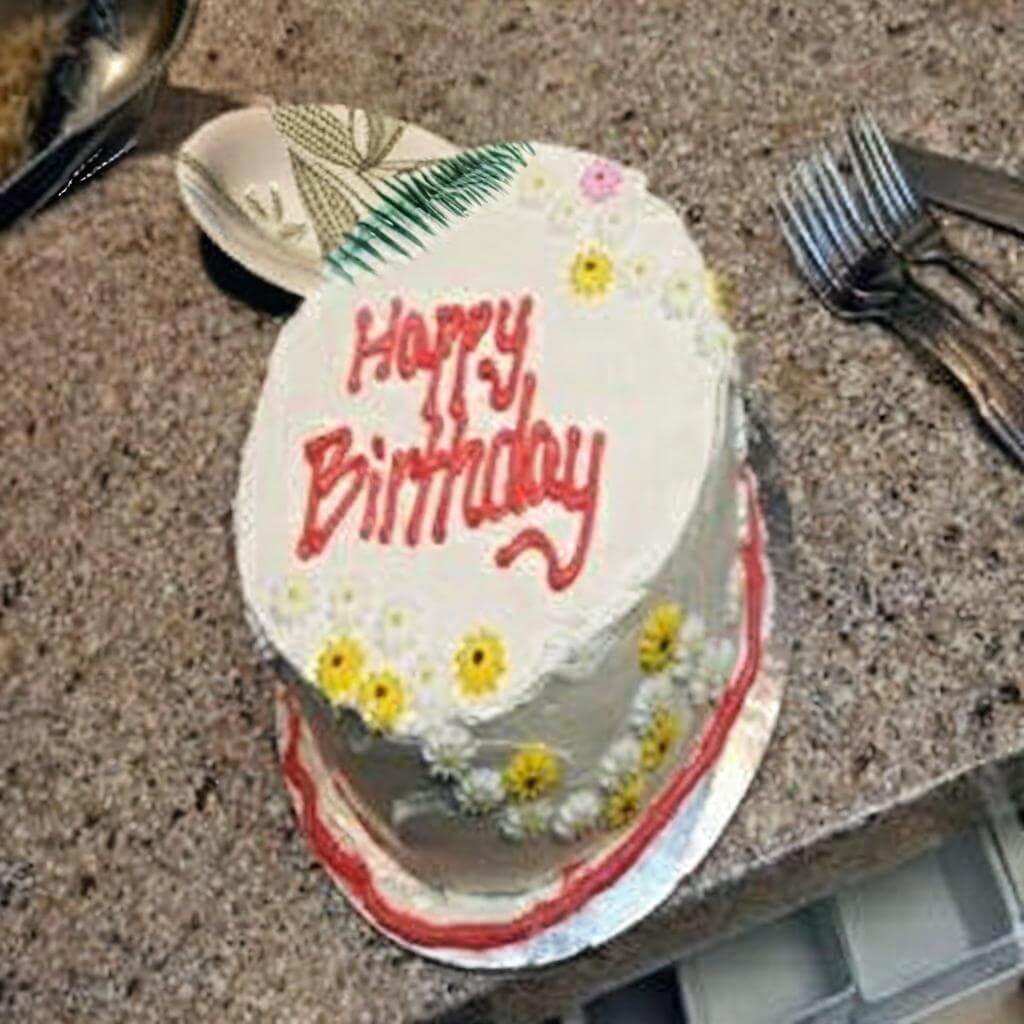}
        & \includegraphics[width=\linewidth]{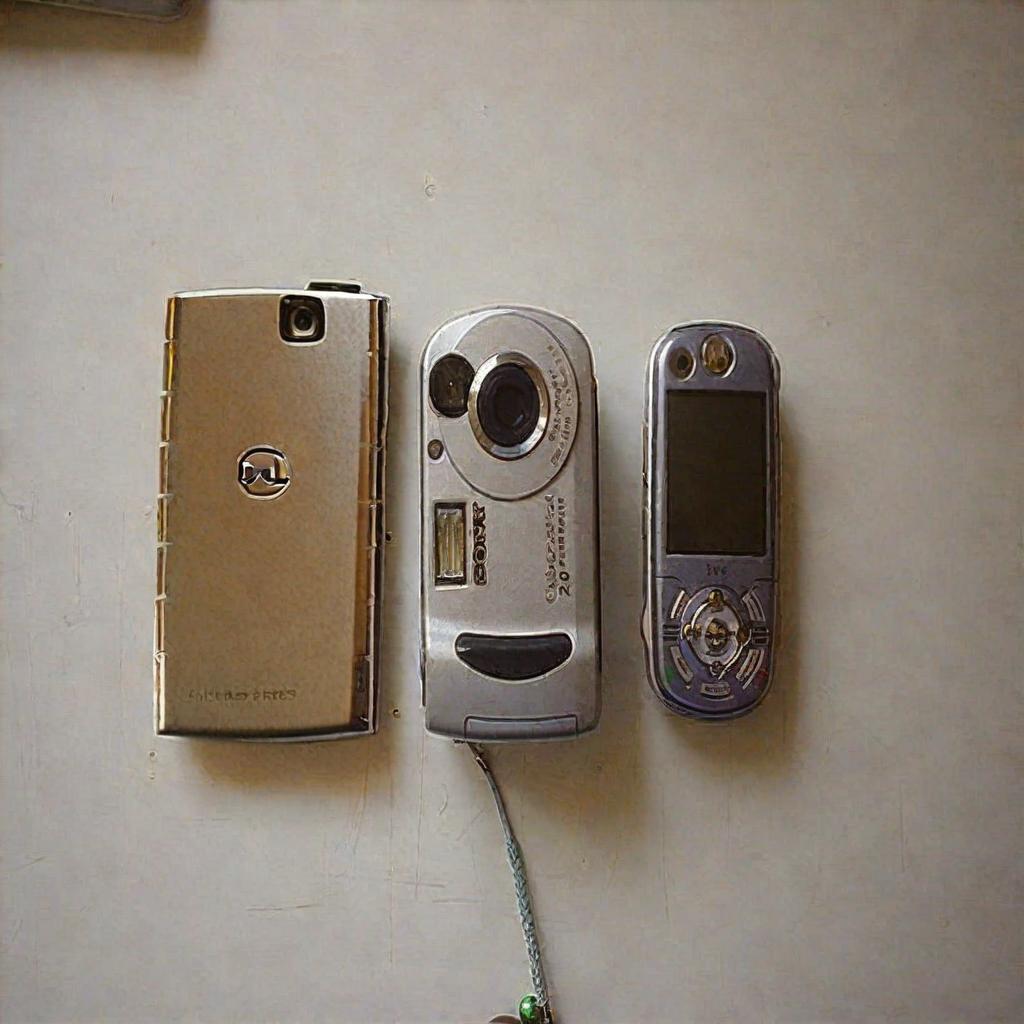}
        & \includegraphics[width=\linewidth]{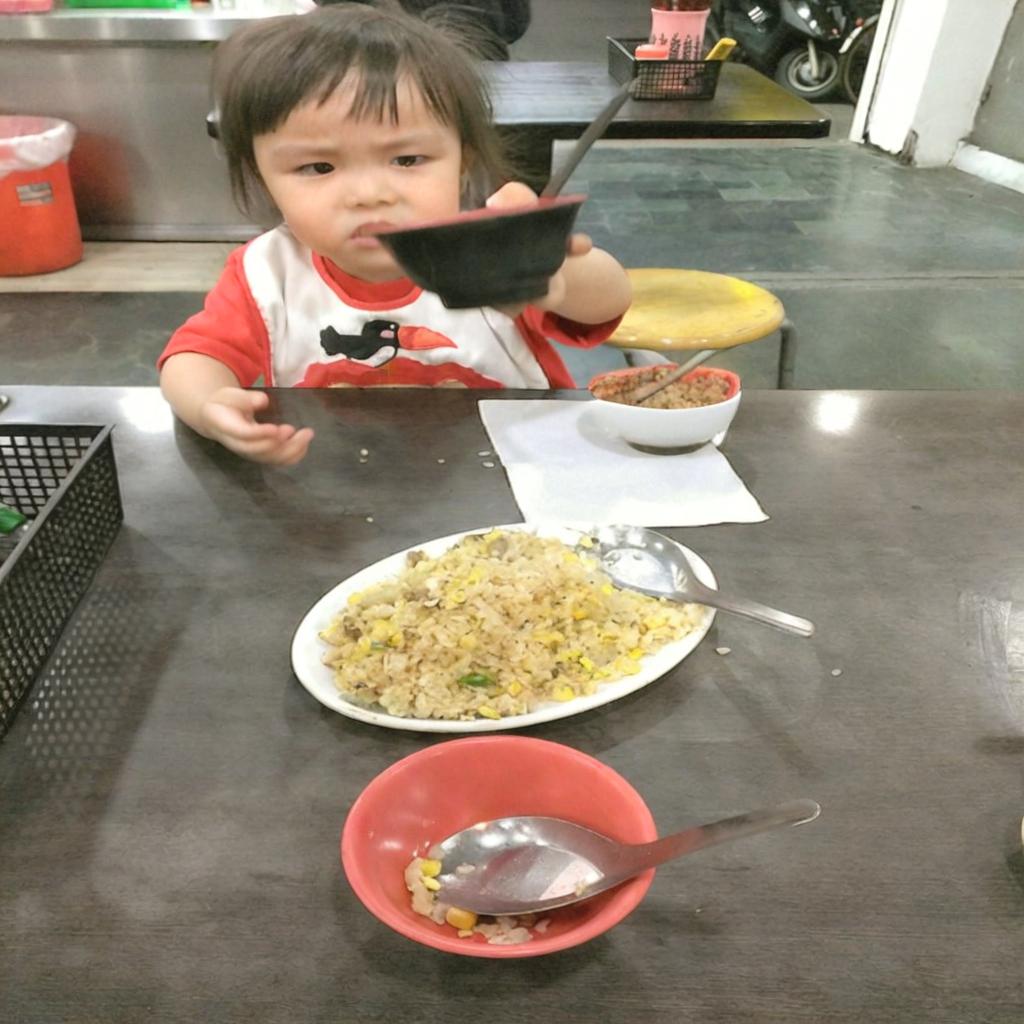} \\

        \rotatebox{90}{SDXLI+HDP}
        & \includegraphics[width=\linewidth]{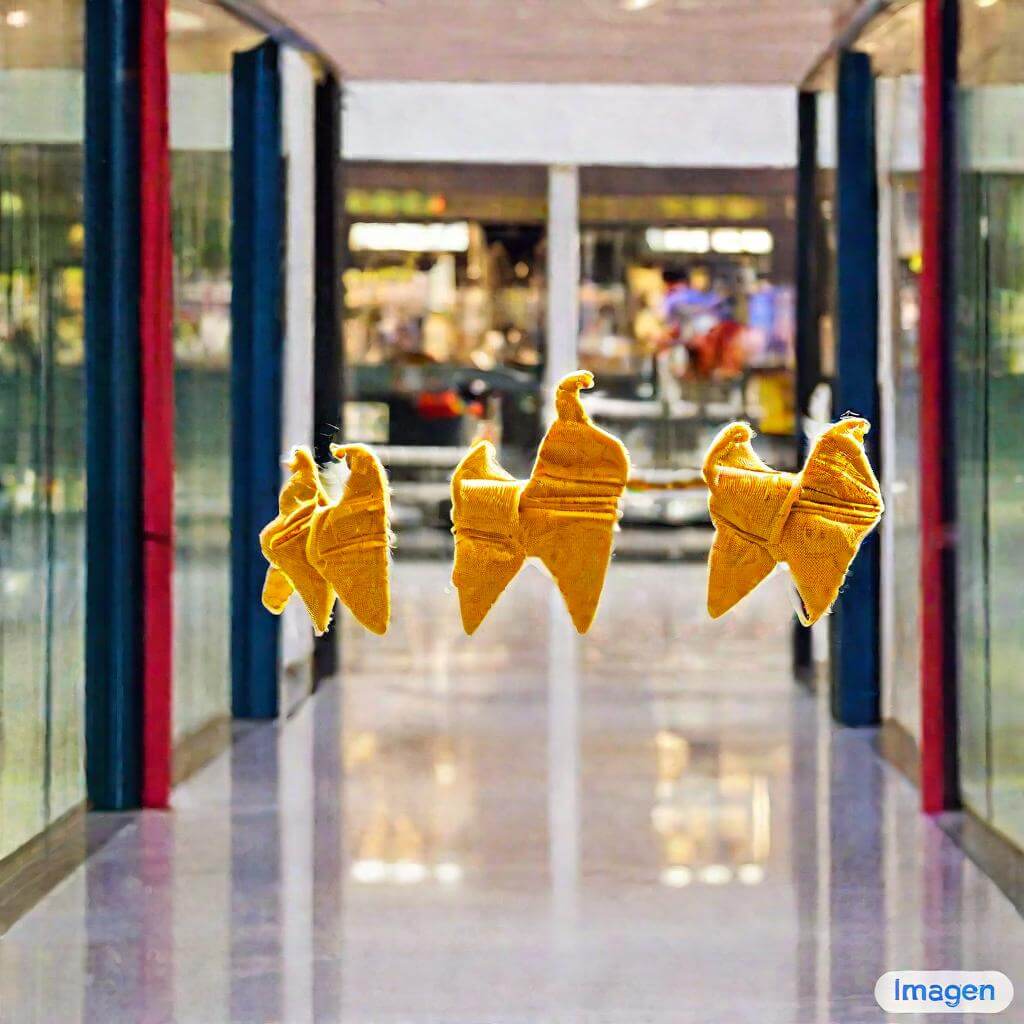}
        & \includegraphics[width=\linewidth]{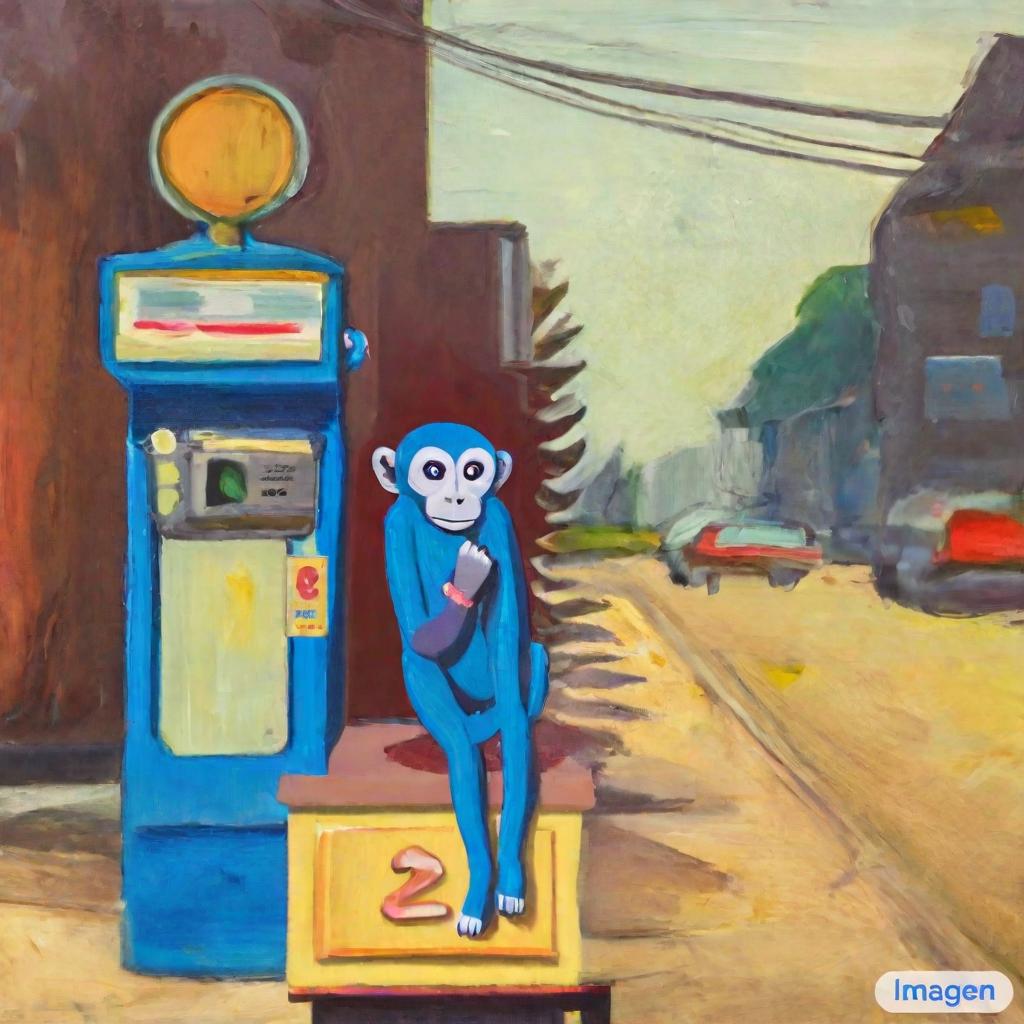}
        & \includegraphics[width=\linewidth]{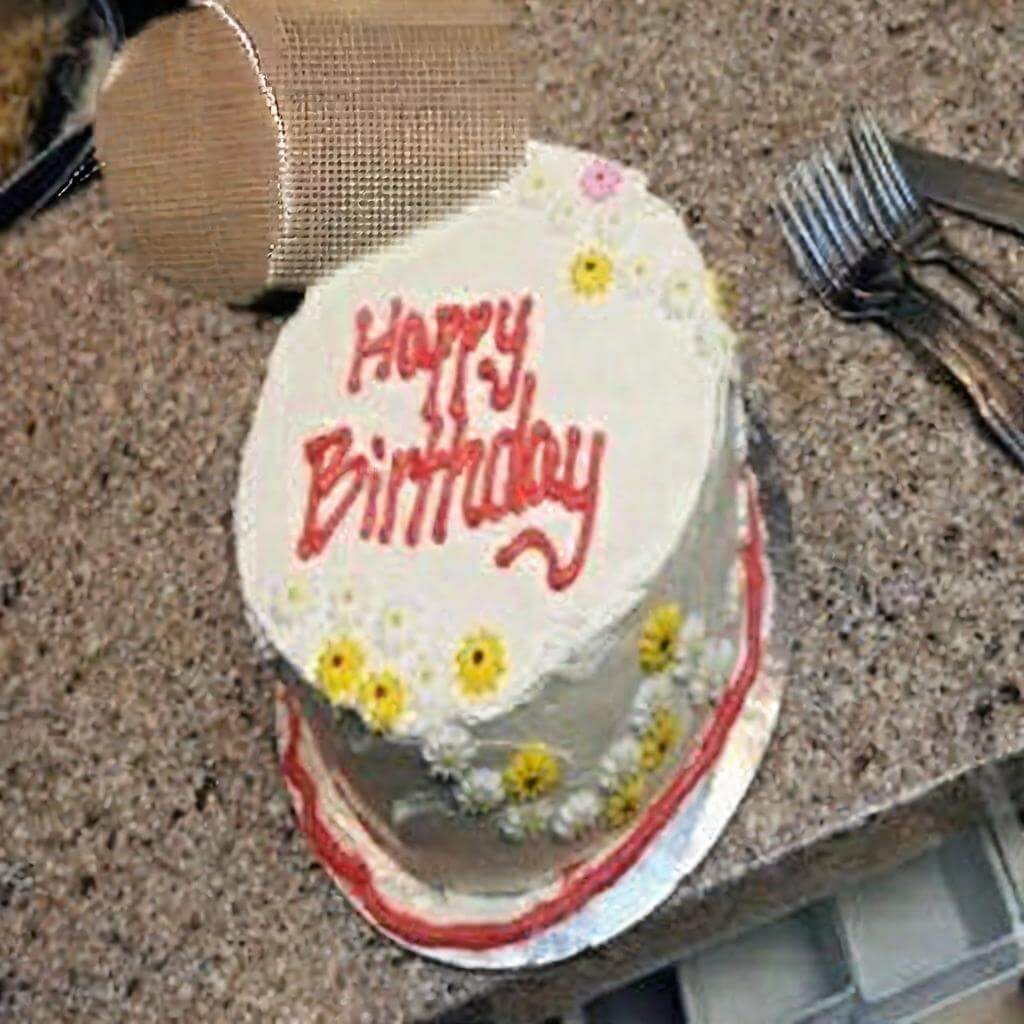}
        & \includegraphics[width=\linewidth]{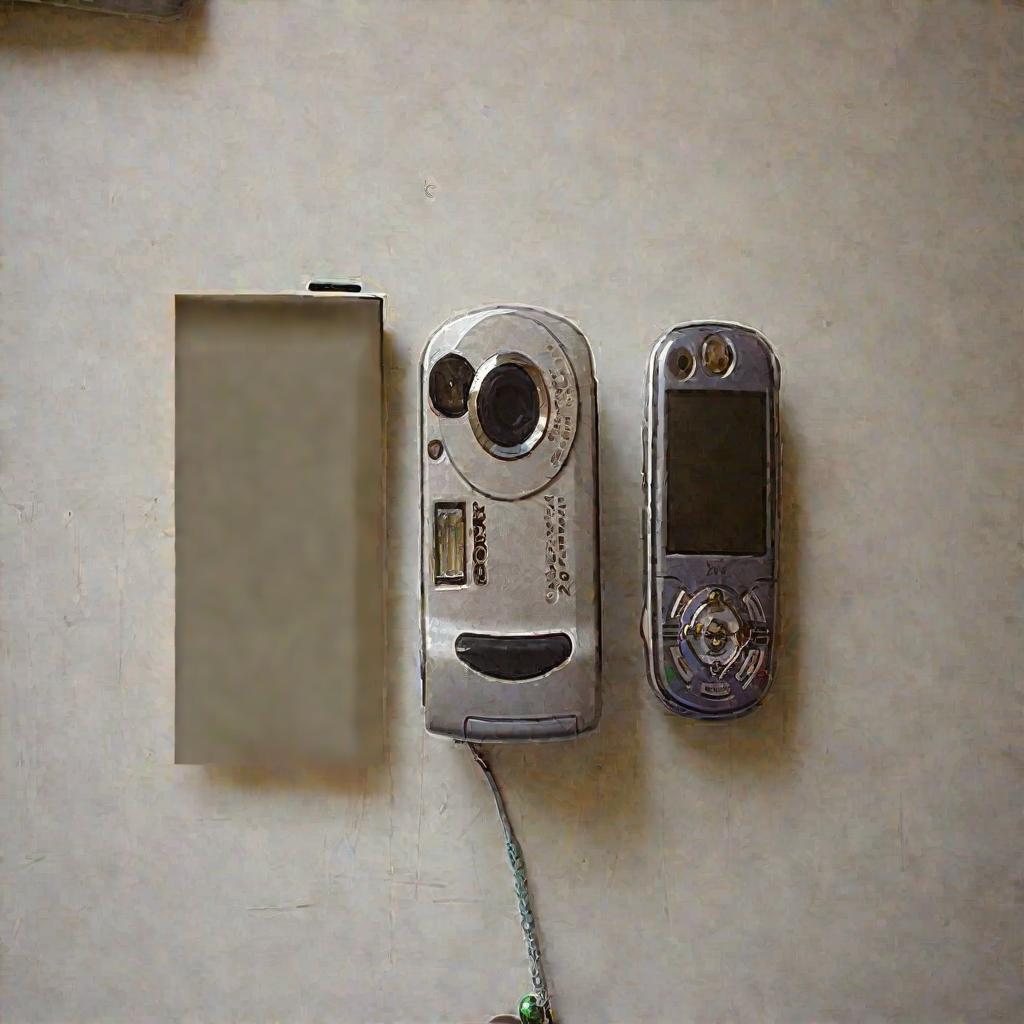}
        & \includegraphics[width=\linewidth]{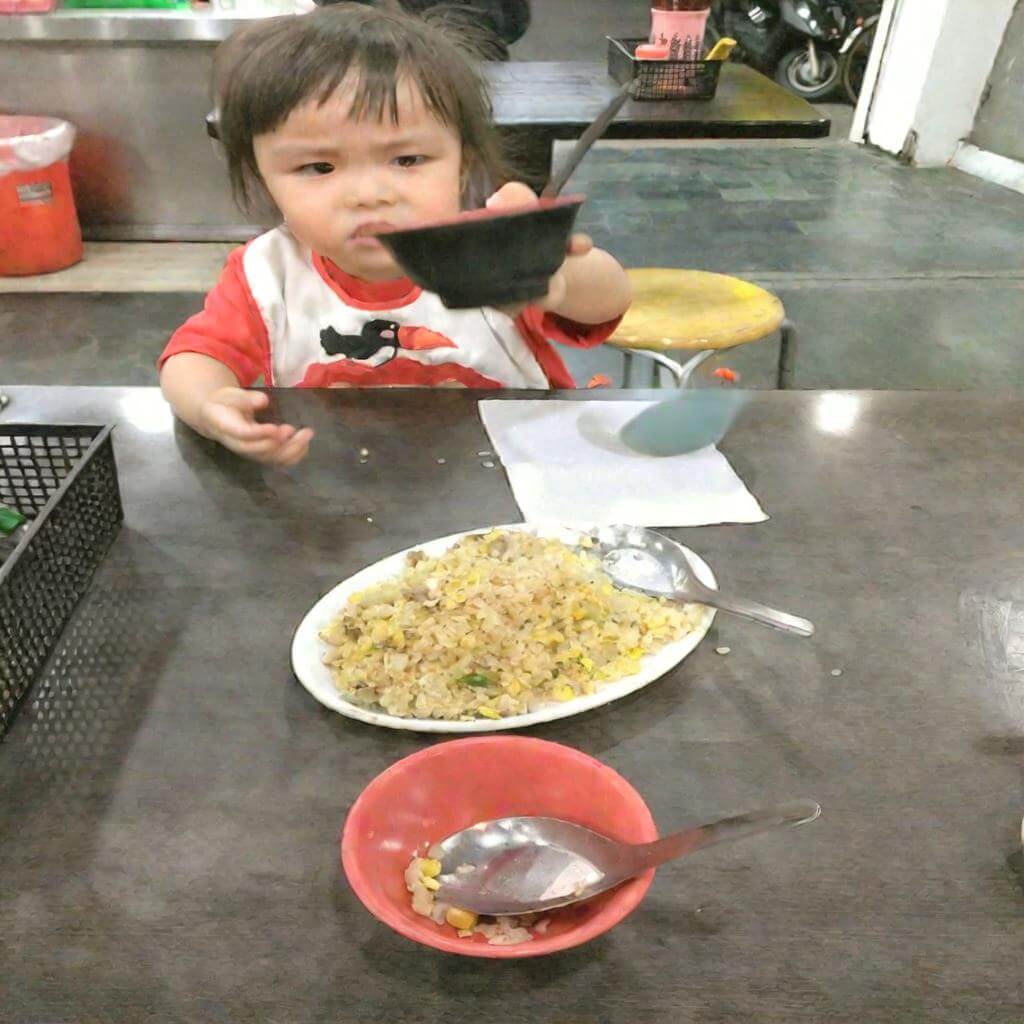} \\

        \rotatebox{90}{SDXLI+Ours}
        & \includegraphics[width=\linewidth]{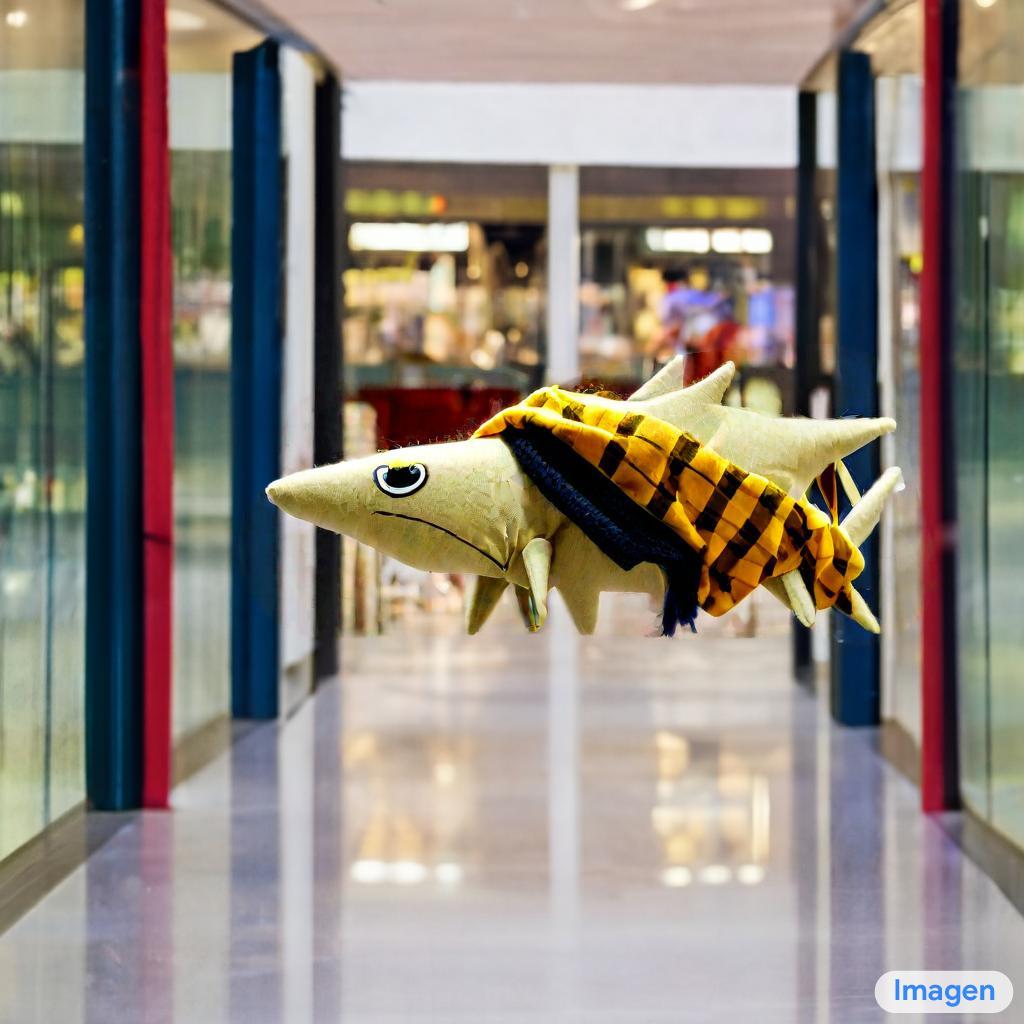}
        & \includegraphics[width=\linewidth]{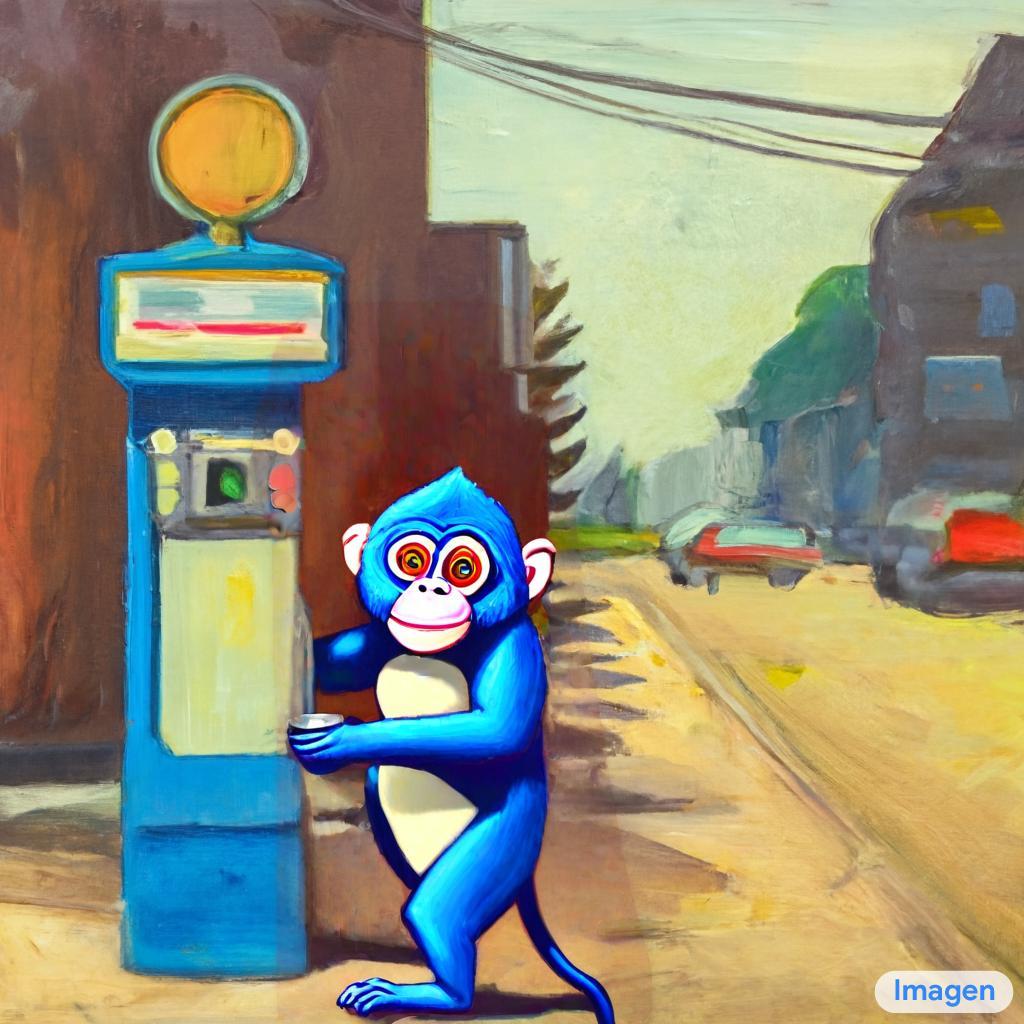}
        & \includegraphics[width=\linewidth]{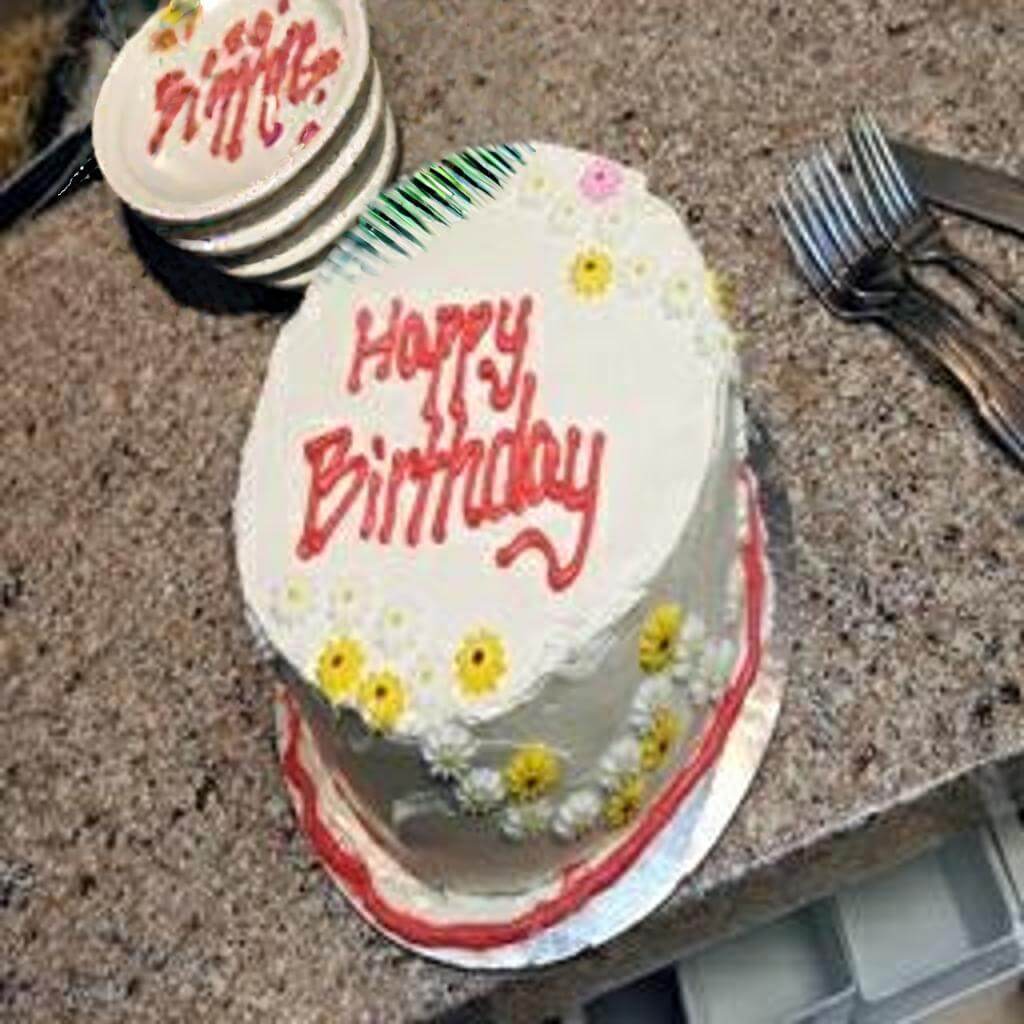}
        & \includegraphics[width=\linewidth]{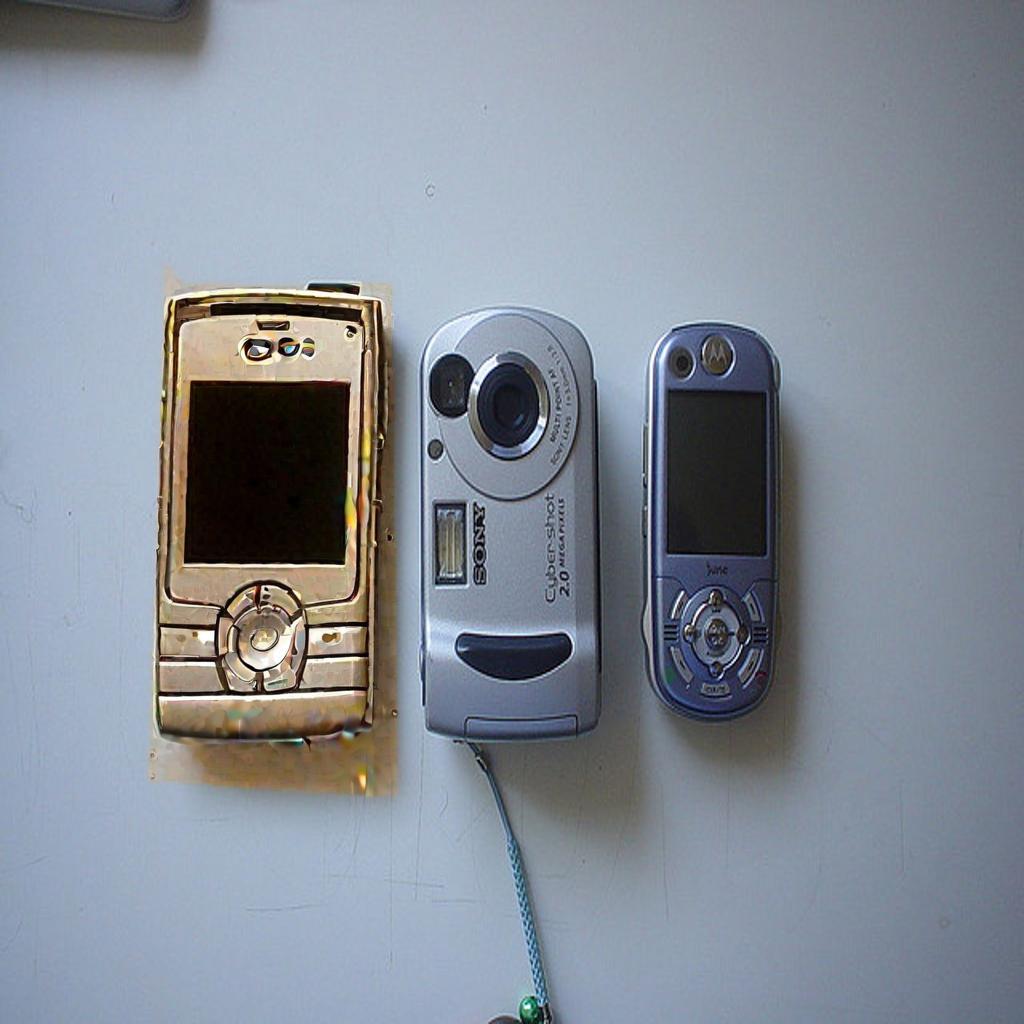}
        & \includegraphics[width=\linewidth]{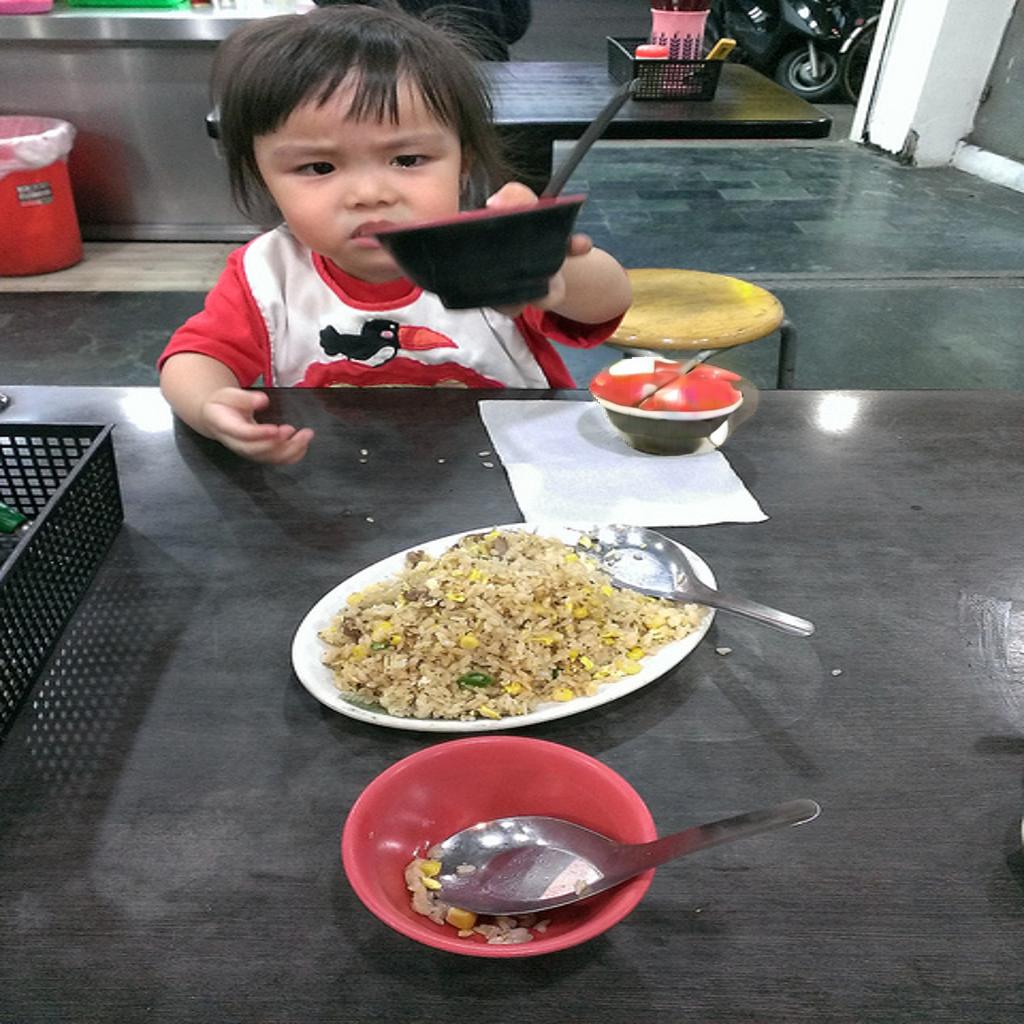} \\

    \end{tabular}
    \caption{Comparison against SDXLI-based approaches. Zoom in for a better view.}
    \label{fig:qualitative-sdxli}
\end{figure*}

\begin{figure*}[t]
    \centering
    \renewcommand{\arraystretch}{0.4} 
    \setlength{\tabcolsep}{0.5pt} 

    \begin{tabular}{
        >{\centering\arraybackslash}b{0.03\linewidth} 
        >{\centering\arraybackslash}p{0.12\linewidth} 
        >{\centering\arraybackslash}p{0.12\linewidth} 
        >{\centering\arraybackslash}p{0.12\linewidth} 
        >{\centering\arraybackslash}p{0.12\linewidth} 
        >{\centering\arraybackslash}p{0.12\linewidth}}

        & Three parrots with green and orange feathers
        & A yellow cloth sawfish doll
        & An American flag hung on a pole
        & A zebra standing on a dirt ground
        & A teddy bear in light brown color \\

        \rotatebox{90}{Input}
        & \includegraphics[width=\linewidth]{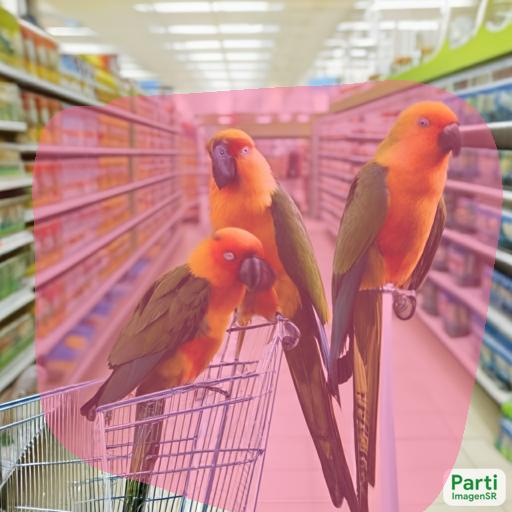}
        & \includegraphics[width=\linewidth]{images/qualitative/editbench/135_init.jpg}
        & \includegraphics[width=\linewidth]{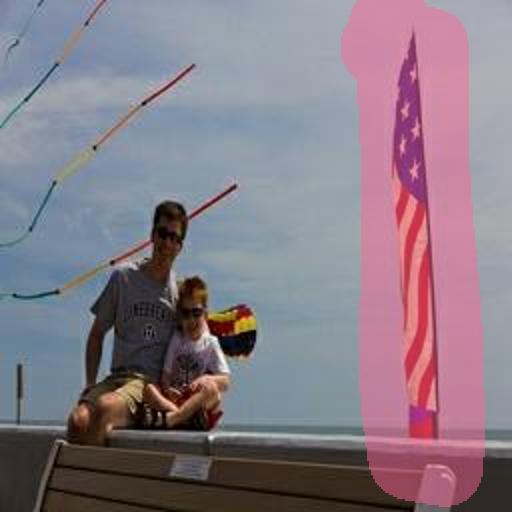}
        & \includegraphics[width=\linewidth]{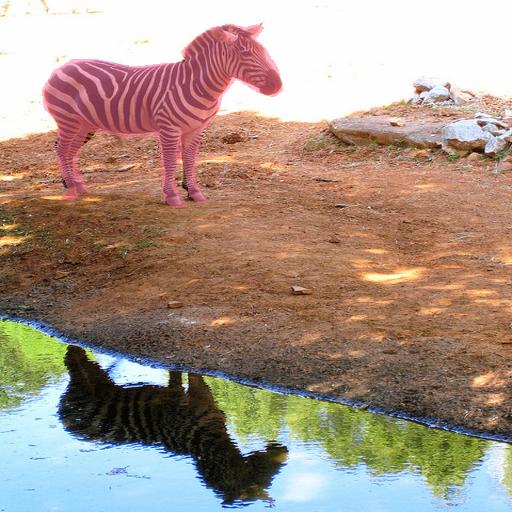}
        & \includegraphics[width=\linewidth]{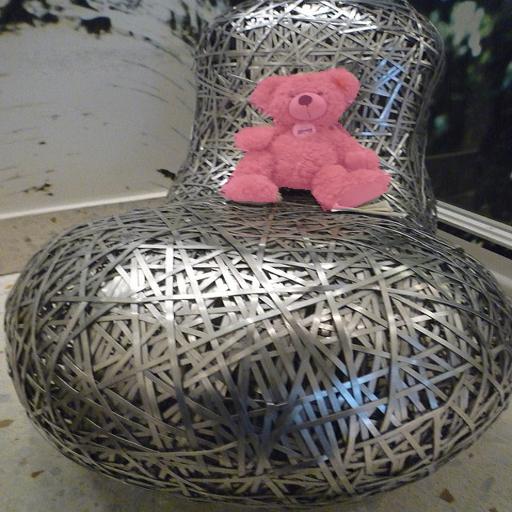} \\

        \rotatebox{90}{SD3I}
        & \includegraphics[width=\linewidth]{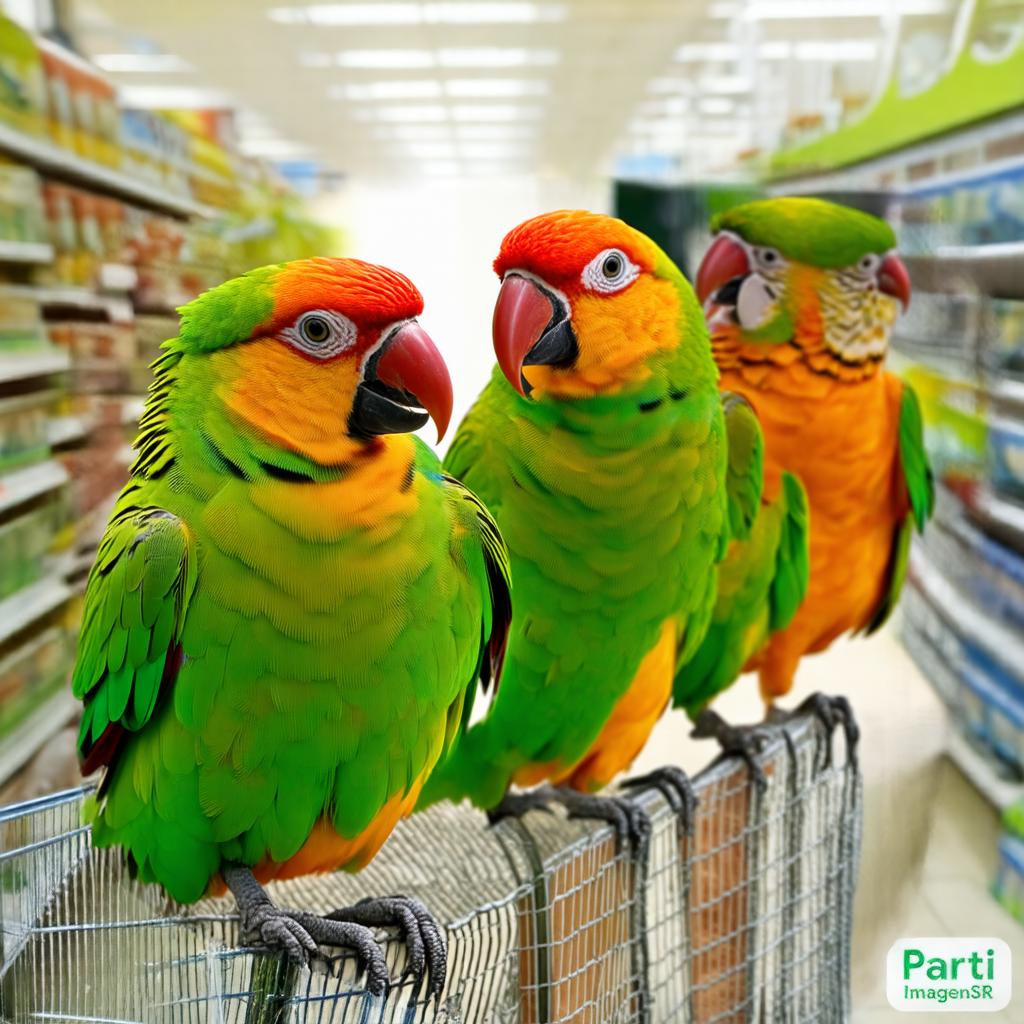}
        & \includegraphics[width=\linewidth]{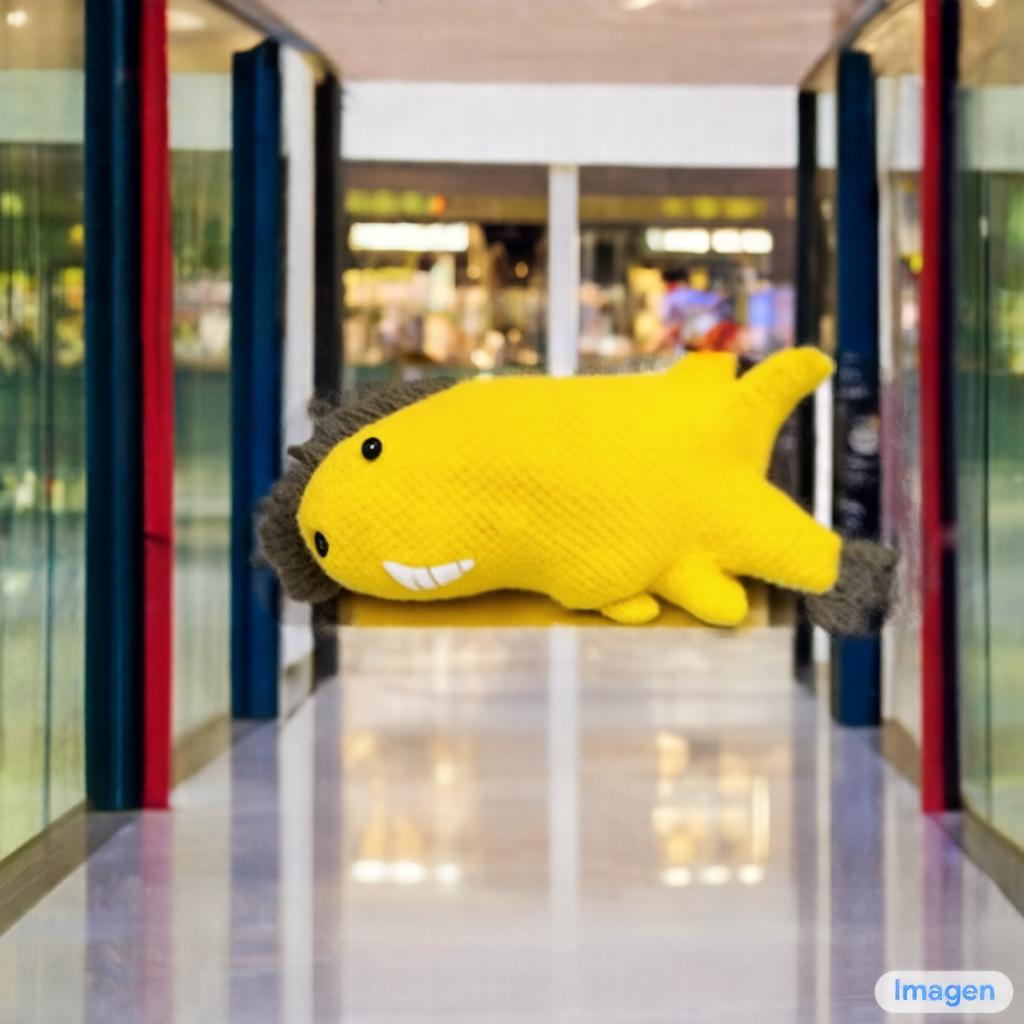}
        & \includegraphics[width=\linewidth]{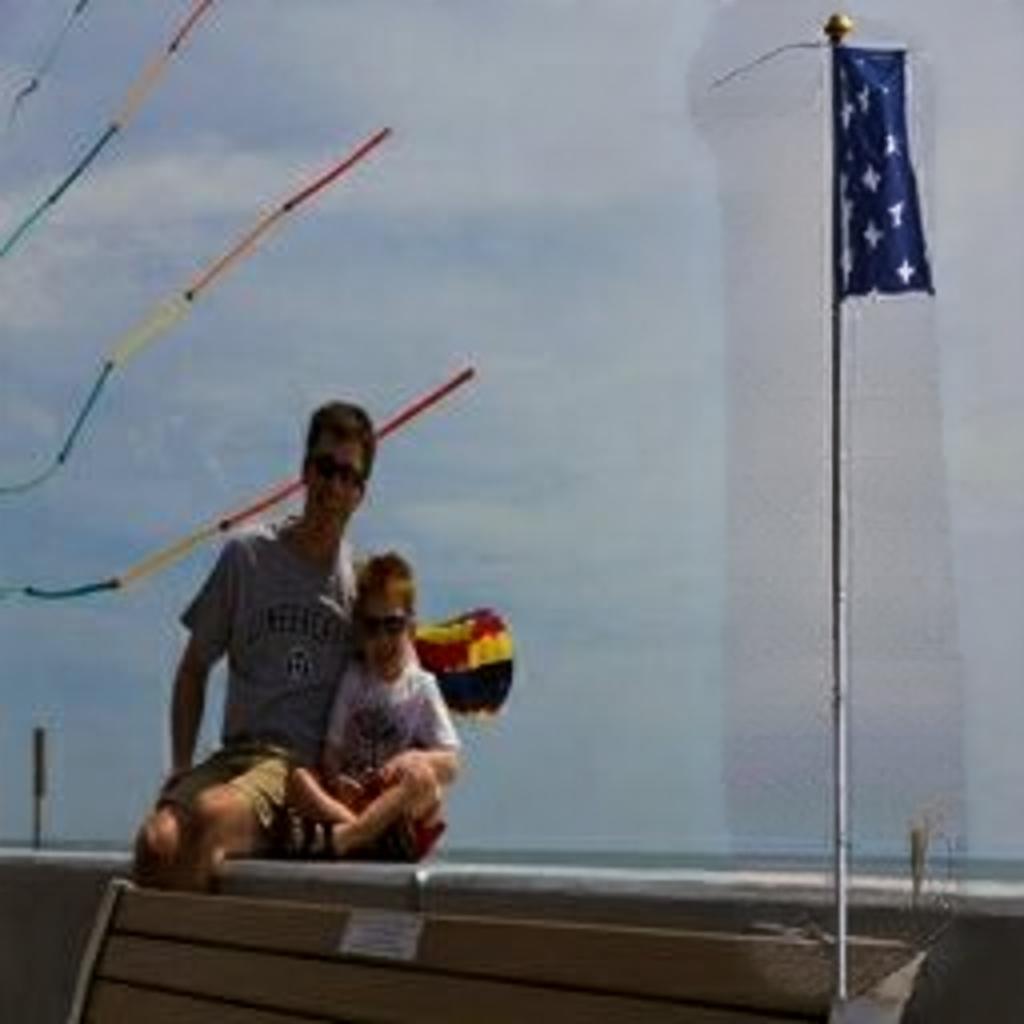}
        & \includegraphics[width=\linewidth]{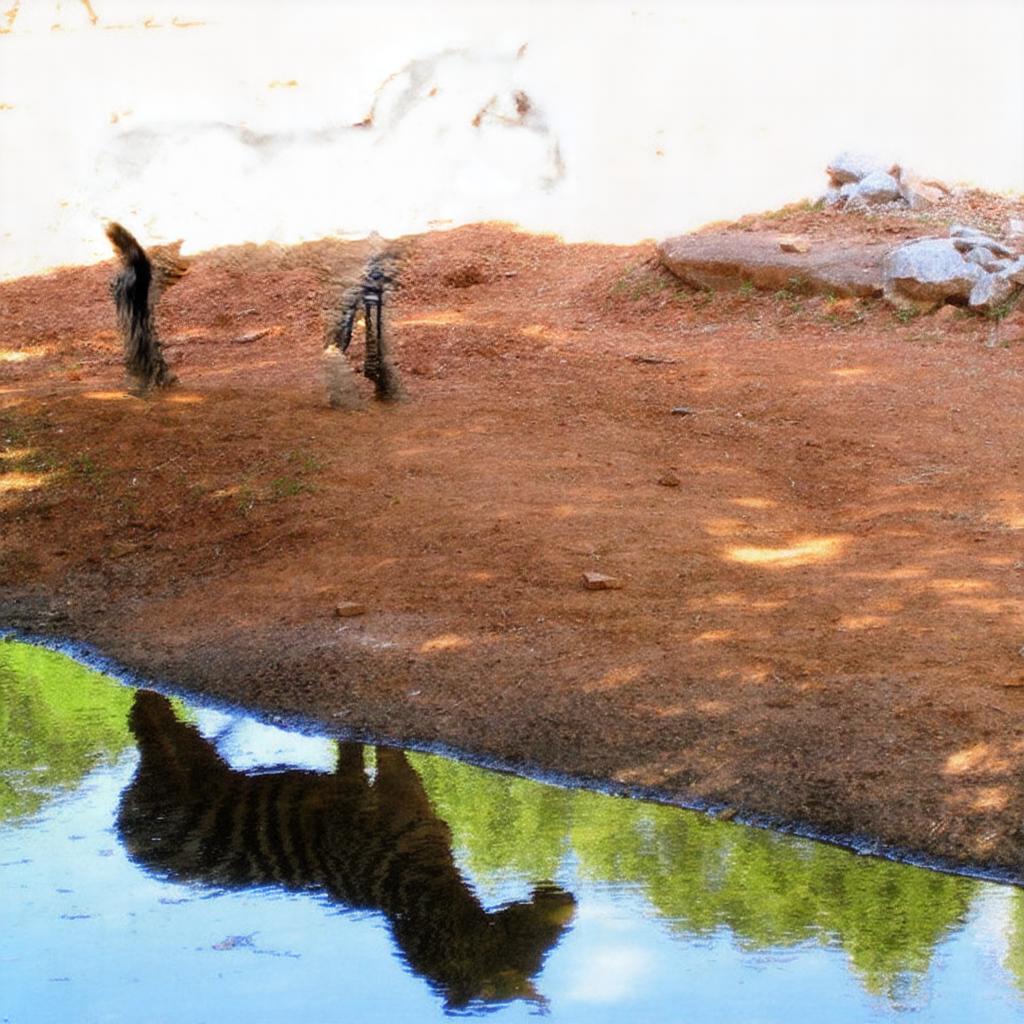}
        & \includegraphics[width=\linewidth]{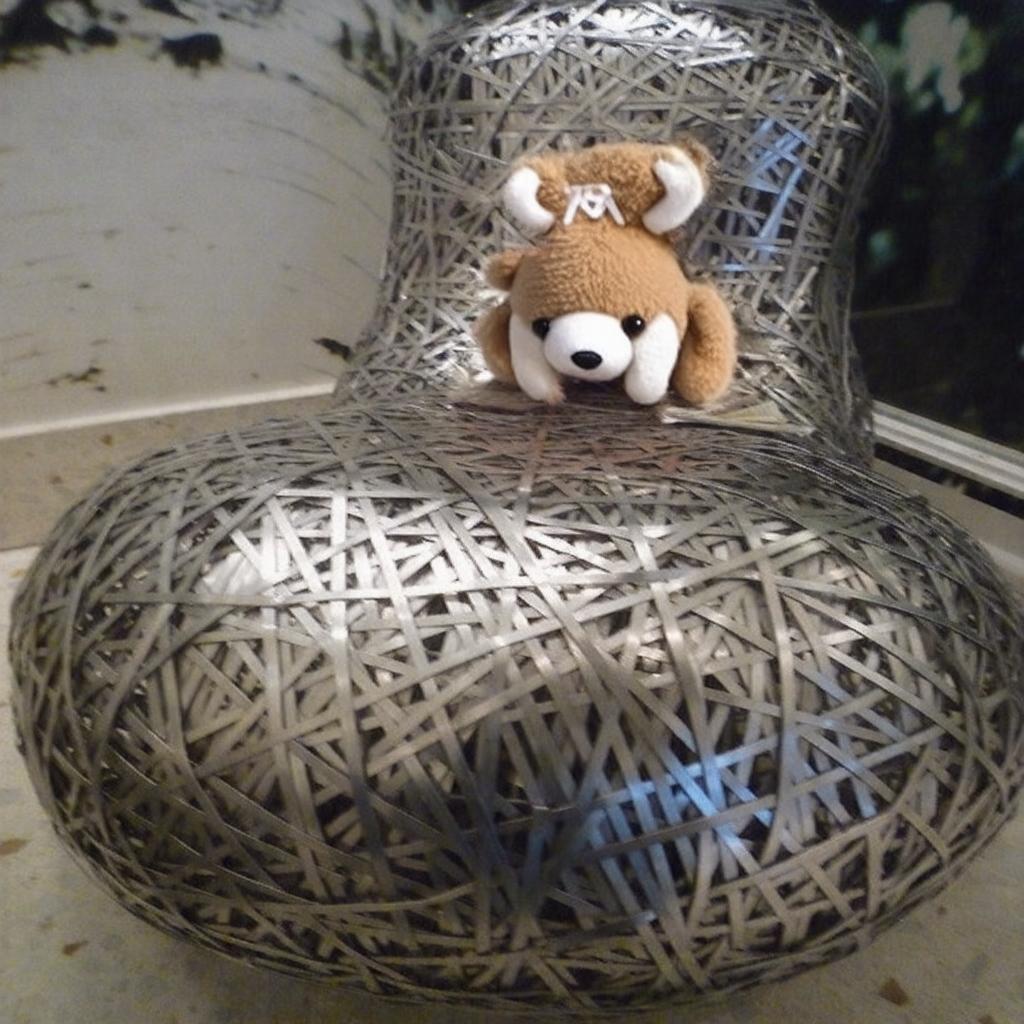} \\

        \rotatebox{90}{SD3I+HDP}
        & \includegraphics[width=\linewidth]{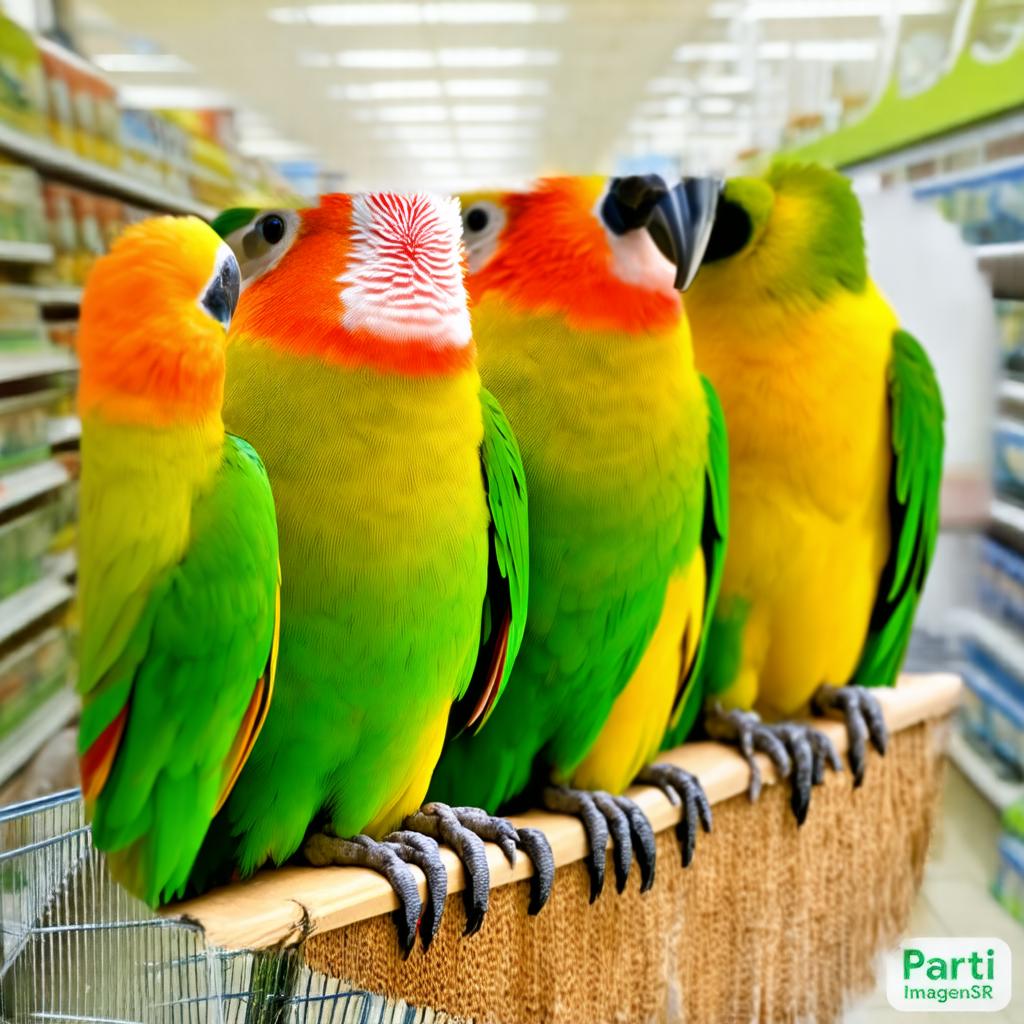}
        & \includegraphics[width=\linewidth]{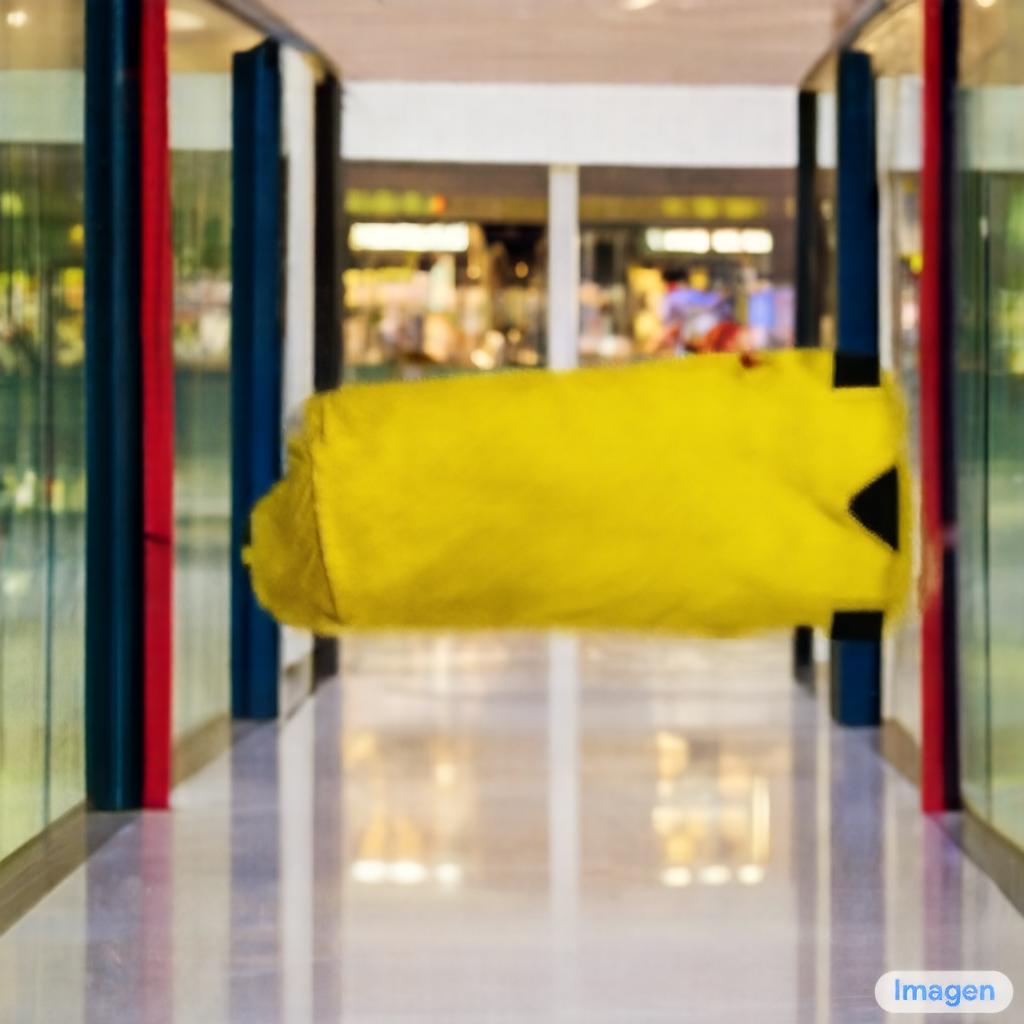}
        & \includegraphics[width=\linewidth]{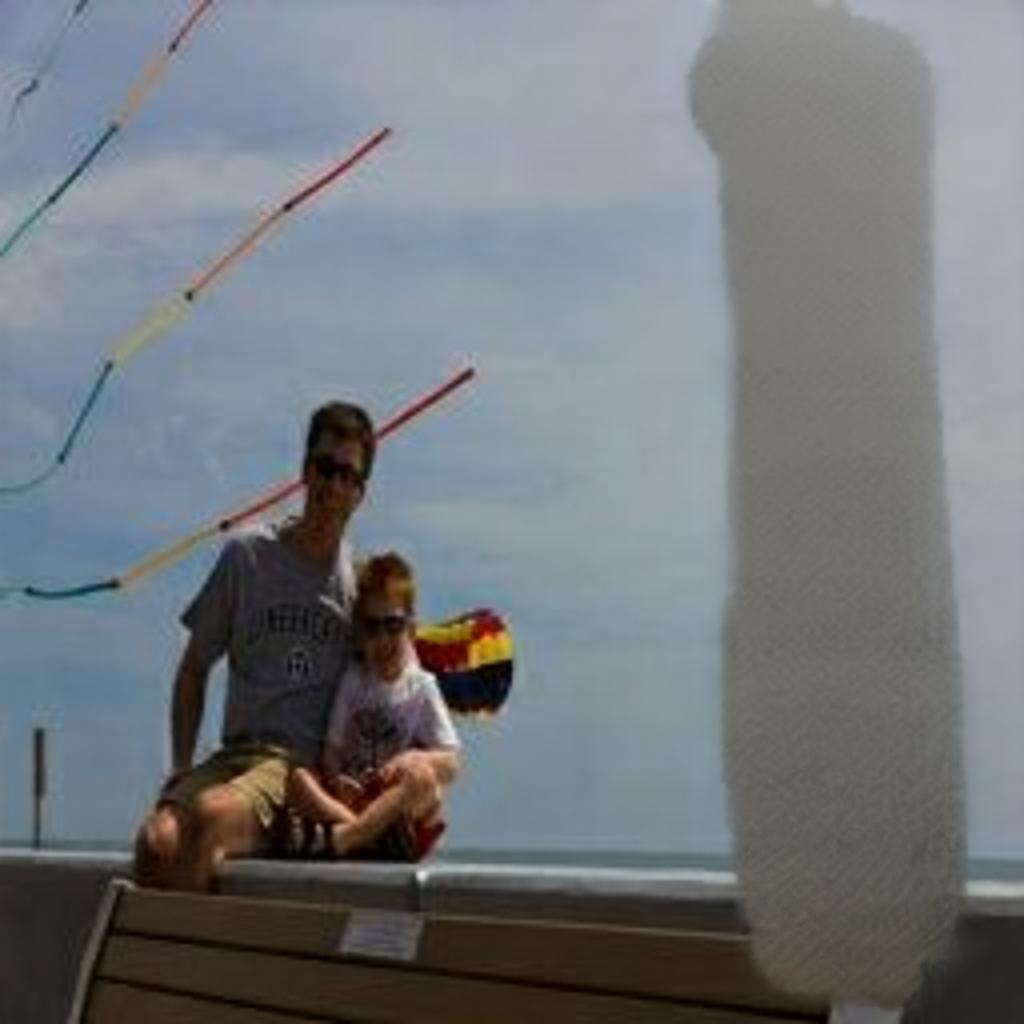}
        & \includegraphics[width=\linewidth]{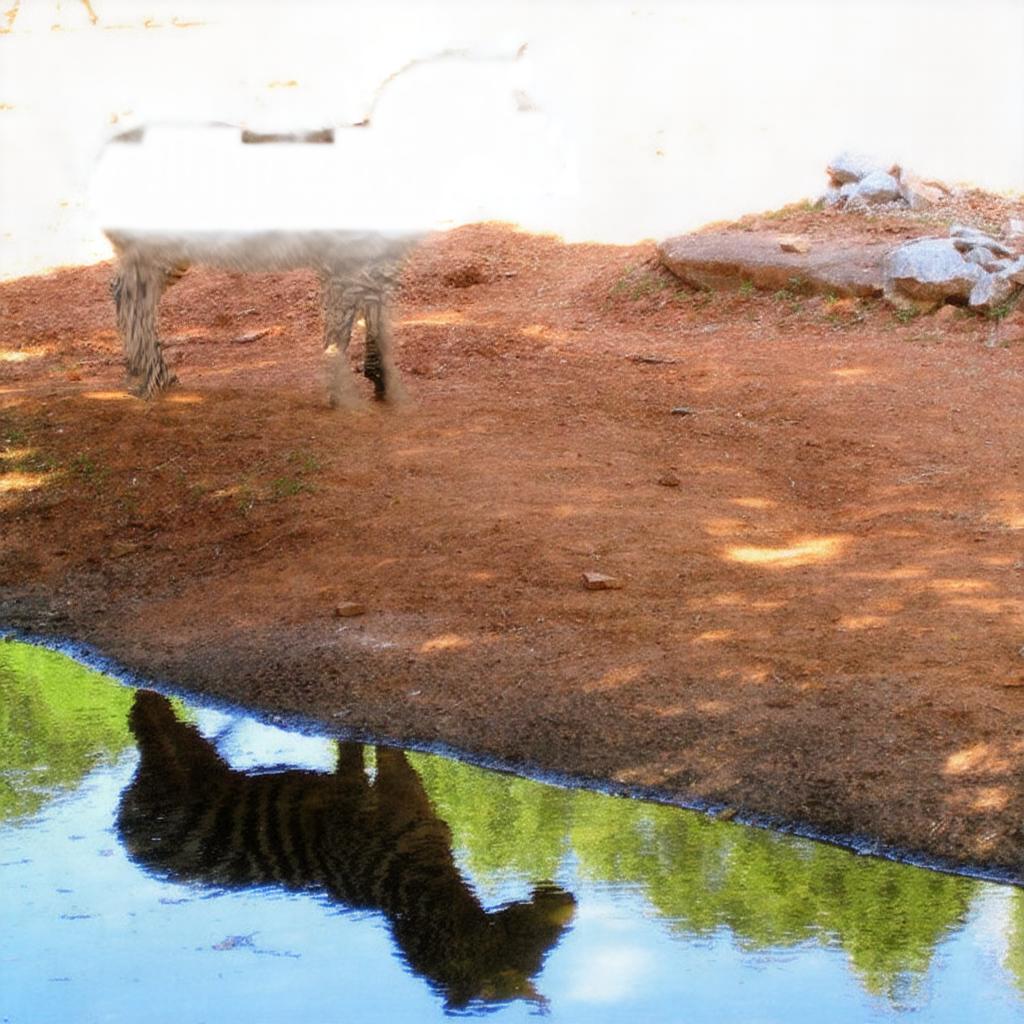}
        & \includegraphics[width=\linewidth]{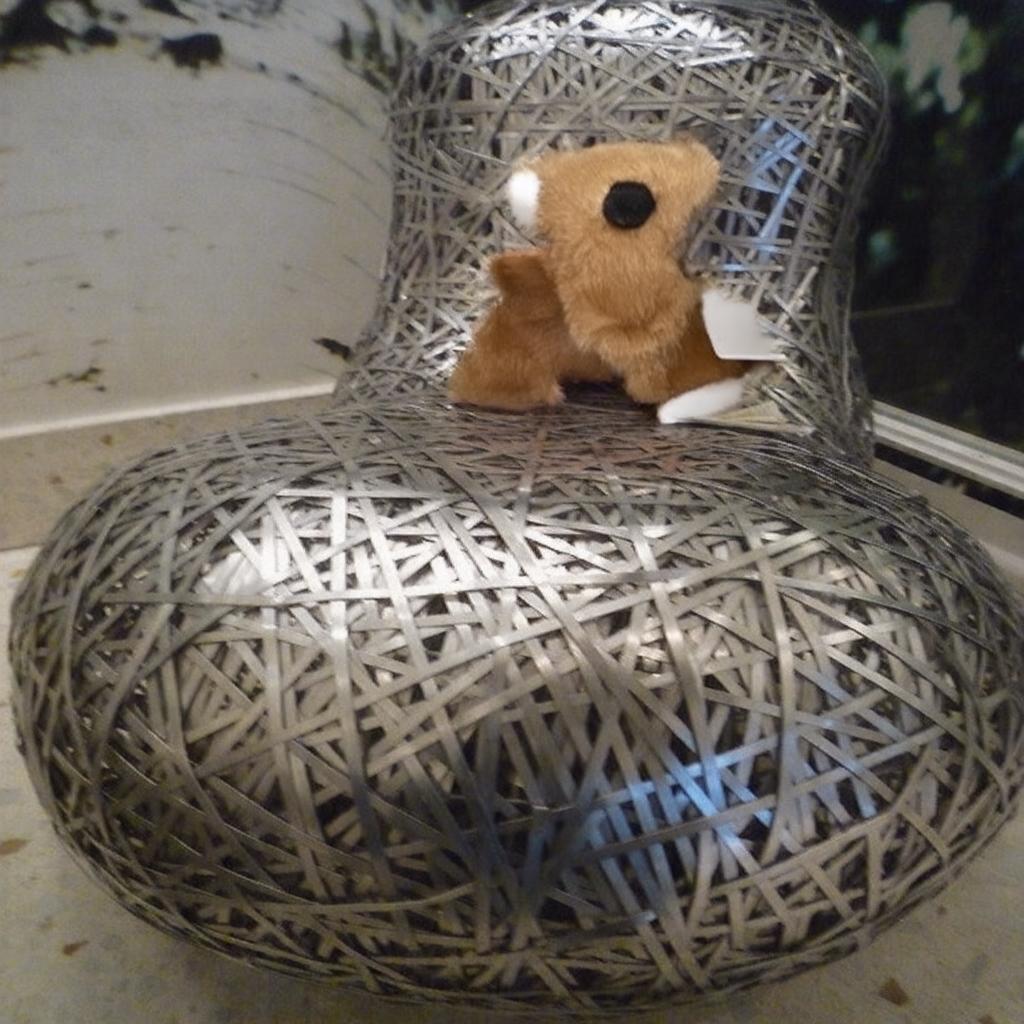} \\

        \rotatebox{90}{SD3I+Ours}
        & \includegraphics[width=\linewidth]{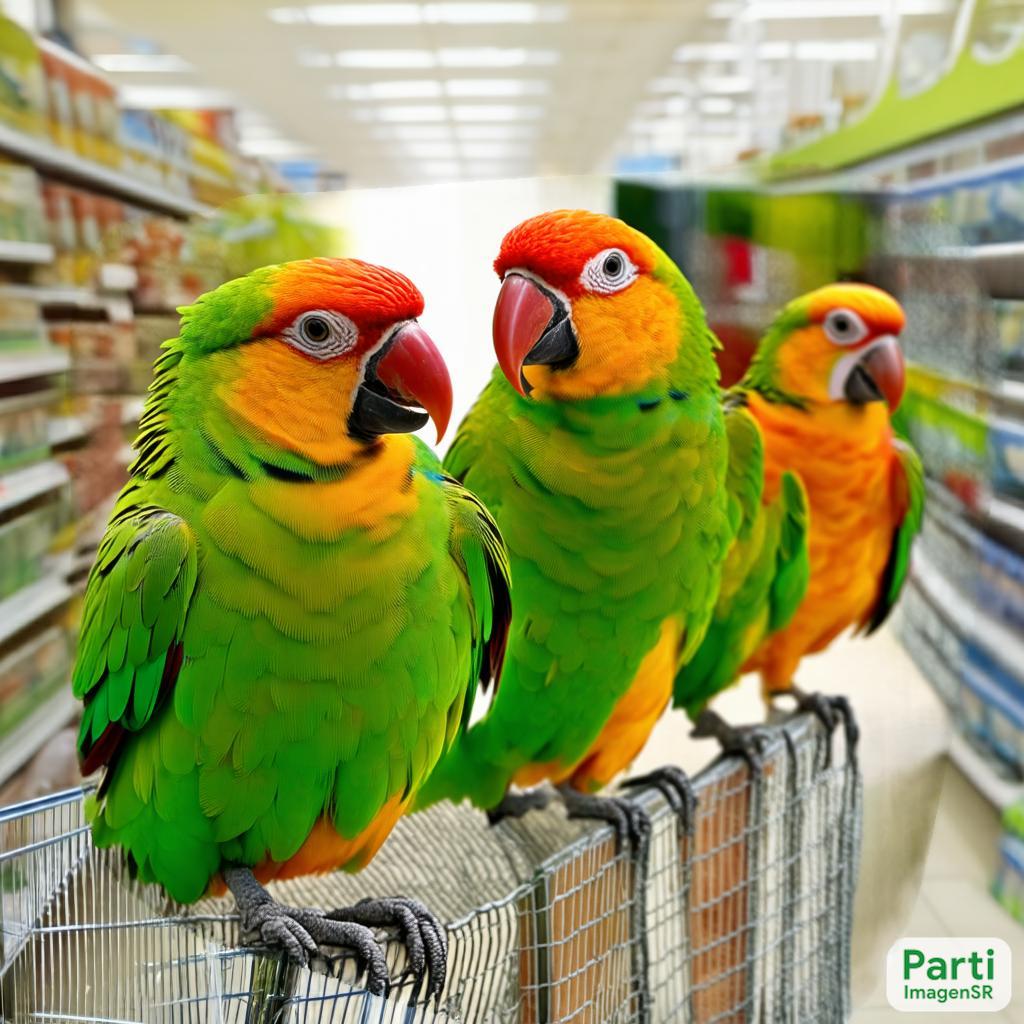}
        & \includegraphics[width=\linewidth]{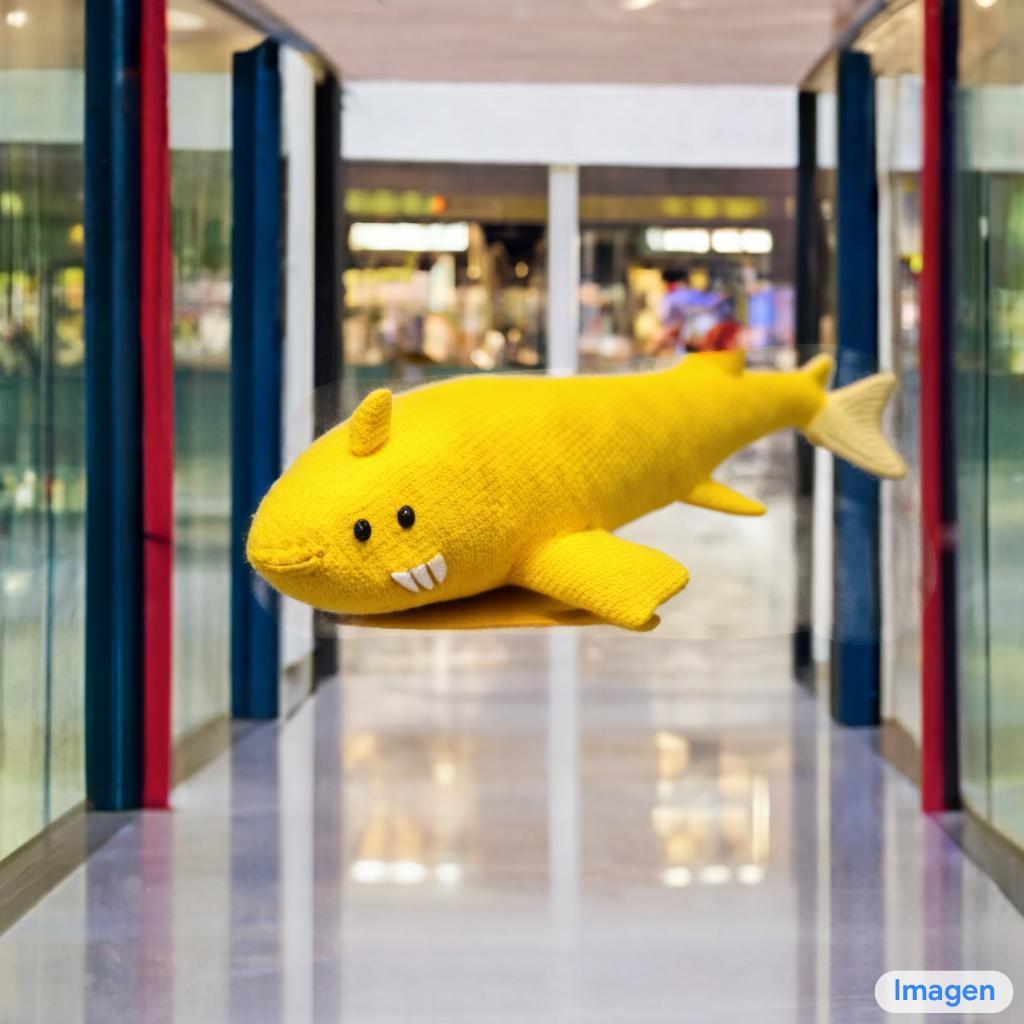}
        & \includegraphics[width=\linewidth]{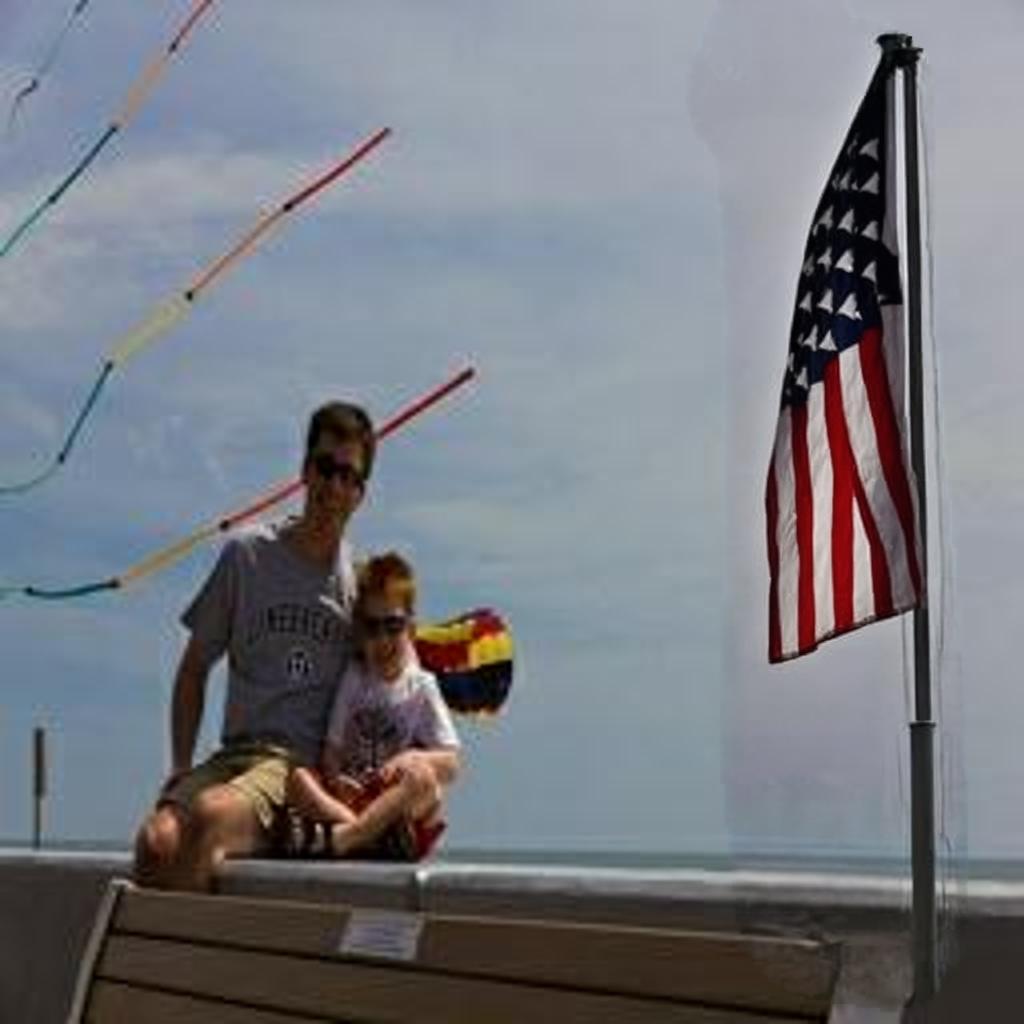}
        & \includegraphics[width=\linewidth]{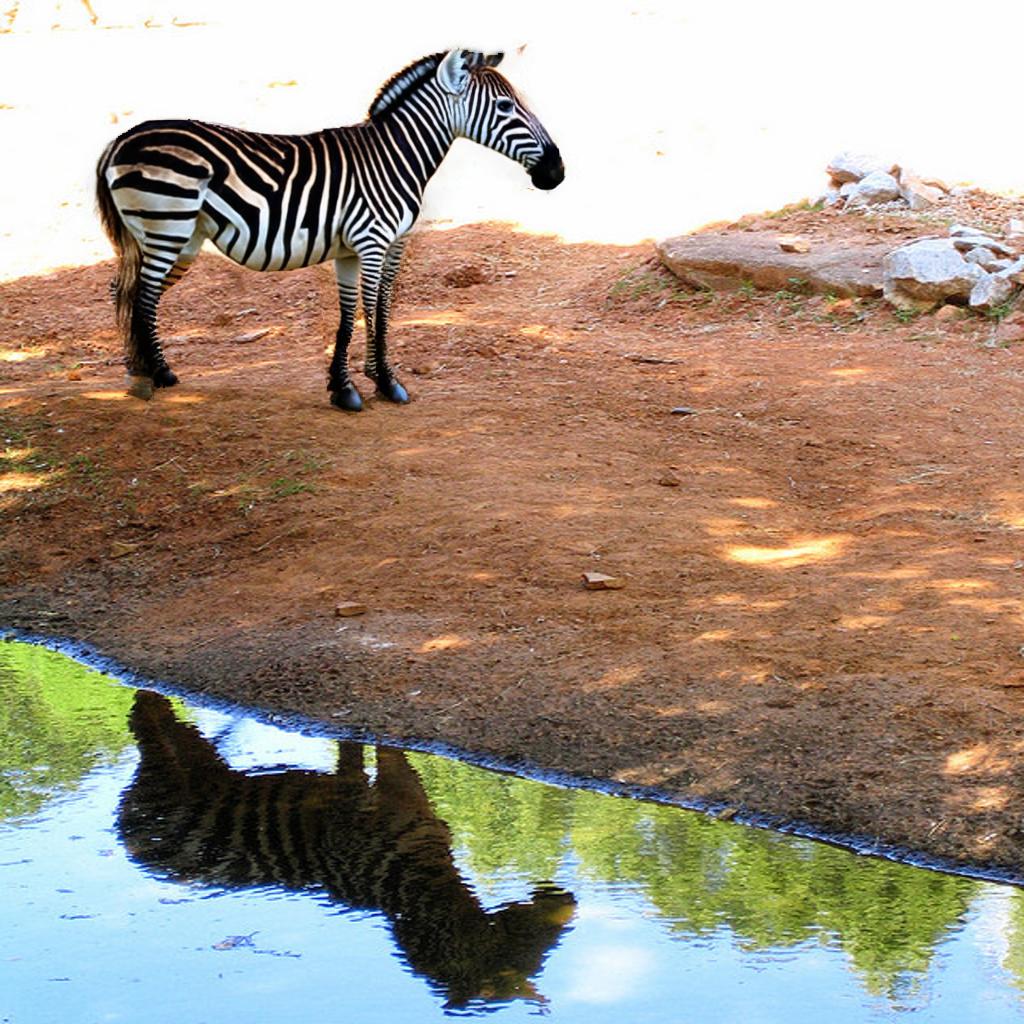}
        & \includegraphics[width=\linewidth]{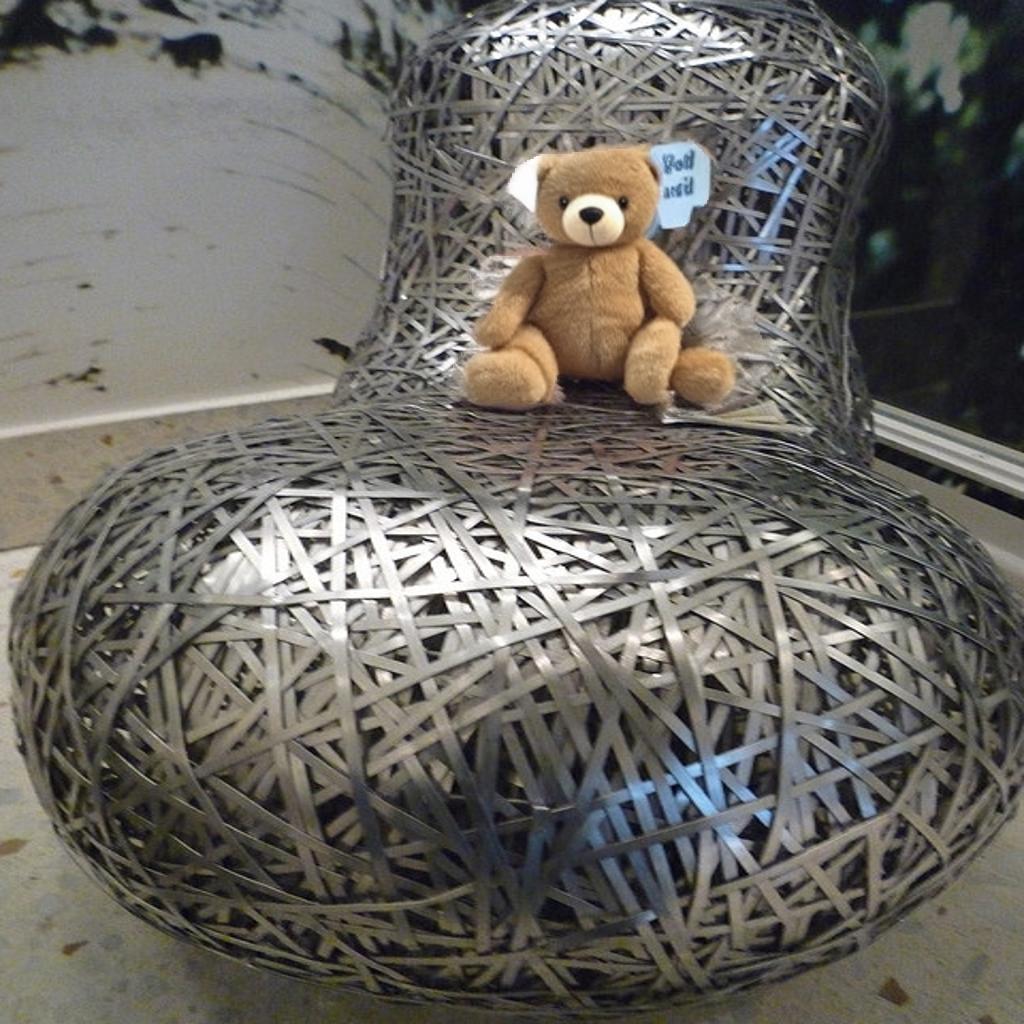} \\

    \end{tabular}
    \caption{Comparison against SD3I-based approaches. Zoom in for a better view.}
    \label{fig:qualitative-sd3i}
\end{figure*}

Fig \ref{fig:qualitative-ppt}, \ref{fig:qualitative-sdxli}, and \ref{fig:qualitative-sd3i} present additional qualitative results, featuring examples from EditBench (first three columns) and MSCOCO (last two columns). These examples focus on challenging scenarios where base models often struggle, producing results that may be visually incoherent, inaesthetic, or misaligned with the prompt. In each of these difficult cases, our method demonstrates a clear and consistent improvement, successfully addressing the shortcomings of the baseline and yielding a higher-quality output.

\section{Limitations and Future Work}
\begin{figure*}[t]
    \centering
    \renewcommand{\arraystretch}{0.4} 
    \setlength{\tabcolsep}{0.5pt} 

    \begin{tabular}{
        >{\centering\arraybackslash}b{0.03\linewidth} 
        >{\centering\arraybackslash}p{0.24\linewidth} 
        >{\centering\arraybackslash}p{0.24\linewidth} 
        >{\centering\arraybackslash}p{0.24\linewidth}}
        
        & Round clock 
        & White chair
        & Black bird\\
        
        \rotatebox{90}{Input}
        & \includegraphics[width=\linewidth]{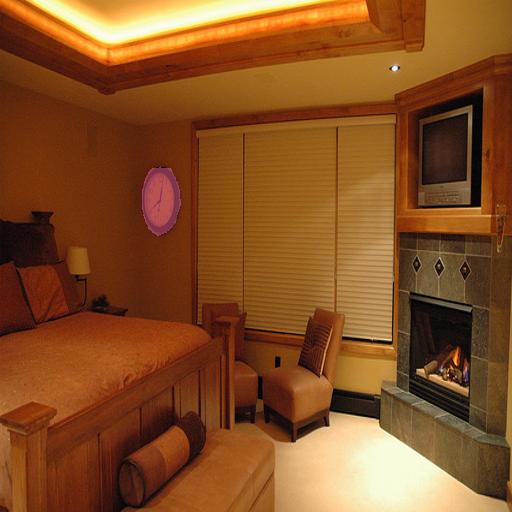}
        & \includegraphics[width=\linewidth]{images/limitations/224_init.jpg}
        & \includegraphics[width=\linewidth]{images/limitations/62_init.jpg}\\
        
        \rotatebox{90}{Output}
        & \includegraphics[width=\linewidth]{images/limitations/104_000000041872.jpg}
        & \includegraphics[width=\linewidth]{images/limitations/224_000000112378.jpg}
        & \includegraphics[width=\linewidth]{images/limitations/62_000000419201.jpg}\\

    \end{tabular}
    \caption{When the masked area is extremely small, the model tends to complete the mask directly with background. Zoom in for a better view.}

    \label{fig:limitations}
\end{figure*}

Although \emph{FreeInpaint} achieves superior performance across various benchmarks and evaluation perspectives, it still possesses a limitation, i.e., difficulty with extremely small regions. As shown in Fig. \ref{fig:limitations}, when the masked area is extremely small, even with our attention steering and decomposed guidance mechanism, the model still tends to complete the mask directly with background. A hierarchical attention steering or a scenario-specific guidance strategy could be a solution and will be our future work. Additionally, our method uses reward models in DeGu as a proxy for human preferences, which may introduce bias.

\end{document}